\documentclass[11pt,letterpaper]{article}

\usepackage[utf8]{inputenc}
\usepackage[T1]{fontenc}
\usepackage{lmodern}
\usepackage[margin=1in]{geometry}
\usepackage{microtype}
\usepackage{amsmath,amssymb,amsthm,mathtools}
\usepackage{graphicx}
\usepackage{tikz}
\usetikzlibrary{arrows.meta,positioning,fit,backgrounds,calc}

\usepackage{adjustbox}
\usepackage{booktabs}
\usepackage{tabularx}
\usepackage{multirow}
\usepackage{enumitem}
\usepackage{algorithm}
\usepackage{algpseudocode}
\usepackage{xcolor}
\usepackage[colorlinks=true,linkcolor=blue!60!black,citecolor=blue!60!black,urlcolor=blue!60!black]{hyperref}
\usepackage[capitalize,noabbrev]{cleveref}

\theoremstyle{definition}

\theoremstyle{plain}

\newtheorem{invariant}{Invariant}[section]

\newcommand{\StandingOut}{\textsc{out}}
\newcommand{\StandingTent}{\textsc{tentative}}
\newcommand{\StandingSupp}{\textsc{supported}}
\newcommand{\StandingLock}{\textsc{locked}}
\newcommand{\sgn}{\operatorname{sgn}}
\newcommand{\clip}{\operatorname{clip}}
\newcommand{\Ind}{\mathbb{I}}

\newcommand{\Darmok}{{\normalfont\scshape Darmok}}
\newcommand{\clue}[1]{\textit{``#1''}}
\newcommand{\ans}[1]{\texttt{#1}}

\title{\textbf{{\normalfont\scshape Darmok}: Championship Cryptic Crossword Solving via\\Constraint Tightening and Resilient Orchestration}}
\author{Siddhartha Chatterjee \\
Department of Computer Science \\
The University of Texas at Austin \\
\texttt{sidchat@utexas.edu} $\cdot$ \href{https://orcid.org/0009-0000-3320-8340}{\texttt{orcid.org/0009-0000-3320-8340}}}
\date{November 2026}

\begin{document}

\maketitle

\begin{abstract}
We present \Darmok{}, 
the first automated system to reach superhuman performance on British cryptic crosswords.
Unlike American-style crosswords, 
cryptic clues disguise an unmarked definition and compositional wordplay beneath a misleading surface reading, 
sparse grids leave nearly half of each word's letters unchecked, 
and tournament rules rank a grid with even a single wrong or blank square behind all error-free solutions regardless of speed.
Competing alongside an elite human field at the 55th \emph{Times} Crossword Championship in London, 
\Darmok{} solved all seven championship puzzles error-free and finished faster than every human competitor in every round.
It solved the Final puzzle in $6\text{m}\,52\text{s}$, a full $3\text{m}\,13\text{s}$ ahead of the human champion.

Today's frontier AI models can on average solve a whole puzzle in a single try with comparable accuracy and speed.
However, cloud API calls still stall, time out, trigger safety refusals, or get stuck in reasoning loops.
To simultaneously achieve tournament-grade accuracy, speed, and reliability, 
\Darmok{} cross-checks clues across three independent models in rounds;
uses verified crossing letters to unlock harder clues;
and uses a whole-puzzle model and dictionary lookups to fill any remaining gaps without overwriting confirmed words.
This design maintained near-zero-error performance across a five-month daily evaluation period.
Under normal benchmark conditions, 
removing backup models, fallback steps, or elaborate prompt instructions barely changes accuracy, 
making the architecture appear over-engineered.
Yet during both the worst days of daily operation and a severe slowdown of our primary model in the live Championship Final, 
the backup paths carried the solve, 
showing that as foundation models master the average case, the central challenge in building complex AI systems shifts to resilience in the worst case.
\end{abstract}

\section{Introduction}
\label{sec:intro}

Since the codebreakers of Bletchley Park used a 12-minute timed solve of the \textit{Daily Telegraph} crossword puzzle number 5{,}062 on January 13, 1942 as a recruitment test for wartime cryptanalysts~\cite{smith1998bletchley}, crossword solving has served as a bellwether for both natural and artificial intelligence under simultaneous linguistic and combinatorial constraints.
For American-style crosswords, where dense grids check every white square in both directions ($100\%$ checking) and even punny or theme-based clues resolve as whole-phrase semantic queries rather than character-level string programs, weighted constraint-satisfaction solvers beginning with \textsc{Proverb}~\cite{shazeer1999aaai,littman2002proverb}, \textsc{WebCrow}~\cite{ernandes2005webcrow}, and \textsc{Dr.Fill}~\cite{ginsberg2011drfill}, and culminating in neural-retrieval hybrids such as the Berkeley Crossword Solver~\cite{wallace2022berkeley}, achieved superhuman tournament performance by combining high-recall candidate retrieval with probabilistic belief propagation or branch-and-bound search.

British cryptic crosswords subvert both halves of that regime.
First, a cryptic clue is a dual specification written in two overlapping grammars (\Cref{fig:primer-clues}).
Read under standard English grammar, the clue forms a coherent, deliberately misleading narrative phrase (the \textit{surface reading}).
Read under the compositional rules of ``Crossword English''~\cite{ximenes1966art,manley2014chambers}, the same token sequence splits at an unmarked boundary into a semantic \textit{definition} (almost always positioned at the extreme start or end of the clue) and an executable \textit{wordplay expression} that nests string-manipulation primitives (anagrams, charades, containers, deletions, reversals, homophones, Spoonerisms, and letter substitutions) to construct the same character string.

Second, British grids are topologically sparse (\Cref{fig:primer-grids}): standard $15\times 15$ blocked \textit{Times} grids leave $40\%$ to $50\%$ of each word's letters \textit{unchecked} (belonging to only a single Across or Down word, which accounts for $61.3\%$ to $65.0\%$ of all physical white cells),\footnote{Because each checked intersection cell is shared by two perpendicular words whereas each unchecked cell belongs to only one word, a grid where $45.9\%$ of word-letter positions are unchecked (\textit{Times}~\#29399, $102/222$) has $102/162 = 63.0\%$ of its physical white squares unchecked (\cref{sec:csp-breakdown}).} while $12\times 12$ barred grids such as the Sunday \textit{Times} \textit{Mephisto} (an advanced weekly cryptic solved against the full \textit{Chambers Dictionary}) contain zero black squares, pack $36$ words separated by thick cell bars, and leave $20.0\%$ to $23.1\%$ of word-letter positions ($33.3\%$ to $37.5\%$ of physical white cells) unchecked alongside arcane dictionary headwords.
Because perpendicular crossings never inspect an unchecked cell, grid geometry alone cannot rescue a solver whose candidate generator fails to parse the internal wordplay.
Consequently, while early grammar-based analyzers such as \textsc{Lacross}~\cite{woodhead1977computer,williams1979computer}, commercial heuristic systems such as \textsc{Crossword Maestro}~\cite{tunstallpedoe2010true}, and single-clue benchmarks such as \textsc{Cryptonite}~\cite{efrat2021cryptonite} and \textsc{Decrypting}~\cite{rozner2021decrypting} established cryptic crosswords as a formidable reasoning challenge, no automated system had previously approached championship human play on full grids under live tournament clocks~\cite{sadallah2024good,saha2025lmcrossword,andrews2024proving,andrews2025reasoning}.

\subsection{Tournament Rules, the Accuracy--Speed Pareto Frontier, and the Live Milestone}
\label{sec:milestone}

\begin{figure}[!htbp]
\centering
\includegraphics[width=\textwidth]{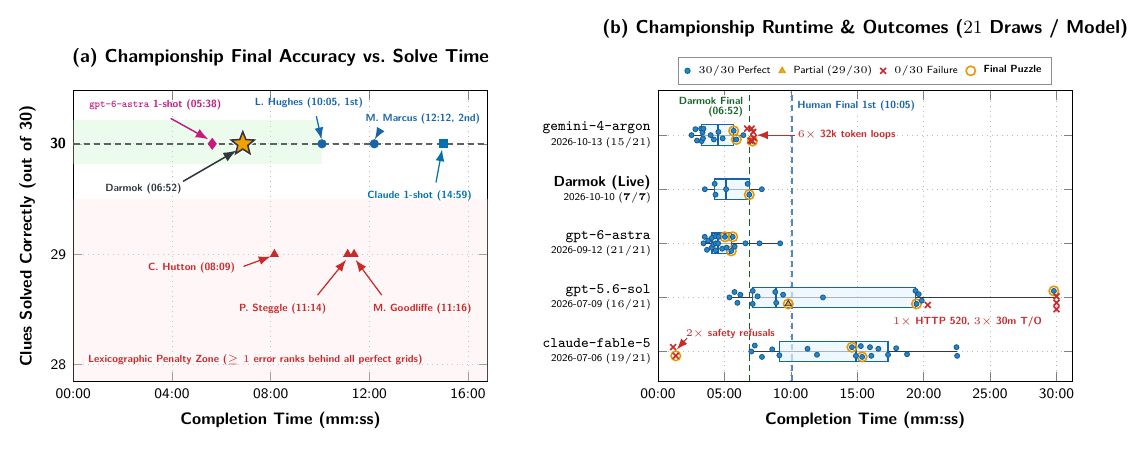}
\caption{\textbf{The tournament accuracy--speed trade-off and single-shot runtime/outcome distributions.}
\textbf{(a)} Accuracy versus completion time in the 2026 \textit{Times} Championship Final ($30$ clues), where any grid with $\ge 1$ error ranks behind all error-free solutions; \Darmok{} finished $30/30$ in $6\text{m}\,52\text{s}$.
\textbf{(b)} Per-run completion times and outcomes across \Darmok{}'s $7$ live championship puzzles and $21$ standalone draws ($7\text{ puzzles} \times 3\text{ draws}$) per single-shot frontier model (ordered reverse-chronologically by date; boxes and vertical bars show IQR and median over completed solves, and gold rings mark Championship Final draws).}
\label{fig:intro-pareto}
\end{figure}

First held in 1970, the annual \textit{Times} Crossword Championship in London is the premier competitive test of cryptic solving~\cite{times2026championship}.
The championship's rules define a strict lexicographic objective over \textbf{correctness first and completion speed second} (\Cref{fig:intro-pareto}a).
Grids are scored by clues rather than by individual squares: a single wrong or blank square marks every clue passing through that square incorrect, and a grid with even one missed clue ranks behind every $100\%$-correct solution regardless of speed.
Competitors solve three puzzles within a $60\text{m}$ window in the Qualifying and Semi-Final rounds and a single Championship Final puzzle on stage; among error-free submissions, standings are ordered by completion time.

The human field data from the 55th \textit{Times} Crossword Championship on October 10, 2026 illustrates how unforgiving this scoring frontier is in practice.
In the Qualifying round ($113$ contestants competing for $57$ Semi-Final berths across three puzzles, $90$ clues), $60$ human competitors achieved a perfect $90/90$, so the qualification cut line between $57\text{th}$ and $60\text{th}$ place was decided purely by solve speed.
The fourth-fastest human finisher on the clock scored $89/90$, dropping to $61\text{st}$ in the standings; likewise for the fifth-fastest human finisher ($88/90$), who dropped to $69\text{th}$.
In the Semi-Final round ($57$ contestants competing for five Final berths on stage), only $11$ achieved $90/90$; the fourth-fastest human finisher on the clock scored $86/90$, dropping to $25\text{th}$ in the standings.
In the five-person Championship Final ($30$ clues), three finalists---including fastest finisher Chloe Hutton ($8\text{m}\,09\text{s}$), Peter Steggle ($11\text{m}\,14\text{s}$), and 14-time champion Mark Goodliffe ($11\text{m}\,16\text{s}$)---committed a single error ($29/30$), awarding the 2026 title to the fastest error-free solver, Liam Hughes ($10\text{m}\,05\text{s}$, $30/30$), ahead of runner-up Matthew Marcus ($12\text{m}\,12\text{s}$, $30/30$).

For an automated solver, this rule eliminates both naive operating regimes (\Cref{fig:intro-pareto}a): under an illustrative independent-clue calculation, a fast solver with $97\%$ per-clue accuracy has only a $0.97^{30} = 40.1\%$ chance of producing a clean 30-clue grid, while a cautious solver that takes $15\text{m}$ to $23\text{m}$ for unconstrained single-shot extended thinking (such as standalone \texttt{claude-fable-5}, which averaged $14\text{m}\,59\text{s}$ across its two completed Final draws [$14\text{m}\,36\text{s}$ and $15\text{m}\,22\text{s}$], with a 21-draw median of $14\text{m}\,54\text{s}$ and maximum of $22\text{m}\,31\text{s}$) loses the speed race to human champions who clear the Final in $8\text{m}$ to $12\text{m}$.

On October 10, 2026, with the advance permission of the \textit{Times} Crossword Championship organizers, we fielded \Darmok{}\footnote{After the eponymous \emph{Star Trek: The Next Generation} episode (Season~5, Episode~2, Stardate 45047.2, 1991), in which communication with the Tamarian species requires decoding utterances whose literal surface grammar conceals allegorical and compositional references---a direct parallel to cryptic crossword clues.} alongside the human competitors at the 55th \textit{Times} Crossword Championship (participating outside the official human prize standings).
Ingesting each round's official puzzle PDF at the starting bell and emitting a locked solution grid and per-clue wordplay explanations before time was called, \Darmok{} solved all seven championship puzzles without a single error ($210/210$ clues), finishing faster than every human competitor in every round.
\begin{itemize}[leftmargin=1.8em,itemsep=2pt]
    \item \textbf{Round 1 (Qualifiers 1--3, 90 clues).}
    $90/90$ correct in $4\text{m}\,20\text{s}$, $4\text{m}\,16\text{s}$, and $3\text{m}\,32\text{s}$.
    Because the three puzzles in a round are solved concurrently across three execution panes, total wall-clock elapsed time for the round is the maximum of the three puzzle times ($4\text{m}\,20\text{s}$), whereas sequential human solve time corresponds to their sum ($12\text{m}\,08\text{s}$).\footnote{The sum of the puzzle times was the operational arrangement agreed with the Championship organizers.}
    \item \textbf{Round 2 (Semi-Finals 1--3, 90 clues).}
    $90/90$ correct in $6\text{m}\,45\text{s}$, $5\text{m}\,08\text{s}$, and $7\text{m}\,49\text{s}$, completing all three concurrently in $7\text{m}\,49\text{s}$ of wall-clock time ($19\text{m}\,42\text{s}$ sequential sum).
    \item \textbf{Championship Final (30 clues).}
    $30/30$ correct in $6\text{m}\,52\text{s}$, finishing $3\text{m}\,13\text{s}$ ahead of 2026 human champion Liam Hughes ($10\text{m}\,05\text{s}$, zero errors) and $1\text{m}\,17\text{s}$ ahead of fastest human finisher Chloe Hutton ($8\text{m}\,09\text{s}$, one error).
\end{itemize}

To establish that this performance was neither a seven-puzzle fluke nor an artifact of pre-training memorization, we trace the system's evolution from our 2025 single-model greedy baseline (\textbf{V1}, fielded at the 2025 Championship) through three operational epochs of our 2026 architecture (\textbf{V2}):
\textbf{Epoch~E0} (May 19 -- July 5; $39$ days, early multi-seat greedy admission), \textbf{Epoch~E1} (July 6 -- August 25; $50$ days, synchronous corroboration admission), and \textbf{Epoch~E2} (August 26 -- October 9; $43$ days, the frozen championship configuration).
We report the performance across three evaluation corpora.
\begin{enumerate}[leftmargin=1.8em,itemsep=2pt]
    \item \textbf{Post-Cutoff Daily Production Record ($132$ days, $3{,}876$ clues).}
    Every usable\footnote{\textit{The Times} publishes its standard daily cryptic crossword six days a week (Monday--Saturday; Sundays feature the blocked Sunday Cryptic and the $12\times12$ barred \textit{Mephisto}).
    Across the $124$ Monday--Saturday publication days plus $14$ Sunday Cryptic editions in the May~19 -- October~9 window, $132$ runs met our pre-registered canonical grading criterion (complete solution-grid ground truth and unperturbed production configuration; $6$ experimental flag-test or incomplete-ground-truth runs are archived separately in the repository).
    Of the $132$ canonical puzzles, $126$ were solved live on the morning of publication and $6$ Epoch~E0 puzzles from a June~17--29 author vacation were batch-backfilled on June~29 and July~5 under strictly date-filtered historical wordlists.} daily \textit{Times} Cryptic puzzle published between May 19 and October 9, 2026, all post-dating the training cutoffs of every model in the ensemble.
    Across all $132$ days, \Darmok{} solved $3{,}844 / 3{,}876$ clues ($99.17\%$, $5$ wrong, $27$ blank, $108/132$ perfect grids); across the 43-day frozen championship epoch (Epoch~E2, each puzzle run live on publication morning), it solved $1{,}259$ of $1{,}261$ clues ($99.84\%$, $1$ wrong, $1$ blank, and $41/43$ flawless grids) in a median puzzle solve time of $5\text{m}\,36\text{s}$.
    \item \textbf{Stratified 16-Puzzle Test Set ($463$ clues).}
    Across a frozen benchmark spanning five difficulty bands on the community \textsc{Snitch} scale,\footnote{The Crossword \textsc{Snitch} (\url{https://times.xwdsnitch.link/crosswords}) aggregates completion times and error rates of experienced human solvers into a difficulty index centered at $100$ (average), where $<90$ is easy and $\ge 160$ is extreme; individual puzzle identities appear in \Cref{tab:app-test-puzzles}.} comprising $1$ Easy ($70$--$90$, $30$ clues), $3$ Moderate ($90$--$110$, $87$ clues), $4$ Hard ($110$--$130$, $116$ clues), $5$ Very Hard ($130$--$160$, $142$ clues), and $3$ Extreme ($160+$, $88$ clues) puzzles, the championship architecture solved $461/463$ clues ($99.57\%$, $2$ errors, $0$ blanks, $14/16$ perfect grids) in a median per-puzzle time of $4\text{m}\,11\text{s}$, compared to $380/463$ ($82.1\%$, median $11\text{m}\,40\text{s}$) for our 2025 V1 baseline and $408/463$ ($88.1\%$, median $5\text{m}\,44\text{s}$) for our June 2026 Epoch~E0 prototype.
    This corpus was primarily used for choosing among multiple design choices for a particular feature.
    \item \textbf{Zero-Shot Transfer to Barred \textit{Mephisto} Grids ($10$ puzzles, $360$ clues).}
    Run unmodified on ten consecutive Sunday \textit{Times} \textit{Mephisto} puzzles, advanced $12\times 12$ barred grids with zero black squares and esoteric \textit{Chambers Dictionary} vocabulary, the championship solver achieved $350/360$ ($97.2\%$, $4$ wrong, $6$ blank, $5/10$ perfect grids) in a median per-puzzle time of $13\text{m}\,32\text{s}$.
    This run was conducted between October 12 and October 14, 2026.
\end{enumerate}

\subsection{The Post-Championship Divergence: Tournament Engineering or Rube Goldberg Machine?}
\label{sec:paradox}

\Darmok{} was engineered over five months of daily runs with a single operational goal: achieving tournament-grade performance under the \textit{Times} Championship's unforgiving zero-error, minimum-time rule.
Closing the gap from our early Epoch~E0 runs ($97.97\%$ daily accuracy, $59.0\%$ perfect grids) to Epoch~E2 ($99.84\%$ accuracy, $95.3\%$ perfect grids) followed a protective layering principle: \textit{first lock in corroborated answers from the per-clue models and crossing-letter retries, and then fill any remaining open clues using strictly non-evicting backup stages (a background whole-grid model and a local dictionary sweep) that are never allowed to overwrite already-accepted words.}
This architecture (detailed in \cref{sec:arch} and \Cref{fig:arch}) coordinates three per-clue reasoning model seats---Gemini (\texttt{gemini-3.1-pro-preview}, Anchor), Claude (\texttt{claude-fable-5}, Peer), and GPT-5 (\texttt{gpt-5.6-sol}, Third Seat)---alongside an asynchronous whole-grid process (\texttt{gpt-6-astra}).

Had we stopped after the championship and written a conventional system description, every one of these stages would have been presented as an essential gear in a neuro-symbolic architecture.
Yet an apparatus built by accreting fallback stages to squeeze out the last $1$--$2$ clues per grid invites a skeptical systems question: \textit{did we build a Rube Goldberg machine\footnote{After cartoonist and engineer Rube Goldberg (1883--1970), whose contraptions perform a simple task through a long, precarious chain of serial mechanisms~\cite{wolfe2000rube}.} whose auxiliary gears were rendered obsolete as the upstream models improved?}

\begin{table}[!htbp]
\centering
\caption{Standalone single-shot frontier models ($3\text{ draws/puzzle} = 21\text{ runs}$ per model, ordered reverse-chronologically by GA date) versus \Darmok{} across all $7$ puzzles of the 2026 \textit{Times} Championship (\cref{sec:tail-oneshot}; median and SD over completed draws).}
\label{tab:intro-oneshot-mini}
\small
\begin{adjustbox}{max width=\textwidth}
\begin{tabular}{@{}llrrrl@{}}
\toprule
\textbf{System / Standalone Model} & \textbf{GA Date} & \textbf{Perfect Grids} & \textbf{Clues} & \textbf{Med Time (SD)} & \textbf{Primary Tail Failure Mode} \\
\midrule
\textbf{\Darmok{} (Live Contest)} & 2026-10-10 & $\mathbf{7/7}$ ($\mathbf{100\%}$) & $\mathbf{210/210}$ & $\mathbf{05{:}08}$ ($01{:}32$) & Survived 6-min Final brownout \\
\texttt{gemini-4-argon} (32k/65k) & 2026-10-13 & $15/21$ ($71.4\%$) & $450/630$ & $04{:}33$ ($01{:}14$) & $6/21$ 32k token-cap loops ($7/9$ at 65k) \\
\texttt{gpt-6-astra} (1-Shot)     & 2026-09-12 & $21/21$ ($100\%$) & $630/630$ & $04{:}32$ ($01{:}38$) & None on $15\times 15$ draws ($0/21$ failures) \\
\texttt{gpt-5.6-sol} (1-Shot)     & 2026-07-09 & $16/21$ ($76.2\%$) & $509/630$ & $08{:}54$ ($06{:}52$) & $3$ timeouts ($30{:}00$), $1$ HTTP 520 \\
\texttt{claude-fable-5} (1-Shot)  & 2026-07-06 & $19/21$ ($90.5\%$) & $570/630$ & $14{:}54$ ($04{:}46$) & $2/21$ safety refusals ($0/30$ clues) \\
\bottomrule
\end{tabular}
\end{adjustbox}
\end{table}

Immediately after the championship, we ran a controlled ablation program to evaluate each component in isolation.
At first glance, four results appeared to confirm the Rube Goldberg critique.
\begin{enumerate}[leftmargin=1.8em,itemsep=2pt]
    \item \textbf{Single-shot frontier models have reached accuracy parity on standard blocked grids (\Cref{tab:intro-oneshot-mini}, \cref{sec:tail-eval}).}
    Handed an entire 30-clue championship puzzle in one prompt, standalone \texttt{gpt-6-astra} solved $21/21$ draws perfectly ($630/630$ clues)\footnote{An observed $0/21$ failure rate corresponds to a one-sided $95\%$ Clopper--Pearson binomial upper bound of $\le 13.3\%$ across $21$ draws ($\le 34.8\%$ across $7$ distinct puzzles, since $3$ draws on $7$ selected puzzles are not $21$ independent draws from the puzzle population).
    Because that same $7$-puzzle small-sample bound applies symmetrically to \Darmok{}'s $7/7$ championship sweep, our primary reliability estimate rests on the 43-day Epoch~E2 daily record ($41/43 = 95.3\%$ perfect grids, $4.7\%$ observed failure rate, $95\%$ Clopper--Pearson upper bound $\le 13.9\%$; \cref{sec:tail-eval}).} in a median of $4\text{m}\,32\text{s}$ ($\text{SD } 1\text{m}\,38\text{s}$), and during the live Championship Final its background feed wrote a complete $30/30$ JSON file at $5\text{m}\,38\text{s}$ ($74.0\text{s}$ before the per-clue pipeline finished cross-verifying the grid at $6\text{m}\,52\text{s}$).
    Meanwhile, \texttt{gemini-4-argon} solved $15/15$ non-truncated draws ($7/9$ on the three hardest puzzles at a 65k-token cap), \texttt{gpt-5.6-sol} solved $16/21$, and \texttt{claude-fable-5} solved $19/21$.
    \item \textbf{Prompt scaffolding shows no detectable accuracy benefit over native extended thinking (\cref{sec:prompt-ablation}, \Cref{app:prompt-p0}).}
    Each per-clue prompt requests a structured JSON output containing the definition, wordplay type, step-by-step explanation, draft answer, confidence score, and final answer (\Cref{app:prompt-p0}).
    Across all $463$ test-set clues evaluated over seven prompt variants on two models ($463 \times 7 \times 2 = 6{,}482$ scored calls), reordering the answer before the explanation, removing self-revision, removing decomposition fields, or stripping domain solving instructions changed accuracy by less than same-prompt replicate noise once a JSON parser sensitivity on Claude was accounted for uniformly across arms.
    \item \textbf{Eliciting verbalized confidence does not perturb object-level answers (\cref{sec:conf-ablation}).}
    Across $463$ clues evaluated over three confidence-schema variants on two models ($463 \times 3 \times 2 = 2{,}778$ calls), removing the confidence fields left the emitted crossword answers identical on $98.4\%$--$98.7\%$ of jointly completed calls on Gemini (whose same-prompt cross-session repeat identity is $99.7\%$) and $95.9\%$--$96.8\%$ on Claude (same-prompt identity $96.7\%$).
    \item \textbf{Multi-model per-clue seats evaluate to zero marginal accuracy gain in the test-set mean (\cref{sec:mean-eval}).}
    In an 80-run leave-one-out stage ablation on the 16-puzzle test set, removing the historical-answer lookup, the whole-grid feed, or corroboration admission cost one clue each (inside replica noise), and removing both secondary per-clue seats (Claude and GPT-5) to run the opening cold pass and crossing-constrained retry passes on the Gemini anchor alone (with the background feed retained) scored $461/463$---matching the full three-seat ensemble in accuracy (while running $\approx 1\text{m}$ slower in diagnostic median solve time).
\end{enumerate}

Taken together, these findings forced us to answer a central question:
\textit{If frontier models can solve championship cryptics in one shot, and a single per-clue seat with a crossing-retry loop matches the full ensemble's accuracy on the test set, why did we build the rest of the system?}

\subsection{Resolution: Failover Redundancy for the Tail, Not Serial Complexity}
\label{sec:resolution}

The resolution lies in the difference between a Rube Goldberg contraption and a resilient distributed system~\cite{vonneumann1956probabilistic}.
A Rube Goldberg machine chains components \textit{in series}, so the failure of any link breaks the whole.
\Darmok{}'s multi-seat orchestration and fallback stages are arranged \textit{in parallel across pass barriers}, combining $N$-version model diversity~\cite{avizienis1985nversion} with active-active replication~\cite{lyons1962tmr,schneider1990implementing} to bound tail risk when reasoning APIs degrade.
In computer architecture and distributed systems, redundancy mechanisms such as error-correcting memory~\cite{hamming1950error}, RAID arrays~\cite{patterson1988raid}, hedged RPCs~\cite{dean2013tail}, and active-active failover replicas~\cite{schneider1990implementing} contribute zero marginal accuracy during fault-free operation; they exist to preserve correctness and latency bounds when primary components stall.

Our telemetry from the Championship Final, our 132-day daily record, and our 135-draw single-shot battery (\Cref{fig:intro-pareto}b) shows that frontier reasoning endpoints are heavy-tailed components whose failure modes are invisible on calm static benchmarks (\cref{sec:tail-eval}).
Across the six qualifying and semi-final puzzles, our Gemini anchor delivered $115$--$119\text{ tokens/s}$ and timed out on only $1$--$8$ clues per puzzle.
Seconds after the Championship Final began, however, the provider experienced an unexpected server-side latency brownout: project-level Google Cloud \texttt{serviceruntime} median request latency jumped from $16\text{s}$ to $50\text{s}$--$190\text{s}$ and effective token delivery collapsed by $72\%$ to $33.0\text{ tokens/s}$, pushing the anchor into its $90\text{s}$ timeout cap on \textbf{$23$ of the $30$ opening-pass clues} (and failing $17$ of $23$ retry calls: $16$ timeouts and $1$ parse error).
By the time the winning $30/30$ grid locked at $6\text{m}\,52\text{s}$, the secondary per-clue seats had supplied $13$ answers without the anchor and the asynchronous \texttt{gpt-6-astra} boundary feed had supplied the remaining $4$ ($17/30 = 56.7\%$ total), while the per-clue seats independently cross-verified $26/30$ entries and $13$ of the $17$ checked entry positions ($13$ of $15$ unique checked physical cells) of the $4$ feed entries.
The 132-day production tail confirms the same failover mechanism: in the worst decile of daily runs by anchor opening-pass silence ($13$ days with $42.9\%$--$100\%$ silence),\footnote{These $13$ worst-decile days comprise $9$ early Epoch~E0 runs where local macOS DNS exhaustion (\texttt{Errno 8}) during 60-socket bursts dropped $100\%$ of Gemini's opening-pass calls ($258/262 = 98.47\%$, $5/9$ perfect grids), $1$ partial-error E0 run ($29/30$, $0/1$ perfect grids), and $3$ Epoch~E1/E2 provider brownout days ($86/86 = 100\%$, $3/3$ perfect grids; \Cref{tab:app-tail}).} overall clue accuracy held at $373/378$ ($98.68\%$ vs.\ $99.23\%$ on the other $119$ days; $8/13 = 61.5\%$ perfect grids overall, with all $5$ imperfect grids occurring in Epoch~E0 prior to corroboration admission) because the secondary seats supplied $32.1\%$ (up to $89.8\%$ across the 2-day peak outage, or $90.0\%$ [$27/30$] single-day peak) of committed answers.

Standalone single-shot frontier models are similarly exposed to tail failures (\Cref{fig:intro-pareto}b, \Cref{tab:intro-oneshot-mini}, \cref{sec:tail-oneshot}).
While September--October 2026 frontier models (\texttt{gemini-4-argon} and \texttt{gpt-6-astra}) compress completed-solve latency below the $10\text{m}\,05\text{s}$ human Final winner benchmark relative to the wide, right-skewed latency tails of the July 2026 cohort (\texttt{gpt-5.6-sol} and \texttt{claude-fable-5}), every standalone family exhibits a distinct tail hazard: \texttt{gemini-4-argon} exhausted its default 32k-token budget inside repetitive verification loops on $6/21$ draws ($0/30$ clues, including $1$ on the Final); \texttt{gpt-6-astra} failed to return before the per-clue solver on $17$ of $20$ barred \textit{Mephisto} runs; \texttt{gpt-5.6-sol} suffered $4/21$ timeouts or HTTP~520 errors plus $1$ partial grid ($29/30$, on the Final); and \texttt{claude-fable-5} suffered $2/21$ safety-classifier refusals ($0/30$ clues, including $1$ on the Final) while taking $7\text{m}\,02\text{s}$ to $22\text{m}\,31\text{s}$ on completed draws.

\subsection{Contributions and Paper Roadmap}

This paper makes three main contributions.
\begin{enumerate}[leftmargin=1.8em,itemsep=2pt]
    \item \textbf{Algorithmic (\cref{sec:arch}).}
    We formulate a synchronous, order-independent pass-barrier admission recurrence (\textit{rebuild-from-kept}) that scores candidate proposals using calibrated seat floors, multi-seat corroboration, and a Jacobi-style crossing support update over prior-pass standings, paired with wholesale grid reconstruction.
    We show that iterative crossing-letter retries ($+13$ clues on the test set) and the mutual backup between corroboration admission and local terminal dictionary sweeps ($+99$ clues on a $1{,}204$-clue replay) form the algorithmic core of full-grid solving.
    \item \textbf{Empirical Evaluation of Common LLM Prompting Choices (\cref{sec:folklore}).}
    Across $9{,}260$ paired per-clue API calls plus an audit of a published symbolic Python-AST wordplay verification engine, we show that for native extended-thinking models: (a) intersecting letter patterns induce a steep $+33$ to $+50\text{ percentage-point}$ solve-rate jump at $\ge 3$ known letters; (b) chain-of-thought field ordering, self-revision, decomposition fields, and domain tutorials produce no detectable accuracy gain, whereas requiring a wordplay explanation acts as a brake on unverified guessing (converting would-be wrong guesses into timeouts); (c) eliciting self-graded confidence leaves emitted answers unchanged on $95.9\%$--$98.7\%$ of jointly completed calls while revealing a sharp calibration knee on Claude ($\ge 85$) versus binary confidence on Gemini; and (d) without its candidate-frequency fallback, a published symbolic-LLM wordplay verifier solves half as many clues as a zero-cost proposal-frequency baseline on its 200-clue validation script ($14.0\%$ vs.\ $28.0\%$).
    \item \textbf{Systems Engineering for Heavy-Tailed Reasoning APIs (\cref{sec:eval,sec:discussion}).}
    Combining our 132-day daily record, 16-puzzle test-set ablations, 10-puzzle \textit{Mephisto} transfer study, 135-draw single-shot frontier battery, and dual-sided Championship Final telemetry, we quantify how static benchmark ablations misprice resilience over volatile reasoning endpoints.
\end{enumerate}

\Cref{sec:related} introduces the dual-grammar structure of cryptic clues, formalizes 2D crossword grid topologies, and reviews related work.
\Cref{sec:arch} defines our operational terminology, presents the \Darmok{} architecture and synchronous corroboration recurrence, and details per-seat calibration and triple-replica execution.
\Cref{sec:folklore} reports our controlled clue-level experiments on crossing constraints, prompt structure, confidence elicitation, and symbolic verification.
\Cref{sec:eval} evaluates the full system across our daily, test-set, and \textit{Mephisto} corpora, contrasting average-case ablations against tail-regime performance in the Championship Final.
\Cref{sec:discussion} discusses implications for compound AI evaluation, \Cref{sec:future-work} outlines open frontiers including a formal cryptic grammar and automated clue setting, and \Cref{sec:conclusion} concludes.
Complete reference tables, the production prompt, and verified wordplay derivations for all cited clues appear in \Cref{app:tables,app:prompts-clues}.

\section{Domain Primer and Related Work}
\label{sec:related}

\begin{figure}[!htbp]
\centering
\includegraphics[width=\textwidth]{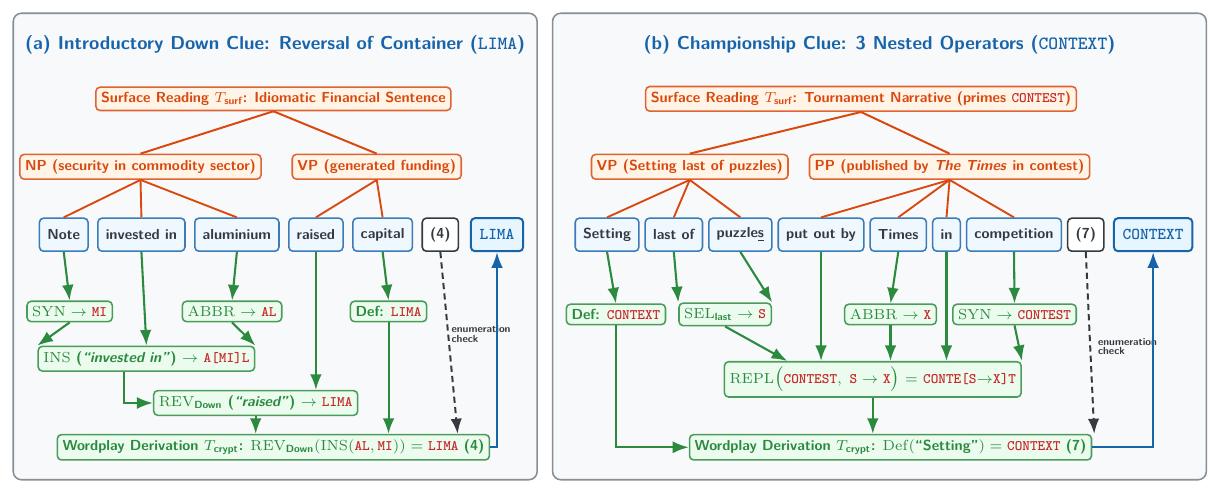}
\caption{\textbf{Dual-grammar anatomy of a cryptic clue.}
Competing surface readings $T_{\text{surf}}$ (top, orange) and compositional wordplay derivations $T_{\text{crypt}}$ (bottom, green) over the same token sequence for \textbf{(a)} an introductory Down clue (\ans{LIMA}; Clue~C1, \Cref{tab:app-clues-1}) and \textbf{(b)} a 2026 Championship clue (\ans{CONTEXT}, Semi-Final~1, clue~28A; Clue~C3, \Cref{tab:app-clues-1}).}
\label{fig:primer-clues}
\end{figure}

To make our algorithmic and systems contributions self-contained for readers outside the (admittedly niche) cryptic crossword community, we first formalize the dual-grammar linguistic structure of cryptic clues (\Cref{fig:primer-clues}) and the geometry of 2D crossword grids (\Cref{fig:primer-grids}), and then review prior work across American and cryptic crossword solving.

\subsection{Anatomy of a Cryptic Clue: Dual Parse Trees and Compositional Primitives}
\label{sec:primer-clues}

In an American-style crossword, a clue is a direct semantic or encyclopedic query whose length is inferred from the grid (e.g., \clue{Capital of Peru} for a $4$-letter slot maps directly to \ans{LIMA});
in British crosswords, the clue instead appends an explicit parenthesized \textit{enumeration} such as (4) that specifies the exact letter count and word structure of the answer (e.g., $(7)$ for a single 7-letter word, $(5,5)$ for a two-word phrase, or $(4\text{-}6)$ for a hyphenated entry).
For the exact same Down entry \ans{LIMA}, a British cryptic setter might instead write:\footnote{In reality, this clue is the creation of the author.}
\begin{center}
\clue{Note invested in aluminium\footnote{Spelled the British way, naturally: an American setter trying to hide \ans{MI} inside \textit{aluminum} would still get \ans{AL}, but neither a British setter nor a British solver would ever forgive the missing \textit{i}.} raised capital (4)} $\longrightarrow$ \ans{LIMA} \quad (\text{Clue~C1, \Cref{tab:app-clues-1}}).
\end{center}
As illustrated in \Cref{fig:primer-clues}a, the setter constructs a single natural-language phrase that simultaneously supports two competing interpretations.
\begin{enumerate}[leftmargin=1.8em,itemsep=2pt]
    \item \textbf{The Surface Reading ($T_{\text{surf}}$, Standard English).}
    Viewed as ordinary prose, the clue parses into an idiomatic English sentence in which words bind according to standard syntactic structure, activating a coherent semantic frame (here, a promissory note invested in the aluminium market that raised financial capital).
    Every word choice, including capitalization, punctuation, and polysemous parts of speech (such as reading a passive participle as an active past-tense verb), is engineered to misdirect direct semantic retrieval.
    \item \textbf{The Cryptic Parse ($T_{\text{crypt}}$, Compositional Wordplay Abstract Syntax Tree [AST]).}
    Viewed as a formal specification under the rules of cryptic grammar~\cite{ximenes1966art,manley2014chambers}, the exact same token sequence partitions at an unmarked boundary into two independent sub-programs that must evaluate to the same uppercase letter string of the enumerated length ($4$):
    \begin{itemize}[leftmargin=1.5em,itemsep=1pt]
        \item \textit{Unmarked Definition.}
        Located almost always at the leftmost or rightmost extremity of the clue (here, the trailing noun \textit{``capital''}, defining the capital of Peru), specifying the semantic category or synonym class of the target answer.
        \item \textit{Compositional Wordplay AST.}
        A typed expression tree built from string-manipulation operators signaled by natural-language keywords (\textit{indicators}) acting on literal, abbreviated, or synonymous operands (\textit{fodder})~\cite{clueclinic2026}.
        In \Cref{fig:primer-clues}a, \textit{``Note''} maps via standard musical notation to the third sol-fa syllable \ans{MI} ($\operatorname{ABBR}$), \textit{``aluminium''} maps to its chemical symbol \ans{AL} ($\operatorname{ABBR}$), the container indicator \textit{``invested in''} inserts \ans{MI} into \ans{AL} ($\ans{A[\textbf{MI}]L}$), and the directional reversal indicator \textit{``raised''} (licensed in a Down clue) reverses \ans{AMIL} to \ans{LIMA}.
    \end{itemize}
\end{enumerate}

Following Manley's standard taxonomy~\cite{manley2014chambers} and Ximenes' fairness principles~\cite{ximenes1966art}, the operator repertoire of cryptic wordplay comprises the following.
The indicator lists for each kind of operator, shown italicized within parentheses, are representative rather than exhaustive. 
More complete lists are readily available online~\cite{clueclinic2026}.
\begin{itemize}[leftmargin=1.8em,itemsep=1pt]
    \item \textbf{Charade ($\operatorname{CHAR}(E_1, \dots, E_m)$).}
    Concatenates component strings $E_1, \dots, E_m$ end-to-end in sequence.
    \item \textbf{Anagram ($\operatorname{ANA}(F, \text{ind}_{\text{ana}})$).}
    Permutes the multiset of letters in explicit fodder $F$, signaled by an indicator of disorder or motion (\textit{new, worried, jerks, entangled}).
    Under strict Ximenean fairness~\cite{ximenes1966art}, the letters to be anagrammed ($\mathsf{Fodder}$) must appear explicitly in the clue or via a direct standard abbreviation, disallowing \textit{indirect anagrams} (anagramming an unstated synonym).
    \item \textbf{Container / Insertion ($\operatorname{INS}(E_{\text{out}}, E_{\text{in}}, \text{ind}_{\text{ins}})$).}
    Embeds string $E_{\text{in}}$ inside $E_{\text{out}}$ (\textit{invested in, takes in, bandages, gorges}).
    \item \textbf{Deletion and Letter Selection ($\operatorname{DEL}_{\text{mode}}(E), \operatorname{SEL}_{\text{pos}}(\langle\text{Lit}\rangle)$).}
    Strips initial (\textit{headless}), final (\textit{curtailment}, \textit{shortened}), outer (\textit{peeled}), or internal (\textit{heartless}, \textit{extremely}) letters, or extracts acrostic/initial (\textit{heads for}), terminal (\textit{last of, close to}), or alternating characters (\textit{regularly}) from literal clue words.
    \item \textbf{Reversal ($\operatorname{REV}_d(E)$) and Cycling ($\operatorname{CYC}_k(E)$).}
    Reverses character order (\textit{returned, knocked back}, or \textit{raised, heading north, climbing} in a Down clue) or cyclically shifts letters (\textit{cycling}).
    \item \textbf{Substitution ($\operatorname{REPL}(E, F_x \to F_y)$) and Subtraction ($\operatorname{MINUS}(E_1, F_2)$).}
    Replaces $F_x$ with $F_y$ inside $E$ (\textit{put out by, changing X for Y}), or deletes the letters of explicit fodder $F_2$ from $E_1$ (\textit{leaving, losing}).
    \item \textbf{Hidden Words ($\operatorname{HID}$), Homophones ($\operatorname{HOM}$), and Spoonerisms ($\operatorname{SPOON}$).}
    Conceals the answer as a proper internal substring of literal clue tokens (\textit{in, fragment of}), maps an expression to a phonetic equivalent (\textit{reportedly, picked up}), or swaps the initial phonetic consonant clusters of a phrase (\textit{Spooner's}).
    \item \textbf{Foreign/Dialect Synonyms ($\operatorname{FSYN}$), Archaisms ($\operatorname{ARCH}$), Cross-References ($\operatorname{XREF}$), Double Definitions, and ``\&lit''.}
    Substitutes foreign-language (French, Latin, German, etc.) or regional/Scots equivalents signaled by provenance indicators (\textit{in France, locally, of old, over the border}), references another clue entry ($n\text{A}/n\text{D}$), pairs two distinct definitions with no wordplay, or constructs an all-in-one clue (``\&lit'') in which the full clue span serves simultaneously as definition and wordplay AST.
\end{itemize}

Championship clues routinely \textbf{compose several primitives into a nested expression tree}, as illustrated in \Cref{fig:primer-clues}b on 2026 Championship Semi-Final~1 clue~28A (\ans{CONTEXT}; Clue~C3, \Cref{tab:app-clues-1}).

\subsection{Grid Topologies and the Breakdown of Static Constraint Satisfaction}
\label{sec:csp-breakdown}

\begin{figure}[!htbp]
\centering
\includegraphics[width=\textwidth]{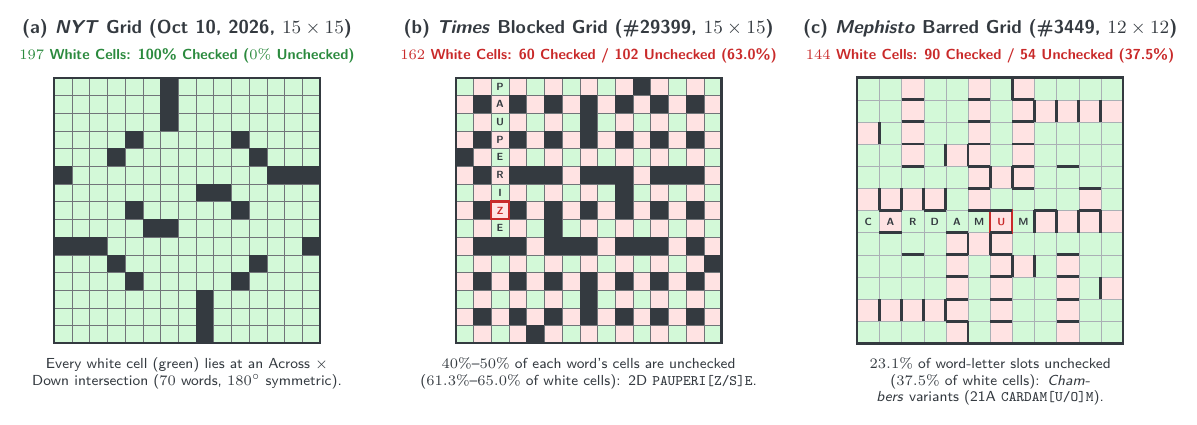}
\caption{\textbf{Grid checking topologies on real symmetric crossword grids.}
Checked intersection cells (green) versus unchecked cells (pink) on \textbf{(a)} a fully checked $15\times 15$ American grid ($0\%$ unchecked), \textbf{(b)} a British $15\times 15$ blocked grid (\textit{Times}~\#29399, $45.9\%$ of word-letter positions / $63.0\%$ of white cells unchecked; 2D \ans{PAUPERI[Z/S]E}, Clue~C5), and \textbf{(c)} a Sunday \textit{Mephisto} $12\times 12$ barred grid (\#3449, $23.1\%$ of word-letter positions / $37.5\%$ of white cells unchecked; 21A \ans{CARDAM[U/O]M}, Clue~C25).}
\label{fig:primer-grids}
\end{figure}

Formally, a standard crossword grid is a finite, non-toroidal 2D rectangular grid $\{1, \dots, R\} \times \{1, \dots, C\}$\footnote{Almost always square; $R = C = 15$ for daily \textit{Times} grids and $R = C = 12$ for \textit{Mephisto}.} containing a subset of playable white cells $\mathcal{V} \subseteq \{1, \dots, R\} \times \{1, \dots, C\}$ and a set of $m$ clues $\mathcal{I} = \{1, \dots, m\}$.
The pattern of white cells must possess either 180-degree or 90-degree rotational symmetry.
Each clue $i \in \mathcal{I}$ has a starting cell $c_{i,1} = (x_i, y_i) \in \mathcal{V}$ (at row $x_i$, column $y_i$), an orientation $d_i \in \{\text{Across}, \text{Down}\}$, and a total letter length $\ell_i \in \mathbb{N}^+$, spanning the contiguous sequence of white cells $\mathbf{c}_i = [c_{i,1}, c_{i,2}, \dots, c_{i,\ell_i}] \in \mathcal{V}^{\ell_i}$ given by the step recurrence
\begin{equation}
c_{i,k} = c_{i,k-1} + \Delta(d_i) \quad (2 \le k \le \ell_i), \quad \text{where } \Delta(\text{Across}) = (0, 1), \;\; \Delta(\text{Down}) = (1, 0).
\end{equation}
Let $\Sigma = \{\ans{A}, \dots, \ans{Z}\}$ denote the uppercase English alphabet.
After an orthographic normalization step that folds characters to uppercase ASCII and strips spaces, hyphens, apostrophes, punctuation, and diacritics, \textit{assigning} a candidate word $a_i \in \Sigma^{\ell_i}$ to clue $i$ places character $a_i[k] \in \Sigma$ into cell $c_{i,k}$.
Whenever an Across clue $i$ and a Down clue $j$ intersect at a shared physical square, say the $p_{ij}$-th letter of clue $i$ and the $p_{ji}$-th letter of clue $j$ ($c_{i,p_{ij}} = c_{j,p_{ji}}$), they enforce the binary letter-equality constraint $a_i[p_{ij}] = a_j[p_{ji}]$.

In American-style crosswords (\Cref{fig:primer-grids}a), solver design rests on two domain properties.
\begin{enumerate}[leftmargin=1.8em,itemsep=2pt]
    \item \textbf{Total grid checking ($0\%$ unchecked cells).}
    Every white cell $v \in \mathcal{V}$ belongs to both an Across clue and a Down clue, so every character position in every word is independently verified by an orthogonal clue.
    \item \textbf{High-recall cold candidate retrieval (a property of direct clues).}
    Because American clues query synonyms, trivia, or phrase completions rather than nested character algebra, candidate retrievers combining historical clue--answer databases and neural bi-encoders routinely achieve top-$10^3$ recall of $95\%$--$99\%$ before examining the grid~\cite{littman2002proverb,ginsberg2011drfill,wallace2022berkeley}.
\end{enumerate}
Landmark American solvers such as \textsc{Proverb}~\cite{shazeer1999aaai,littman2002proverb}, \textsc{WebCrow}~\cite{ernandes2005webcrow}, \textsc{Dr.Fill}~\cite{ginsberg2011drfill}, and the Berkeley Crossword Solver~\cite{wallace2022berkeley}, together with constraint-programming solvers~\cite{beacham2001cp,anbulagan2008cp} and Monte Carlo tree search systems~\cite{mcts2022icaps,mcts2024aij}, populate each clue up front with a static list of $10^2$--$10^4$ candidates and calibrated log-probabilities, reducing grid completion to maximum-weight assignment via loopy belief propagation~\cite{murphy1999loopy,littman2002proverb} or integer linear programming~\cite{ginsberg2011drfill}, analogous to championship Scrabble engines such as \textsc{Maven}~\cite{sheppard2002scrabble}.

British cryptic crosswords violate both properties simultaneously.
\begin{enumerate}[leftmargin=1.8em,itemsep=2pt]
    \item \textbf{Unchecked cells ($40\%$--$50\%$ of word-letter positions on standard blocked grids; $20\%$--$23\%$ on barred \textit{Mephisto} grids).}
Standard British blocked grids (\Cref{fig:primer-grids}b) do not provide full crosschecking, unlike US-style crosswords. 
In \textit{The Times}, each entry has unchecked cells, almost always in alternating positions; for an entry of $2n$ or $2n+1$ cells, there may be at most $n$ unchecked cells. 
The rules for barred \textit{Mephisto} grids are different: an entry of $n$ cells should have no more than $\lfloor n/3 \rfloor$ unchecked cells. 
This is because the vocabulary is typically more obscure than that found in standard cryptics.

    Across the $30$ standard $15\times 15$ \textit{Times} grids in our repository, $40.0\%$ to $50.0\%$ of the letter positions in every individual word (mean $43.7\%$; $61.3\%$ to $65.0\%$ of physical white squares, median $63.4\%$) are \textit{unchecked} (pink in \Cref{fig:primer-grids}b).
    In $12\times 12$ barred \textit{Mephisto} grids (\Cref{fig:primer-grids}c, \#3449), black squares are omitted and $36$ words are packed into $144$ white cells separated by thick bars, leaving $20.0\%$ to $23.1\%$ of all word-letter positions (mean $21.4\%$; $33.3\%$ to $37.5\%$ of physical white cells, mean $35.3\%$) unchecked.
    While barred \textit{Mephisto} grids check a higher fraction of word-letter positions than blocked \textit{Times} grids ($78.6\%$ vs.\ $56.3\%$ mean), their difficulty arises from obscure \textit{Chambers Dictionary} headwords and spelling variants that differ on a single unchecked square (such as \ans{CARDAMUM}/\ans{CARDAMOM} or \ans{RIEVE}/\ans{REIVE}, and British \ans{-ISE}/\ans{-IZE} endings such as \ans{PAUPERISE}/\ans{PAUPERIZE}; Clues~C5, C23, and C25).
    On any unchecked cell, perpendicular crossings provide zero constraint; the correct letter can be determined only by using the clue's wordplay AST to establish which spelling is required.
    \item \textbf{Rapid tail-off of cold candidate lists.}
    Because surface readings actively misdirect retrieval and wordplay expressions compose multiple string operations, prompting a language model for top-$K$ candidates without crossing letters yields rapid probability decay after the top $1$--$2$ hypotheses and surfaces plausible definition-only guesses that match the enumeration $\ell_i$ while violating the wordplay. We observed this pattern directly in our early multi-candidate pass@$k$ trials ($k \in \{1,2,4,6,8,10\}$ across $18$ V1-missed clues and $3$ models, $324$ runs; \cref{sec:verifier-falsification}), where $96$ of the $99$ correct hits ($97.0\%$) occurred at rank~1 ($89$) or rank~2 ($7$) and ranks~3--10 combined yielded only $3$ correct answers across $1{,}296$ slots.
\end{enumerate}

\subsection{Related Work}
\label{sec:related-work}

Prior research on cryptic crosswords has focused primarily on isolated clue-level parsing rather than full-grid tournament solving.
Early rule-based systems such as \textsc{Lacross}~\cite{woodhead1977computer,williams1979computer} and Tunstall-Pedoe's commercial solver \textsc{Crossword Maestro}~\cite{tunstallpedoe2010true} demonstrated that explicit lexicon lookup and hand-engineered grammar rules could decompose single-step anagrams, charades, and containers, while Hardcastle~\cite{hardcastle2007thesis,hardcastle2007w07} formalized the dual-constraint structure of cryptic clue generation.
In the modern NLP literature, Efrat et al.~\cite{efrat2021cryptonite} introduced \textsc{Cryptonite} ($470{,}000+$ clues from \textit{The Times} and \textit{The Telegraph}) and Rozner et al.~\cite{rozner2021decrypting} introduced wordplay-disjoint splits from \textit{The Guardian}, establishing that standard sequence-to-sequence models struggled to generalize across unseen wordplay compositions.
More recent studies evaluated foundation models and prompting strategies on isolated cryptic clues~\cite{sadallah2024good,sadallah2025challenging,saha2025lmcrossword} and explored LLM-synthesized Python verification scripts~\cite{andrews2024proving,andrews2025reasoning} (which we audit experimentally in \cref{sec:verifier-falsification}).
In statistical physics, Hartmann~\cite{hartmann2024percolation} recently analyzed crossword-puzzle percolation on random lattices, showing that cooperative crossing constraints induce a percolation transition in solved cluster growth as single-word solve probability increases.
To our knowledge, \Darmok{} is the first automated system to solve full-grid British cryptic crosswords at championship human speed and zero-error reliability.

\section{Architecture: Order-Independent Constraint Orchestration}
\label{sec:arch}

\begin{figure}[!htbp]
\centering
\includegraphics[width=\textwidth]{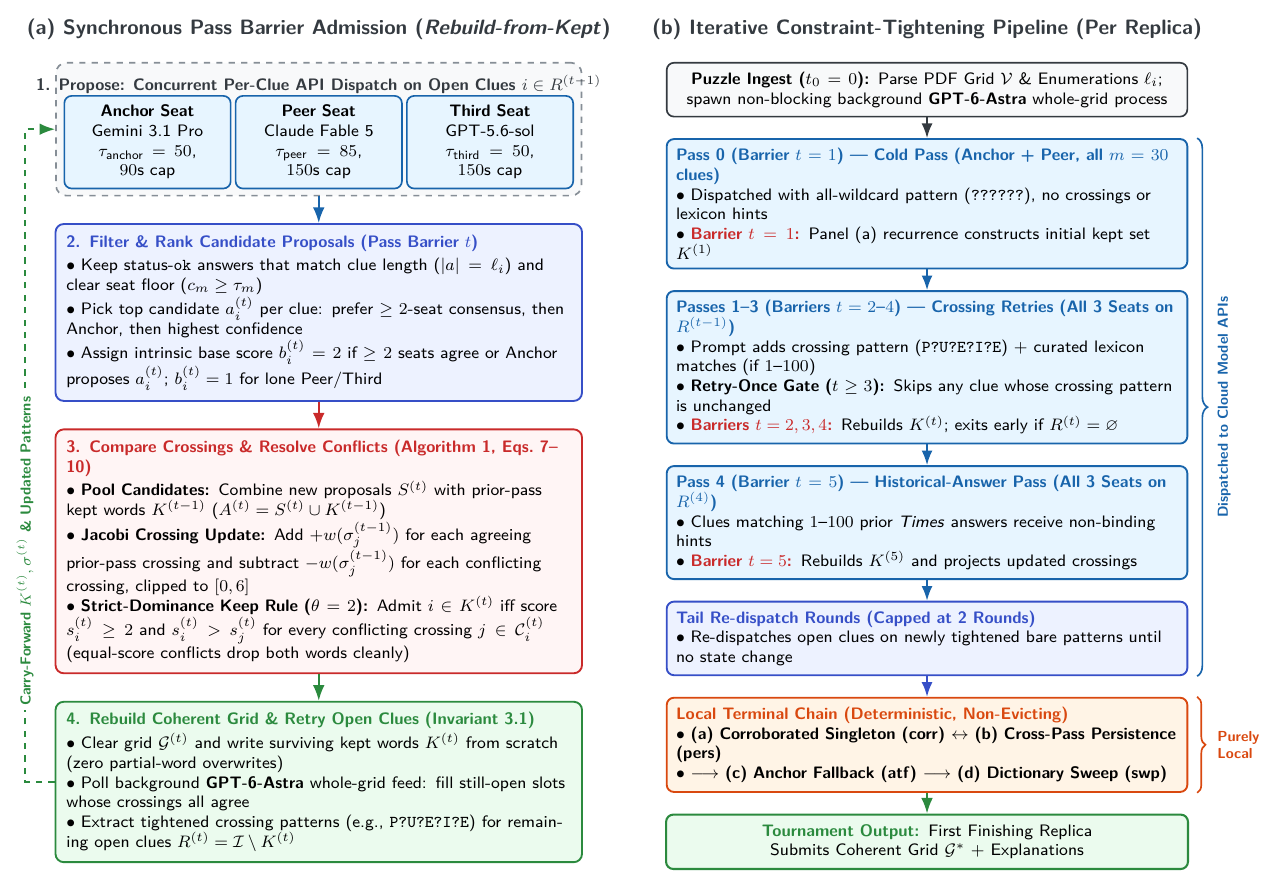}
\caption{\textbf{Architecture of the \Darmok{} system.}
\textbf{(a)} Synchronous pass-barrier admission (\textit{rebuild-from-kept}, \Cref{alg:corrob}), which scores qualifying per-clue proposals against the carry-forward kept set $K^{(t-1)}$ alongside the non-evicting \texttt{gpt-6-astra} feed and reconstructs the coherent grid $\mathcal{G}^{(t)}$ from scratch (\Cref{inv:coherence}).
\textbf{(b)} Iterative constraint-tightening pipeline executed per puzzle replica across cloud passes (Pass~0 through Tail Rounds) and the local terminal chain (\textsc{corr} $\leftrightarrow$ \textsc{pers} $\to$ \textsc{atf} $\to$ \textsc{swp}).}
\label{fig:arch}
\end{figure}

\Darmok{} treats full-grid cryptic crossword solving as an \textit{iterative, synchronous constraint-tightening loop}, analogous to classical arc-consistency propagation~\cite{mackworth1977consistency}, executed over asynchronous, heavy-tailed reasoning models (\Cref{fig:arch}).
Because prompting a language model to emit hundreds of cold candidates per clue fails on compositional cryptic wordplay, the orchestrator at each pass instead dispatches all currently unsolved (\textit{residual open}) clues in parallel across distinct reasoning model seats~\cite{avizienis1985nversion}; it admits only those proposals that survive a synchronous corroboration and crossing-consistency recurrence at the pass barrier, projects the resulting kept set onto the grid to induce tighter wildcard patterns (e.g., \texttt{P?U?E?I?E}), and re-dispatches the remaining open clues conditioned on those patterns in the next pass.

\paragraph{Operational Terminology and Post-Hoc Grading Metrics.}
To distinguish runtime solver concepts from post-hoc benchmark evaluation metrics across \cref{sec:arch,sec:folklore,sec:eval}, we separate our terminology into two groups (with mathematical symbols summarized in \Cref{tab:notation}).
At \textbf{runtime}, the orchestrator operates without access to ground truth (GT) using four concepts:
\begin{itemize}[leftmargin=1.8em,itemsep=1pt]
    \item \textbf{Model Seat ($m \in \mathcal{M}$).}
    One of the three per-clue reasoning model API endpoints queried at each pass: \textbf{Anchor} (\texttt{gemini-3.1-pro-preview}), \textbf{Peer} (\texttt{claude-fable-5}), and \textbf{Third Seat} (\texttt{gpt-5.6-sol}).
    \item \textbf{Qualifying Proposal.}
    On clue $i$, a response $r_i = (m, a, c)$ from seat $m$ with answer string $a$ and self-reported confidence $c$ is \textit{qualifying} if its API status is \texttt{ok}, its normalized letter count matches the clue's enumeration ($\vert{}a\vert{} = \ell_i$), and its self-reported confidence clears the seat's floor ($c \ge \tau_m$).
    \item \textbf{Pass Barrier, Kept Set ($K^{(t)}$), and Residual Open Set ($R^{(t)}$).}
    A \textit{pass barrier} $t \in \{1, \dots, T\}$ is the synchronization point at the end of dispatch pass $t$ after all in-flight per-clue API requests have returned or hit their timeout caps.
    At barrier $t$, the admission recurrence selects a conflict-free (\textit{coherent}) subset of clues $K^{(t)} \subseteq \mathcal{I}$ whose answers agree on every shared intersection cell and writes them to the physical grid $\mathcal{G}^{(t)}$; $R^{(t)} = \mathcal{I} \setminus K^{(t)}$ is the set of remaining unsolved clues.
    \item \textbf{Boundary Feed (\textsc{bf}).}
    The asynchronous background process running \texttt{gpt-6-astra} on the full puzzle in a single prompt, polled only at pass barriers to fill open clues in $R^{(t)}$ without evicting kept entries in $K^{(t)}$.
\end{itemize}
During \textbf{post-hoc evaluation} against known puzzle solutions ($\text{GT}$), we report three grading metrics:
\begin{itemize}[leftmargin=1.8em,itemsep=1pt]
    \item \textbf{Admissible Answer.}
    In clue-level evaluations (\cref{sec:folklore}), a qualifying response that also matches the ground-truth solution ($a = \text{GT}$) is termed \textit{admissible}.
    \item \textbf{Silent Rate and Jointly Completed (\texttt{both-ok}) Subset.}
    A seat's \textit{silent rate} on a pass is the fraction of dispatched clues that fail to return a valid JSON response due to timeouts, socket/API errors, or truncated outputs ($1 - \text{Avail}$).
    When comparing two prompt arms on the same clue set, the \textit{jointly completed} (\texttt{both-ok}) subset comprises clues where both arms returned status \texttt{ok}.
    \item \textbf{First-to-Correct and Unique Solve.}
    Among clues committed correctly by model passes, a seat is credited with \textit{first-to-correct} if it returned the ground-truth answer at the earliest wall-clock timestamp, and with a \textit{unique solve} if no other seat ever proposed the ground-truth answer on any pass.
\end{itemize}

\begin{table}[!htbp]
\centering
\caption{Summary of mathematical notation used in the pass-barrier recurrence (\cref{sec:corrob-admission}) and pipeline (\cref{sec:pipeline}).}
\label{tab:notation}
\footnotesize
\begin{adjustbox}{max width=\textwidth}
\begin{tabular}{@{}llp{10.6cm}@{}}
\toprule
\textbf{Symbol} & \textbf{Domain / Value} & \textbf{Meaning} \\
\midrule
$\mathcal{I},\; \mathcal{V},\; \ell_i$ & $\{1,\dots,m\},\; \mathbb{N}^2,\; \mathbb{N}^+$ & Clue index set ($m \approx 28$--$36$), playable white cells, and normalized letter length of clue $i$. \\
$\mathcal{X}(i),\; p_{ij}$ & $\subseteq \mathcal{I} \setminus \{i\},\; \{1,\dots,\ell_i\}$ & Perpendicular clues intersecting clue $i$, and the letter index in clue $i$ shared with $j \in \mathcal{X}(i)$. \\
$\mathcal{M},\; \tau_m$ & $\{\text{anchor}, \text{peer}, \text{third}\},\; [0,100]$ & Per-clue model seats and confidence floors ($\tau_{\text{anchor}}=50, \tau_{\text{peer}}=85, \tau_{\text{third}}=50$). \\
$t,\; \mathcal{Q}_i^{(t)}$ & $\{1,\dots,T\},\; \subset \Sigma^{\ell_i}$ & Pass barrier index ($t=1$ for Pass~0) and distinct qualifying candidate strings for clue $i$ in pass $t$. \\
$\mathcal{M}_i^{(t)}(a)$ & $\subseteq \mathcal{M}$ & Subset of model seats that proposed qualifying candidate $a \in \mathcal{Q}_i^{(t)}$ for clue $i$ in pass $t$. \\
$a_i^{(t)},\; b_i^{(t)}$ & $\Sigma^{\ell_i},\; \{1, 2\}$ & Asserted candidate string and intrinsic base score ($b=2$ for Anchor or $\ge 2$-seat consensus; $b=1$ for lone Peer/Third). \\
$S^{(t)},\; K^{(t)},\; R^{(t)}$ & $\subseteq \mathcal{I}$ & Clues with fresh qualifying proposals in pass $t$ ($S^{(t)}$), kept clues admitted at barrier $t$ ($K^{(t)}$), and open residual ($R^{(t)} = \mathcal{I} \setminus K^{(t)}$). \\
$A^{(t)},\; \mathcal{X}^{(t)}(i),\; \mathcal{C}_i^{(t)}$ & $\subseteq \mathcal{I}$ & Asserting set $A^{(t)} = S^{(t)} \cup K^{(t-1)}$, active crossings $\mathcal{X}(i) \cap A^{(t)}$, and letter-conflicting active crossings. \\
$\sigma_i^{(t)},\; w(\sigma_i^{(t)})$ & $\{\StandingOut, \StandingTent, \StandingSupp, \StandingLock\},\; \{0,1,2,3\}$ & Discrete verification standing label of clue $i$ after barrier $t$ and its integer crossing-support weight. \\
$s_i^{(t)},\; \theta,\; \mathcal{G}^{(t)}$ & $[0, 6],\; \theta = 2,\; \mathcal{V} \to \Sigma \cup \{\varnothing\}$ & Clipped Jacobi corroboration score of clue $i$ at barrier $t$, admission threshold ($\theta=2$), and rebuilt grid state. \\
\bottomrule
\end{tabular}
\end{adjustbox}
\end{table}

\subsection{The Pathology of Greedy Incremental Admission}
\label{sec:greedy-pathology}

Our 2025 baseline solver (V1, which scored $380/463 = 82.1\%$ on our stratified test set) and our initial June 2026 Epoch~E0 prototype ($408/463 = 88.1\%$) relied on \textit{greedy incremental admission}: as each clue's API calls returned, if a proposal cleared a confidence threshold and agreed with whatever letters were currently written in the grid, it was immediately committed.
Later proposals that conflicted with an incumbent answer could overwrite it only by exceeding its cell-level confidence.
In practice, V1 was a genuine Rube Goldberg contraption~\cite{wolfe2000rube} of interacting heuristics and ad-hoc cell-override rules\footnote{The V1 override rules were patched as late as 04:00~BST on the morning of the October 2025 championship.} in which fixing one failure mode routinely destabilized another.

Forensic analysis across the $39$ daily runs of Epoch~E0 and our early development logs revealed that greedy incremental admission suffers from two fundamental flaws.
\begin{enumerate}[leftmargin=1.8em,itemsep=2pt]
    \item \textbf{Latency-driven arrival-order stochasticity and cascading error propagation.}
    Because commercial LLM response times vary widely from call to call, incremental admission is order-dependent on network and model thinking time and therefore inherently stochastic.
    Whichever clue's API call happens to return first writes its letters into the grid immediately; when a model makes a confident cold error that finishes quickly, those false crossing letters are injected into perpendicular clues and condition subsequent retries on false constraints, causing cascading abstentions or forced errors.
    \item \textbf{Partial-word overwrites and state divergence.}
    When conflict resolution operates at the granularity of individual grid squares rather than whole words, a later high-confidence crossing answer can overwrite shared intersections of an earlier word without cleanly evicting and re-solving the displaced entry.
    For example, on \textit{Times} Cryptic~\#29570 (June 16, 2026), both primary models cold-proposed the plausible anagram \ans{REPOSITED} for clue~21A (\clue{Ropes tied tight and put down again (9)}, ground truth \ans{REDEPOSIT}; Clue~C7, \Cref{tab:app-clues-1}).
    When perpendicular Down answers later committed at higher priority, cell-level overrides overwrote only the shared intersections, mutating the grid cells of 21A into the non-word \ans{REDOPISET} while leaving 21A's internal status marked as solved (one of four such state-divergence bugs observed during early development).
\end{enumerate}

\subsection{Synchronous Corroboration-Scored Admission (\textit{Rebuild-from-Kept})}
\label{sec:corrob-admission}

To eliminate latency-order dependence and partial-word overwrites by construction, we replaced incremental admission with a synchronous recurrence executed at the barrier after each dispatch pass $t \in \{1, \dots, T\}$ (\cref{alg:corrob}; Pass~0 corresponds to barrier $t=1$, Pass~1 to $t=2$, and so on).

\paragraph{Model Seats, Calibration Floors, and Deterministic Candidate Selection.}
Let $\mathcal{M} = \{\text{anchor}, \text{peer}, \text{third}\}$ denote the three per-clue model seats: \textbf{Gemini} (\texttt{gemini-3.1-pro-preview}) as Anchor, \textbf{Claude} (\texttt{claude-fable-5}) as Peer, and \textbf{GPT-5} (\texttt{gpt-5.6-sol}) as Third Seat.\footnote{Throughout this paper, pre-release or preview frontier model snapshots evaluated during the 2026 development period whose commercial release names remain under vendor embargo are denoted by standardized release aliases (\texttt{claude-fable-5}, \texttt{gpt-5.6-sol}, \texttt{gpt-6-astra}, \texttt{gemini-4-argon}, \texttt{gemini-3.8-flash}) paired with their exact token budgets, timeout envelopes, and reasoning-effort parameters; public API strings (\texttt{gemini-3.1-pro-preview}, \texttt{gemini-2.5-flash}) are given verbatim.}
Each seat $m \in \mathcal{M}$ is assigned an empirical confidence floor $\tau_m \in [0, 100]$, whose calibration is detailed in \cref{sec:timeouts-replicas} (\Cref{fig:calibration}a):
\begin{equation}
\label{eq:floors}
\tau_{\text{anchor}} = 50, \qquad \tau_{\text{peer}} = 85, \qquad \tau_{\text{third}} = 50.
\end{equation}

At pass barrier $t$, let $\mathcal{Q}_i^{(t)} \subset \Sigma^{\ell_i}$ denote the set of distinct qualifying candidate strings proposed for clue $i$ in pass $t$, and for each $a \in \mathcal{Q}_i^{(t)}$, let $\mathcal{M}_i^{(t)}(a) \subseteq \mathcal{M}$ be the subset of seats that proposed $a$.
Each candidate $a \in \mathcal{Q}_i^{(t)}$ receives an \textit{intrinsic base score} $b_i^{(t)}(a) \in \{1, 2\}$:
\begin{equation}
\label{eq:base-score}
b_i^{(t)}(a) = \begin{cases}
2 & \text{if } \text{anchor} \in \mathcal{M}_i^{(t)}(a) \text{ or } \left|\mathcal{M}_i^{(t)}(a)\right| \ge 2, \\[4pt]
1 & \text{otherwise (a lone qualifying peer or third-seat proposal).}
\end{cases}
\end{equation}
Whenever $\mathcal{Q}_i^{(t)} \neq \varnothing$, the barrier selects a single winning candidate $a_i^{(t)} \in \mathcal{Q}_i^{(t)}$ for clue $i$ by maximizing the tuple:
\begin{equation}
\label{eq:cand-select}
a_i^{(t)} = \operatorname*{arg\,max}_{a \in \mathcal{Q}_i^{(t)}} \left(\, b_i^{(t)}(a),\; \left|\mathcal{M}_i^{(t)}(a)\right|,\; \sum_{m \in \mathcal{M}_i^{(t)}(a)} c_m,\; \max_{m \in \mathcal{M}_i^{(t)}(a)} \operatorname{prio}(m) \,\right),
\end{equation}
and sets $b_i^{(t)} = b_i^{(t)}(a_i^{(t)})$.
In \cref{eq:cand-select}, candidates are ranked first by intrinsic base score $b_i^{(t)}(a)$, then by the number of agreeing seats $|\mathcal{M}_i^{(t)}(a)|$, and then by summed confidence $\sum c_m$.
In the rare event that two distinct single-seat proposals tie on both base score and confidence, fixed seat priority $\operatorname{prio}(\text{anchor}) = 3 > \operatorname{prio}(\text{peer}) = 2 > \operatorname{prio}(\text{third}) = 1$ provides a deterministic tie-break so that candidate selection never depends on API arrival order.
Let $S^{(t)} = \{i \in \mathcal{I} : \mathcal{Q}_i^{(t)} \neq \varnothing\}$ denote the set of clues that produce a qualifying candidate in pass $t$.

\begin{figure}[!htbp]
\centering
\includegraphics[width=\textwidth]{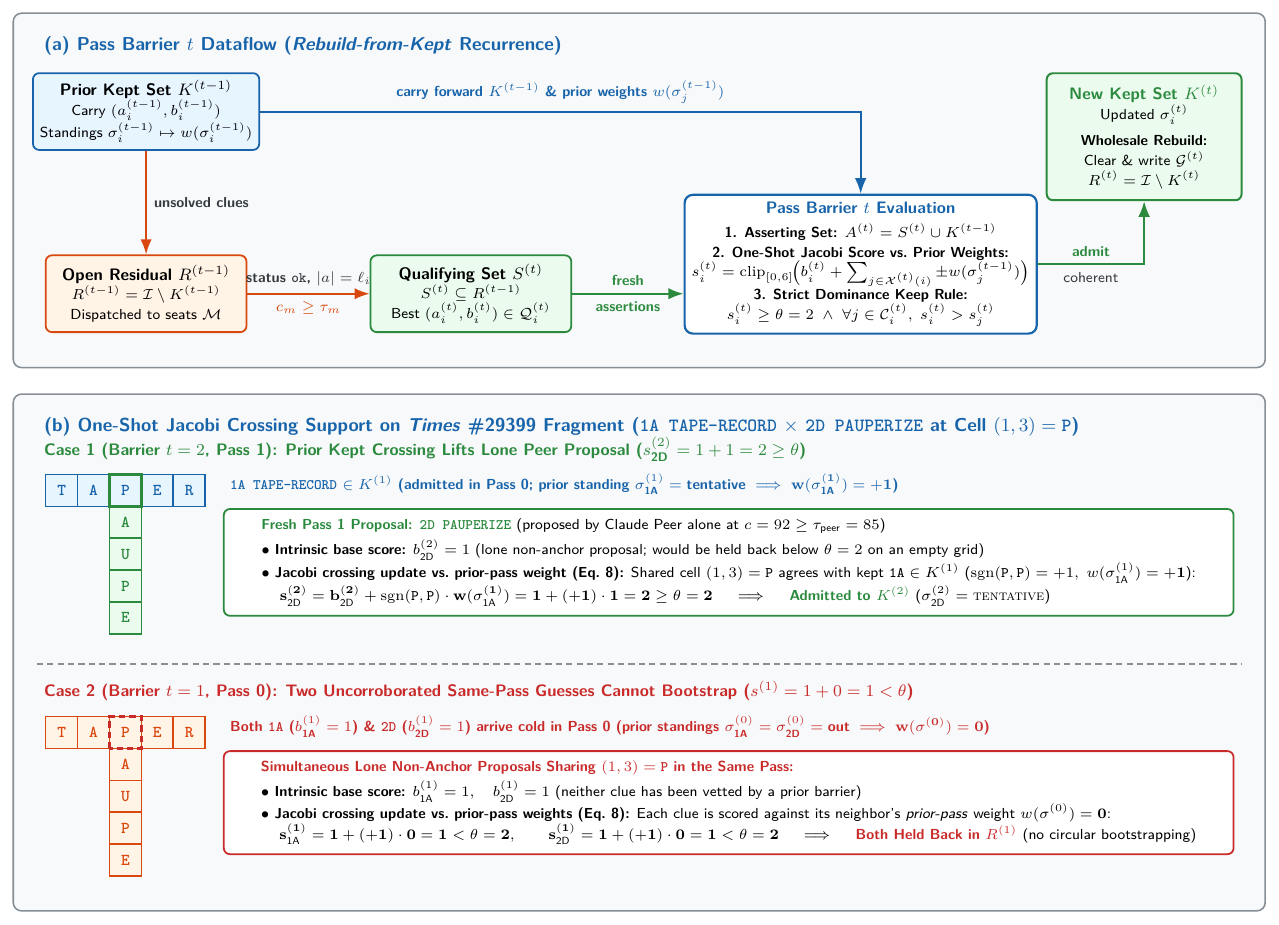}
\caption{\textbf{Pass-barrier dataflow and one-shot Jacobi crossing support.}
\textbf{(a)} At pass barrier $t$, fresh qualifying proposals $S^{(t)}$ and prior kept clues $K^{(t-1)}$ pool into $A^{(t)}$ and are scored against prior-pass weights $w(\sigma_j^{(t-1)})$.
\textbf{(b)} Illustration on \textit{Times}~\#29399 (Clues~C4 and C5): a kept Pass~0 crossing (\texttt{1A TAPE-RECORD}, $w=1$) lifts a lone Pass~1 Peer proposal (\texttt{2D PAUPERIZE}, $b=1$) to $s_{\text{2D}}^{(2)} = 2 \ge \theta$ (Case~1), whereas two uncorroborated same-pass guesses ($w=0$) cannot bootstrap each other into admission (Case~2).}
\label{fig:pass-barrier}
\end{figure}

\paragraph{Carry-Forward Asserting Set and Standing Weights.}
As illustrated in \Cref{fig:pass-barrier}a, to avoid wasting API calls, clues that were already admitted into the grid at the end of pass $t-1$, denoted by the \textit{kept set} $K^{(t-1)} \subseteq \mathcal{I}$ (initialized to $K^{(0)} = \varnothing$), are not re-dispatched in pass $t$.
However, if a newly solved crossing clue in pass $t$ conflicts with an earlier kept word in $K^{(t-1)}$, both words must compete for inclusion in the new kept set $K^{(t)}$ within the same barrier evaluation.
We therefore define the pass-$t$ \textit{asserting set} $A^{(t)}$ as the union of newly proposed clues and prior-pass kept clues:
\begin{equation}
\label{eq:asserting-set}
A^{(t)} = S^{(t)} \cup K^{(t-1)}.
\end{equation}
Each newly proposed clue $i \in S^{(t)}$ asserts its fresh pair $(a_i^{(t)}, b_i^{(t)})$, while each carry-forward clue $i \in K^{(t-1)} \setminus S^{(t)}$ retains its prior-pass pair $(a_i^{(t)}, b_i^{(t)}) = (a_i^{(t-1)}, b_i^{(t-1)})$.

To track how strongly each word has been verified by previous passes, every clue $i \in \mathcal{I}$ carries a discrete verification standing $\sigma_i^{(t)} \in \{\StandingOut, \StandingTent, \StandingSupp, \StandingLock\}$ (initialized to $\sigma_i^{(0)} = \StandingOut$) that summarizes its epistemic status at the end of barrier $t$: $\StandingOut$ (unsolved or below threshold), $\StandingTent$ (admitted at the threshold $s_i^{(t)} = \theta = 2$ without full crossing verification), $\StandingSupp$ (reinforced above the threshold, $s_i^{(t)} > 2$, by agreeing crossings), and $\StandingLock$ (every perpendicular crossing of clue $i$ is active and agrees with its letters).
In our codebase and telemetry logs, $\sigma_i^{(t)}$ is recorded as a human-readable state label, while the score recurrence maps $\sigma_i^{(t)}$ to an integer standing weight $w(\sigma_i^{(t)}) \in \{0, 1, 2, 3\}$:
\begin{equation}
\label{eq:standing-weights}
w(\StandingOut) = 0, \qquad w(\StandingTent) = 1, \qquad w(\StandingSupp) = 2, \qquad w(\StandingLock) = 3.
\end{equation}
Separating the discrete standing transition $\sigma_i^{(t)}$ from its numerical weight $w(\sigma_i^{(t)})$ prevents unbounded score inflation across passes: regardless of how many passes a word survives, its single-crossing contribution to any perpendicular neighbor is bounded by $w(\sigma_i^{(t)}) \le 3$.

\paragraph{Jacobi-Style One-Shot Crossing Support Update.}
To see how intersecting words in $A^{(t)}$ boost or penalize one another at pass barrier $t$, consider the grid fragment from \textit{Times}~\#29399 shown in \Cref{fig:pass-barrier}b.
Suppose Pass~0 (barrier $t=1$) commits 1A \ans{TAPE-RECORD} (Clue~C4; $b_{1\text{A}}^{(1)} = 2$, $\sigma_{1\text{A}}^{(1)} = \StandingTent$, weight $w = 1$), placing the crossing letter \ans{P} at row~1, column~3 ($c_{1\text{A},3} = c_{2\text{D},1} = (1,3)$, the first letter of Down clue~2D, \clue{Make poor value bandages more expensive in A\&E (9)} $\to$ \ans{PAUPERIZE}; Clue~C5).\footnote{In the full puzzle (\Cref{fig:primer-grids}b), 2D also crosses 10A \ans{MAUNDER} (Clue~C6) at \ans{U}, 12A \ans{SET ON A PEDESTAL} at \ans{E}, 14A \ans{CLIPPINGS} at \ans{I}, and 19A \ans{NEEDS} at \ans{E}; for clarity, \Cref{fig:pass-barrier}b isolates the single intersection between 1A and 2D.}
In Pass~1 ($t=2$), suppose Claude alone proposes \ans{PAUPERIZE} at confidence $92$ ($b_{2\text{D}}^{(2)} = 1$).
On an empty grid, a lone Peer proposal ($b=1$) is held back below the admission threshold $\theta = 2$; here, however, because \ans{PAUPERIZE} agrees with the already-kept crossing 1A \ans{TAPE-RECORD} at letter \ans{P}, 1A lends its prior-pass standing weight $+w(\sigma_{1\text{A}}^{(1)}) = +1$ to 2D, lifting 2D's total corroboration score to $s_{2\text{D}}^{(2)} = 1 + 1 = 2 \ge \theta$ so that it is admitted (\Cref{fig:pass-barrier}b, Case~1).
Conversely, if two lone non-anchor guesses ($b=1$) for 1A and 2D arrive in the same cold pass with no prior standing ($\sigma^{(0)} = \StandingOut$, $w = 0$) and happen to share the letter \ans{P}, each is evaluated against the other's prior-pass weight $w(\sigma^{(0)}) = 0$, leaving both at $s^{(1)} = 1 + 0 = 1 < \theta$ so they cannot bootstrap each other into admission (\Cref{fig:pass-barrier}b, Case~2).

In numerical analysis terms, updating each clue's support score using its neighbors' \textit{prior-pass} standings $\sigma_j^{(t-1)}$ rather than their current-pass standings $\sigma_j^{(t)}$ follows a \textbf{synchronous Jacobi update schedule} rather than an in-place Gauss--Seidel sweep~\cite{saad2003iterative}: within each barrier evaluation, the score calculation is strictly independent of clue ordering, requires no inner iteration loop, and prevents unvetted same-pass guesses ($\sigma_j^{(t-1)} = \StandingOut, w = 0$) from amplifying one another into admission before surviving a pass barrier.

Formally, for any clue $i \in A^{(t)}$, let $\mathcal{X}(i) \subseteq \mathcal{I} \setminus \{i\}$ denote the perpendicular clues that share a checked cell with $i$, let $\mathcal{X}^{(t)}(i) = \mathcal{X}(i) \cap A^{(t)}$ be its active crossings at barrier $t$, and let
\begin{equation}
\label{eq:conflicts}
\mathcal{C}_i^{(t)} = \left\{\, j \in \mathcal{X}^{(t)}(i) \;\middle|\; a_i^{(t)}[p_{ij}] \neq a_j^{(t)}[p_{ji}] \,\right\}
\end{equation}
be the subset of active crossings whose asserted letters conflict with $a_i^{(t)}$.
Each clue $i \in A^{(t)}$ computes its corroboration score $s_i^{(t)} \in [0, 6]$ in a single synchronous step against $\sigma_j^{(t-1)}$:
\begin{equation}
\label{eq:score-update}
s_i^{(t)} = \clip_{[0,\,6]}\!\left( b_i^{(t)} + \sum_{j \in \mathcal{X}^{(t)}(i)} \sgn\!\left(a_i^{(t)}[p_{ij}],\, a_j^{(t)}[p_{ji}]\right) \cdot w\!\left(\sigma_j^{(t-1)}\right) \right),
\end{equation}
where $\sgn(x, y) = +1$ if $x = y$ and $-1$ if $x \neq y$, and $\clip_{[0,6]}(z) = \min(6, \max(0, z))$.
Clipping $s_i^{(t)}$ to the interval $[0, 6]$ (twice the maximum single-crossing weight, $w(\StandingLock) = 3$) ensures that a long 14-letter entry with seven checked intersections cannot accumulate an unbounded support score that renders it immune to eviction if two high-confidence perpendicular words later contradict it.

\paragraph{Strict Dominance Keep Rule and Standing Transition.}
We fix the \textit{admission threshold}, the minimum corroboration score required for a candidate word to be written into the grid, at $\theta = 2$, so that a word requires either an intrinsic base score $b_i^{(t)} = 2$ (two-seat agreement or an Anchor proposal) or an intrinsic base score $b_i^{(t)} = 1$ (a lone Peer or Third-Seat proposal) corroborated by at least one kept crossing ($\ge +1$).
Clue $i \in A^{(t)}$ is admitted into the new kept set $K^{(t)}$ if and only if its score reaches $\theta = 2$ and it \textit{strictly outranks} every conflicting crossing clue $j \in \mathcal{C}_i^{(t)}$:
\begin{equation}
\label{eq:keep-rule}
K^{(t)} = \left\{\, i \in A^{(t)} \;\middle|\; s_i^{(t)} \ge \theta \;\land\; \forall j \in \mathcal{C}_i^{(t)},\; s_i^{(t)} > s_j^{(t)} \,\right\}.
\end{equation}
For completeness, we define the full-crossing lock indicator $\text{is\_lock}_i^{(t)} = \Ind\bigl[\mathcal{X}(i) \neq \varnothing \;\land\; \mathcal{X}(i) \subseteq A^{(t)} \;\land\; \mathcal{C}_i^{(t)} = \varnothing\bigr]$, which equals $1$ if and only if clue $i$ has at least one perpendicular crossing and every perpendicular crossing of $i$ is active in $A^{(t)}$ and agrees with $a_i^{(t)}$.
While only kept entries $i \in K^{(t)}$ enter the rebuilt grid $\mathcal{G}^{(t)}$ and carry their base score $b_i^{(t)}$ into $A^{(t+1)}$ (\cref{eq:asserting-set}), the standing $\sigma_i^{(t)}$ carried into barrier $t+1$ is updated via:
\begin{equation}
\label{eq:standing-assign}
\sigma_i^{(t)} = \begin{cases}
\StandingLock & \text{if } i \in A^{(t)} \text{ and } \text{is\_lock}_i^{(t)} = 1, \\[4pt]
\StandingSupp & \text{if } i \in A^{(t)},\; \text{is\_lock}_i^{(t)} = 0, \text{ and } s_i^{(t)} > \theta, \\[4pt]
\StandingTent & \text{if } i \in A^{(t)},\; \text{is\_lock}_i^{(t)} = 0, \text{ and } s_i^{(t)} = \theta, \\[4pt]
\StandingOut  & \text{if } i \notin A^{(t)} \text{ or } \bigl(\text{is\_lock}_i^{(t)} = 0 \text{ and } s_i^{(t)} < \theta\bigr).
\end{cases}
\end{equation}
In plain operational terms, \cref{eq:keep-rule,eq:standing-assign} enforce three behaviors:
\begin{enumerate}[leftmargin=1.8em,itemsep=2pt]
    \item \textbf{Immediate lock promotion when all crossings agree.}
    Although two uncorroborated guesses ($b=1$) cannot bootstrap each other into $K^{(t)}$ in the same pass (\Cref{fig:pass-barrier}b, Case~2), once a word is admitted ($i \in K^{(t)}$, $s_i^{(t)} \ge 2$) and all of its perpendicular crossings are also active and agreeing in $A^{(t)}$ ($\text{is\_lock}_i^{(t)} = 1$), its standing is promoted immediately to $\StandingLock$ ($w = 3$) for the next pass.
    \item \textbf{One-barrier standing memory for non-kept assertions.}
    If an asserting word $i \in A^{(t)} \setminus K^{(t)}$ is held out of the grid, either because it tied or lost a crossing conflict ($s_i^{(t)} \le s_j^{(t)}$) or because it was a lone $b_i^{(t)} = 1$ proposal whose perpendicular crossings all agreed with it, it is excluded from $K^{(t)}$ and $\mathcal{G}^{(t)}$, but carries its assertion standing $\sigma_i^{(t)}$ into barrier $t+1$ (contributing weight $w(\sigma_i^{(t)})$ if it re-asserts in $S^{(t+1)}$).
    \item \textbf{Wholesale grid reconstruction.}
    After $K^{(t)}$ is computed, the physical grid $\mathcal{G}^{(t)}$ is cleared and written from scratch using only $\{(i, a_i^{(t)}) : i \in K^{(t)}\}$.
\end{enumerate}

Because strict inequality ($s_i^{(t)} > s_j^{(t)}$) drops both sides of any conflict tie from $K^{(t)}$, the kept set $K^{(t)}$ is guaranteed to be conflict-free:

\begin{invariant}[Boundary Grid Coherence]
\label{inv:coherence}
At every pass barrier $t \in \{1, \dots, T\}$, for all $i, j \in K^{(t)}$ with $j \in \mathcal{X}(i)$, $a_i^{(t)}[p_{ij}] = a_j^{(t)}[p_{ji}]$.
Consequently, wholesale projection of $K^{(t)}$ onto $\mathcal{V}$ is order-independent and free of partial-word corruption.
\end{invariant}
\begin{proof}
Suppose for contradiction that $i, j \in K^{(t)}$ intersect at cell $v = c_{i,p_{ij}} = c_{j,p_{ji}}$ with $a_i^{(t)}[p_{ij}] \neq a_j^{(t)}[p_{ji}]$.
Then $j \in \mathcal{C}_i^{(t)}$ and $i \in \mathcal{C}_j^{(t)}$, so \cref{eq:keep-rule} requires both $s_i^{(t)} > s_j^{(t)}$ and $s_j^{(t)} > s_i^{(t)}$, which is impossible; subsequent boundary-feed admissions verify cell compatibility against $\mathcal{G}^{(t)}$ before placement.
\end{proof}

\begin{algorithm}[!htbp]
\caption{Synchronous Corroboration-Scored Boundary Admission (\textit{Rebuild-from-Kept})}
\label{alg:corrob}
\begin{algorithmic}[1]
\Require Pass barrier $t$, round responses $\{P_i^{(t)}\}_{i \in \mathcal{I}}$, prior state $(K^{(t-1)}, \{a_i^{(t-1)}, b_i^{(t-1)}, \sigma_i^{(t-1)}\}_{i \in \mathcal{I}})$
\Ensure Updated kept set $K^{(t)}$ and coherent grid state $\mathcal{G}^{(t)}$
\State $S^{(t)} \gets \varnothing$
\ForAll{$i \in \mathcal{I}$}
    \State $\mathcal{Q}_i^{(t)} \gets \{\text{normalized } a : (m, a, c_m) \in P_i^{(t)} \text{ has status \texttt{ok}, } \vert{}a\vert{}=\ell_i, \text{ and } c_m \ge \tau_m\}$
    \If{$\mathcal{Q}_i^{(t)} \neq \varnothing$}
        \State Select unique $a_i^{(t)} \in \mathcal{Q}_i^{(t)}$ via \cref{eq:base-score,eq:cand-select}; set $b_i^{(t)} \gets b_i^{(t)}(a_i^{(t)})$ and $S^{(t)} \gets S^{(t)} \cup \{i\}$
    \EndIf
\EndFor
\State $A^{(t)} \gets S^{(t)} \cup K^{(t-1)}$ \Comment{Carry forward kept clues $i \in K^{(t-1)} \setminus S^{(t)}$}
\ForAll{$i \in A^{(t)}$}
    \State $\mathcal{X}^{(t)}(i) \gets \mathcal{X}(i) \cap A^{(t)}$; \quad $\mathcal{C}_i^{(t)} \gets \{j \in \mathcal{X}^{(t)}(i) : a_i^{(t)}[p_{ij}] \neq a_j^{(t)}[p_{ji}]\}$
    \State $s_i^{(t)} \gets \clip_{[0,6]}\!\left(b_i^{(t)} + \sum_{j \in \mathcal{X}^{(t)}(i)} \sgn(a_i^{(t)}[p_{ij}], a_j^{(t)}[p_{ji}]) \cdot w(\sigma_j^{(t-1)})\right)$
\EndFor
\State $K^{(t)} \gets \bigl\{i \in A^{(t)} : s_i^{(t)} \ge \theta \;\land\; \forall j \in \mathcal{C}_i^{(t)},\; s_i^{(t)} > s_j^{(t)}\bigr\}$
\State Assign assertion standing $\sigma_i^{(t)}$ for $i \in A^{(t)}$ and $\sigma_i^{(t)} \gets \StandingOut$ for $i \notin A^{(t)}$ via \cref{eq:standing-assign}
\State Clear grid $\mathcal{G}^{(t)}$ and write $a_i^{(t)}$ for all $i \in K^{(t)}$ (carrying $b_i^{(t)}$ only for $i \in K^{(t)}$); set residual $R^{(t)} \gets \mathcal{I} \setminus K^{(t)}$
\State \Return $K^{(t)}, \mathcal{G}^{(t)}$
\end{algorithmic}
\end{algorithm}

Notice how \cref{alg:corrob} handles the three main operational cases cleanly.
\begin{itemize}[leftmargin=1.8em,itemsep=2pt]
    \item \textbf{Cold consensus or anchor admission.}
    Any candidate agreed upon by $\ge 2$ seats or proposed by the Anchor receives $b_i^{(1)} = 2 = \theta$ and enters $K^{(1)}$ unless contradicted by an equal- or higher-scored crossing.
    \item \textbf{Single-seat holding and crossing lift.}
    A lone Peer ($\ge 85$) or Third-Seat ($\ge 50$) proposal receives $b_i^{(t)} = 1 < \theta$ and is held back on an empty grid, but enters $K^{(t)}$ as soon as at least one kept crossing agrees with it ($+w(\sigma_j^{(t-1)}) \ge +1$) without an uncompensated conflict.
    \item \textbf{Clean whole-word eviction.}
    Suppose an isolated wrong word $i$ enters $K^{(1)}$ in Pass~0 with $s_i^{(1)} = 2$ and $\sigma_i^{(1)} = \StandingTent$ ($w = 1$), as on \textit{Times}~\#29570 clue~21A (\ans{REPOSITED}; Clue~C7).
    At barrier $t = 2$, when a fresh perpendicular word $j \in S^{(2)}$ ($b_j^{(2)} = 2, \sigma_j^{(1)} = \StandingOut$) conflicts with incumbent $i$ ($-w(\sigma_i^{(1)}) = -1$) while agreeing with another kept crossing $k \in K^{(1)}$ ($+w(\sigma_k^{(1)}) \ge +1$), the challenger reaches $s_j^{(2)} \ge 2 - 1 + 1 = 2$ while incumbent $i$ stays at $s_i^{(2)} = 2$.
    Under strict inequality ($s_i^{(2)} > s_j^{(2)}$), a $2 = 2$ tie drops \textit{both} words back into $R^{(2)}$ (and if $j$ has two kept supporting crossings, $s_j^{(2)} \ge 3 > 2$ admits $j$ and drops $i$), evicting the wrong incumbent in its entirety so it can be re-solved against the updated crossing pattern.
\end{itemize}
In offline replay across $135$ banked daily runs ($132$ canonical May~19 -- October~9 daily runs plus $3$ early May diagnostic runs), replacing greedy admission with \cref{alg:corrob} gained $+34$ correct clues ($3{,}804 \to 3{,}838$), reduced grid errors from $10$ to $4$, eliminated the 21A \ans{REDOPISET} corruption, and maintained $100\%$ grid coherence.
In live paired testing across $40$ hard puzzles ($80$ runs), synchronous barrier admission gained $+18$ correct clues ($1{,}156 \to 1{,}174$, $\bar{\Delta}/\mathrm{SE} = 2.57$), halved unfilled clues ($33 \to 15$), reduced API cost by $41\%$, and cut solve time by $27\%$.

\subsection{The Iterative Constraint-Tightening Pipeline}
\label{sec:pipeline}

Around \cref{alg:corrob}, the full solving pipeline separates active barrier passes (Pass~0 through the Tail Rounds, which invoke \cref{alg:corrob} and can evict a contradicted kept word) from strictly non-evicting auxiliary stages (the asynchronous boundary feed and the local terminal chain, which may only fill still-open clues in $R^{(t)}$ without ever overwriting an incumbent kept answer in $K^{(t)}$; \Cref{fig:arch}b).

\begin{enumerate}[leftmargin=1.8em,itemsep=3pt]
    \item \textbf{Pass 0 (Unconstrained Cold Pass, Barrier $t=1$).}
    All $m$ clues (typically $30$) are dispatched concurrently with an all-wildcard pattern (e.g., \texttt{??????}) to the two primary seats (Gemini Anchor and Claude Peer), using a per-seat concurrency cap of $33$ and a $150\text{ ms}$ inter-request pacing stagger (ramping $30$ same-seat requests over $4.5\text{s}$ to prevent socket/DNS bursts).
    At barrier $t=1$, \cref{alg:corrob} computes $K^{(1)}$ and writes it to the grid.

    \item \textbf{Passes 1--3 (Crossing-Constrained Retries with Retry-Once Gate, Barriers $t=2\text{--}4$).}
    While the residual open set $R^{(t-1)} = \mathcal{I} \setminus K^{(t-1)}$ is non-empty, open clues are re-dispatched for up to three passes to all three seats (adding GPT-5 at per-seat concurrency caps of $10$, $10$, and $8$ for Gemini, Claude, and GPT-5, sized to stay below per-minute token rate limits on retry batches, where typically $|R^{(1)}| \le 10$ clues remain open).
    Each prompt provides the clue's current crossing wildcard pattern (e.g., \texttt{P?U?E?I?E}).
    In addition, the runner queries a composite curated lexicon ($183{,}139$ entries, priority-ordered with $128{,}188$ entries from our primary local wordlist first, followed by $54{,}951$ additional entries from a DataMuse- and \textit{Collins English Dictionary}-validated wordlist) and the UK Advanced Cryptics Dictionary (\textsc{UKACD}, $250{,}496$ normalized entries in \texttt{UKACD18plus.txt}) against the wildcard pattern: if the pattern has at least one known crossing letter and matches between $1$ and $100$ lexicon entries, those matching words are appended to the prompt as non-binding candidate hints (if $0$ or $>100$ entries match, only the bare wildcard pattern is sent).
    Passes~$2$ and~$3$ also include each seat's own prior-pass attempt for multi-turn self-correction, while a \textit{Retry-Once Gate} skips re-dispatch in Passes~2--3 (barriers $t = 3, 4$) for any clue whose wildcard pattern has not changed since its last completed dispatch.

    \item \textbf{Pass 4 (Historical-Answer Pass, Barrier $t=5$).}
    Any clue still open in $R^{(4)}$ whose crossing pattern matches between $1$ and $100$ entries in the point-in-time historical \textit{Times} answer list (filtered strictly to puzzles published prior to the target puzzle's publication date) is dispatched with those historical answers as non-binding hints, followed by barrier $t=5$ of \cref{alg:corrob}.

    \item \textbf{Tail Re-dispatch Rounds (Capped at 2 Rounds).}
    If late admissions or evictions in Pass~3 or Pass~4 alter any crossing pattern, up to two additional three-seat passes are dispatched on the residual using bare tightened wildcard patterns, terminating as soon as no kept answer or crossing pattern changes or after the 2-round cap.

    \item \textbf{Asynchronous Whole-Grid Boundary Feed (\textsc{bf}).}
    At solve start ($t_0 = 0$), the runner spawns a non-blocking background process that sends the entire puzzle grid and all $30$ clues in a single prompt to \texttt{gpt-6-astra}.
    At the end of each main pass barrier (after \cref{alg:corrob} has reconstructed $K^{(t)}$), the runner checks whether the background process has written its output file.
    If so, it scans still-unfilled clues $i \in R^{(t)}$ in canonical clue order (Across before Down, then ascending clue number): any clue whose feed answer matches $\ell_i$, clears confidence $c_i^{\text{feed}} \ge 80$, and agrees with all currently filled crossing cells is admitted into $\mathcal{G}^{(t)}$ and added to $K^{(t)}$ with carry-forward base score $b_i^{(t)} = 2$ and standing $\sigma_i^{(t)} = \StandingTent$, without ever evicting an incumbent kept answer.
    If the per-clue pipeline completes the grid before the whole-grid process returns, the background process is terminated immediately ($2.0\text{s}$ reap wait).

    \item \textbf{Local Terminal Chain (Capped at 10 Iterations).}
    For any clue still open after the last model pass barrier, a deterministic, purely local post-processing chain (making zero API calls) executes four non-evicting stages in sequence.
    \begin{enumerate}[label=(\alph*),leftmargin=1.6em,itemsep=1pt]
        \item \textit{Corroborated Singleton (\textsc{corr}).}
        Commits a UKACD wildcard-pattern singleton if and only if that word was previously proposed by at least one model seat on an earlier pass.
        \item \textit{Cross-Pass Persistence (\textsc{pers}).}
        Iterates jointly with \textsc{corr} (up to a 10-iteration cap), committing an earlier status-\texttt{ok}, crossing-consistent proposal either when two distinct seats proposed it across passes or when a single seat re-proposed it after at least one strict crossing-pattern tightening.
        \item \textit{Anchor Terminal Fallback (\textsc{atf}).}
        Commits the highest-confidence qualifying Gemini Anchor proposal ($\ge 50$) from earlier passes if it remains consistent with all current crossings (logged under the event tag \texttt{gemini\_terminal\_fallback}).
        \item \textit{Terminal Dictionary Sweep (\textsc{swp}).}
        Because tournament scoring penalizes a blank cell and an incorrect cell identically, filling any remaining open clue with an uncorroborated UKACD singleton (Tier-1), a point-in-time historical-answer singleton (Tier-2), or a deterministic merged-lexicon pick matching the crossing pattern is weakly dominant on that clue (and strictly improves expected score whenever lexicon match accuracy is $>0$) without risking any already-committed crossing word.
    \end{enumerate}
\end{enumerate}

\subsection{Empirical Calibration of Per-Seat Timeout Caps}
\label{sec:timeouts-replicas}

\begin{figure}[!htbp]
\centering
\includegraphics[width=\textwidth]{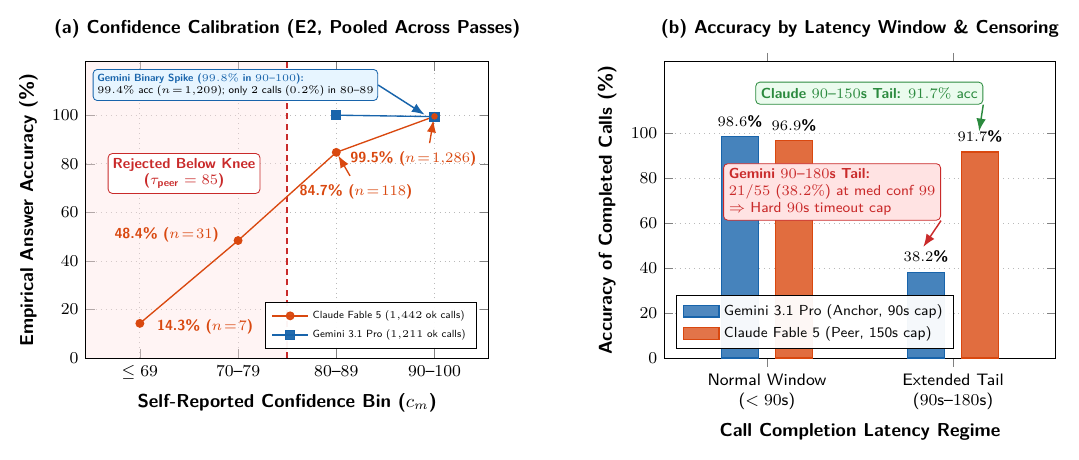}
\caption{\textbf{Asymmetric per-seat confidence calibration and late-tail latency survival.}
\textbf{(a)} Empirical accuracy by self-reported confidence bin across the 43-day Epoch~E2 daily corpus ($1{,}211$ Gemini and $1{,}442$ Claude status-\texttt{ok} calls; \Cref{tab:app-prod}, Panel~B), showing binary confidence on Gemini versus a graded curve with a sharp knee at $\tau_{\text{peer}} = 85$ on Claude.
\textbf{(b)} Answer accuracy for normal completions ($<90\text{s}$) versus extended-tail completions ($90$--$180\text{s}$), motivating Gemini's $90\text{s}$ timeout cap and Claude's $150\text{s}$ cap.}
\label{fig:calibration}
\end{figure}

Because each pass barrier in \cref{sec:pipeline} waits for all in-flight calls in that pass to finish or time out, a single hung API call delays the entire pass up to the seat's timeout cap.\footnote{Across the $43$ Epoch~E2 daily puzzles, median total puzzle solve time is $336\text{s}$ ($5\text{m}\,36\text{s}$), with Pass~0 taking a median of $163\text{s}$, Pass~1 taking $156\text{s}$, and a median of $86.5\text{s}$ per day ($\text{p90} = 255.6\text{s}$) spent waiting at pass barriers after the last status-\texttt{ok} return for straggler calls that ultimately time out.}
Rather than choosing timeout caps heuristically, we calibrated per-seat timeouts and token budgets against empirical confidence and latency-survival distributions (\Cref{fig:calibration}).
\begin{itemize}[leftmargin=1.8em,itemsep=2pt]
    \item \textbf{Gemini Anchor ($90\text{s}$ timeout, $11{,}378$ output tokens).}
    Across $1{,}211$ completed calls pooled across passes in our 43-day Epoch~E2 daily corpus (\Cref{fig:calibration}a), Gemini's self-reported confidence is binary ($99.8\%$ in $[90, 100]$ at $99.4\%$ accuracy, and $98.6\%$ accuracy across normal $<90\text{s}$ completions), making $\tau_{\text{anchor}} = 50$ effectively non-binding on status-\texttt{ok} anchor responses.
    To test whether raising Gemini's $90\text{s}$ timeout cap would recover additional solves, we re-ran $200$ timed-out Gemini calls under a relaxed $180\text{s}$ cap.
    While $98/200$ calls completed under the $180\text{s}$ cap, $48$ of those $98$ finished in $<90\text{s}$ upon simple re-dispatch (reflecting transient provider queue variance), leaving $50$ calls that still required $90$--$180\text{s}$ of thinking (plus $5$ baseline calls that slightly exceeded $90\text{s}$, for $55$ total completions in the $90$--$180\text{s}$ window).
    That $90$--$180\text{s}$ tail was only \textbf{$38.2\%$ accurate ($21/55$) while still reporting a median confidence of $99$} (\Cref{fig:calibration}b).
    In other words, when Gemini gets stuck on a dead-end parse for more than $90\text{s}$, prolonged thinking produces high-confidence confabulations that degrade grid accuracy if admitted into \cref{alg:corrob}.

    Confirming this threshold across our full test-set logs, among all $207$ Gemini calls that timed out at $90\text{s}$ across the 16-puzzle benchmark, none cost a final grid answer because every such clue was solved on the subsequent crossing-constrained pass or by a peer seat (and similarly on daily runs such as \textit{Times}~\#29611 clue~15D \ans{EXECUTOR}; Clue~C8, \Cref{tab:app-clues-1}).
    \item \textbf{Claude Peer ($150\text{s}$ timeout, $20{,}000$ output tokens).}
    Unlike Gemini, Claude exhibits a smoothly graded confidence curve with a steep drop between its $[80,89]$ and $[70,79]$ decile bins (\Cref{fig:calibration}a), motivating the selected operating threshold $\tau_{\text{peer}} = 85$.
    Across our 10-day pre-freeze latency characterization corpus ($266$ Claude-solved clues), Claude's p90 latency was $58$--$62\text{s}$, and $4.1\%$ ($11/266$) of its correct solves required between $90\text{s}$ and $150\text{s}$ of extended thinking at $91.7\%$ accuracy (\Cref{fig:calibration}b).

    Capping Claude at $90\text{s}$ would abort productive deep reasoning; capping at $150\text{s}$ retains it while bounding hangs.
    \item \textbf{GPT-5 Third Seat ($150\text{s}$ cap, $13{,}652$ output tokens).}
    Across all $76$ clues in our development corpus where GPT-5 was the sole seat to propose the correct answer during Passes~1--3, its p90 latency was $78\text{s}$ and its maximum productive latency was $142\text{s}$ (with $\tau_{\text{third}} = 50$ acting as a sanity filter since lone Third-Seat proposals always receive $b=1 < \theta$).
    Tightening its timeout cap from $240\text{s}$ to $150\text{s}$ retained $100\%$ ($76/76$) of its unique solves while trimming mean puzzle solve time by $12.6\text{s}$.
\end{itemize}

\subsection{Triple-Replica Execution and Dispatch Decorrelation}
\label{sec:triple-replica}

Finally, to bound run-to-run stochasticity and tail latency under the championship's speed tie-breaker~\cite{dean2013tail}, the Championship Final configuration launches $D = 3$ active-active solve replicas in parallel on the single Final puzzle (indexed as Replica~0, Replica~1, and Replica~2, or \texttt{run\_0}, \texttt{run\_1}, and \texttt{run\_2}) and submits the grid of the \textit{first replica to finish} (a first-finisher latency race rather than majority voting across replicas).

In the three-puzzle Qualifying and Semi-Final rounds, each of the three execution panes solves one of the three distinct round puzzles concurrently ($D=1$ per puzzle), so all three tournament rounds issue the exact same opening wave of $3 \times 30 = 90$ concurrent requests per primary seat.
When running three replicas on the Final, the replicas share a single background \texttt{gpt-6-astra} whole-grid process and apply deterministic \textit{Replica Dispatch Permutation} (\textsc{r3-rdp}, seeds $\{\texttt{None}, 2, 3\}$ for \texttt{run\_0}, \texttt{run\_1}, and \texttt{run\_2}): \texttt{run\_0} issues every pass batch in canonical clue order, while \texttt{run\_1} and \texttt{run\_2} permute the flattened $(\text{clue}, \text{seat})$ dispatch list using a SHA-256 digest of $(\text{seed}, \text{phase\_id}, \text{batch\_index})$ while preserving the $150\text{ ms}$ per-seat pacing stagger in Pass~0.
Because the three replicas share the same host, project quota, and background feed process, they are not statistically independent against project-wide brownouts,\footnote{Concentrating all three panes onto the single Final puzzle on one host and API project carries a clear self-contention hazard, but matched the operational arrangement agreed with the Championship organizers, wherein the three puzzle panes used for the three concurrent Qualifying and Semi-Final puzzles were assigned to run three replicas of the single Final puzzle.} but permuting clue dispatch order ensures that different clues occupy the earliest dispatch slots across replicas.

\section{Evaluating Common LLM Prompting Choices: Controlled Clue-Level Experiments}
\label{sec:folklore}

\begin{figure}[!htbp]
\centering
\includegraphics[width=\textwidth]{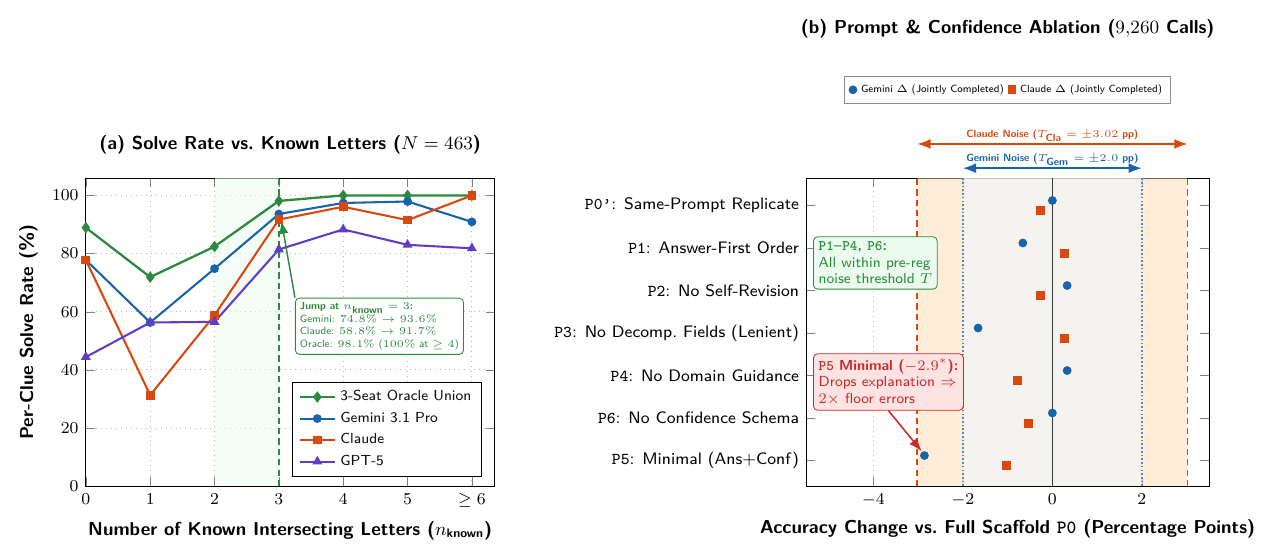}
\caption{\textbf{Controlled clue-level experiments across $N = 463$ test-set clues.}
\textbf{(a)} Per-clue solve rate across strata of known intersecting letters $n_{\text{known}}$ (\cref{sec:crossings-amp}, \Cref{tab:app-crossings}), showing saturation above $90\%$ at $n_{\text{known}} \ge 3$ ($n_{\text{known}}=0$ [$n=9$] comprises short 4--5 letter clues).
\textbf{(b)} Paired admissible accuracy change relative to the full production scaffold (\texttt{P0}) on jointly completed calls across prompt and confidence-elicitation arms \texttt{P0'}--\texttt{P6} (\cref{sec:prompt-ablation,sec:conf-ablation}, \Cref{tab:prompt-summary}): only the Minimal prompt (\texttt{P5}, dropping the wordplay explanation) degrades accuracy outside replicate noise.}
\label{fig:folklore}
\end{figure}

In building neuro-symbolic reasoning systems, practitioners often bundle mechanisms such as external constraint feedback, multi-field chain-of-thought (CoT) prompt schemas, draft-and-revise instructions, verbalized confidence scoring, and symbolic post-hoc verifiers, and credit the bundle as a whole for system performance.
Before evaluating full-grid orchestration in \cref{sec:eval}, we test four common clue-level design choices in controlled experiments across our 463-clue stratified test set (\Cref{fig:folklore}): (1) crossing-letter constraint amplification (\cref{sec:crossings-amp}), (2) structured CoT prompt decomposition (\cref{sec:prompt-ablation}), (3) verbalized confidence elicitation (\cref{sec:conf-ablation}), and (4) symbolic Python-AST wordplay verification (\cref{sec:verifier-falsification}).

\subsection{Crossings as the Primary Reasoning Amplifier}
\label{sec:crossings-amp}

A per-clue solver whose single-call cold accuracy is only $44.1\%$--$63.6\%$ on unconstrained clues (\Cref{tab:app-crossings}, Panel~A) still reaches $99.6\%$--$99.8\%$ accuracy on full grids because intersecting letters rapidly narrow the candidate space.
To measure how solve rates vary with intersecting letters across the difficulty spectrum, we evaluated all $N = 463$ clues of the stratified 16-puzzle test set across all three model seats ($463 \times 3 = 1{,}389$ calls\footnote{In this crossing-amplification experiment, the secondary seats used the \texttt{claude-opus-4-7} and \texttt{gpt-5} snapshots prior to the final seat freeze; because the steep jump to $>90\%$ saturation at $n_{\text{known}} \ge 3$ appears identically on the frozen \texttt{gemini-3.1-pro-preview} anchor ($56.3\% \to 74.8\% \to 93.6\%$) and across both secondary seats, the structural pattern is invariant to snapshot version.}) under the crossing-letter patterns encountered during retry passes.
Following production dispatch rules, clues whose intersecting letter pattern matched $\le 100$ curated lexicon entries ($n = 320$, $69.1\%$) were dispatched with the wildcard pattern and candidate list, whereas clues with $> 100$ matches ($n = 143$, $30.9\%$) were dispatched cold without candidates (because the $320$ vs.\ $143$ subsets partition the test set by natural retry state and bundle pattern letters with candidate lists when matches $\le 100$, the differences below represent between-stratum operational contrasts rather than a randomized single-variable treatment).

\Cref{fig:folklore}a (and \Cref{tab:app-crossings} in \Cref{app:tables}) reveals a steep, saturating accuracy jump across crossing strata (providing the clue-level empirical mechanism that drives grid-level cluster percolation~\cite{hartmann2024percolation}).
On the $>100$-match cold stratum, single-call accuracy is $63.6\%$ ($91/143$) on Gemini, $44.1\%$ ($63/143$) on Claude, and $47.6\%$ ($68/143$) on GPT-5; on the $\le 100$-match crossing-constrained stratum, accuracy is higher by $+33.0$ to $+50.3$ percentage points across seats, reaching $96.6\%$ ($309/320$) on Gemini and $94.4\%$ ($302/320$) on Claude.
Stratifying separately by the integer count of known crossing letters $n_{\text{known}}$ (\Cref{fig:folklore}a; \Cref{tab:app-crossings}, Panel~B) and by continuous crossing density $n_{\text{known}} / \ell_i$ (Panel~C) shows how crossing letters narrow the search space:
\begin{itemize}[leftmargin=1.8em,itemsep=2pt]
    \item At $n_{\text{known}} = 1$--$2$ known crossing letters (or in the $1\%$--$25\%$ density band; aside from $n=9$ short clues at $n_{\text{known}}=0$ that solved at $77.8\%$), wildcard patterns typically match $>100$ lexicon entries, and single-seat solve rates climb from $31.3\%$--$56.3\%$ ($n_{\text{known}}=1$) to $56.5\%$--$74.8\%$ ($n_{\text{known}}=2$, $82.4\%$ three-seat oracle union).
    \item At $n_{\text{known}} = 3$ crossing letters (and similarly in the $26\%$--$40\%$ density band), single-seat solve rates jump by an additional $+18.8$ to $+32.9$ percentage points, reaching $93.6\%$ ($146/156$) on Gemini, $91.7\%$ ($143/156$) on Claude, and $81.4\%$ ($127/156$) on GPT-5 ($98.1\%$ oracle union).
    \item At $n_{\text{known}} \ge 4$ (or $>55\%$ density), the three-seat oracle union reaches \textbf{$100.0\%$ ($135/135$ clues)}.
\end{itemize}
Providing anchored crossing letters in the prompt prunes false candidate definitions, anagrams, and containers during the model's extended-thinking trace.
Two additional findings from this experiment directly informed \Darmok{}'s pipeline (\cref{sec:pipeline}): first, on the $n = 71$ test-set clues that failed cold in Pass~0, crossing-constrained re-dispatch solved $63/71$ ($88.7\%$); second, on clues already solved cold in Pass~0 ($n = 307$), re-dispatching with candidate lists preserved the correct answer only $88.3\%$--$95.4\%$ of the time per seat, demonstrating that constrained retries should only be dispatched on the uncommitted residual $R^{(t)}$ rather than re-running over already-corroborated cold commits.

\subsection{Prompt Scaffolding versus Native Extended Thinking}
\label{sec:prompt-ablation}

A common design pattern in applied LLM engineering is that compositional word puzzles require elaborate chain-of-thought prompt schemas~\cite{sadallah2024good,saha2025lmcrossword} that instruct the model to identify the definition, classify the wordplay operator, write a step-by-step derivation, emit an unverified draft answer, critique its draft, and finally emit a revised answer.
Our production Pass~0 prompt (\texttt{P0}, reproduced verbatim in \Cref{app:prompts-clues} with inline markers indicating the exact lines modified by each variant) implemented this full scaffold as a structured JSON schema with eight fields (\texttt{definition}, \texttt{wordplay\_type}, \texttt{is\_and\_lit}, \texttt{wordplay\_explanation}, \texttt{answer\_unverified}, \texttt{confidence.level}, \texttt{confidence.reason}, and final \texttt{answer}), accompanied by domain solving guidance and enumeration examples.

To test whether any part of this visible scaffold improves solving accuracy in models with native extended thinking, we ran a pre-registered 7-arm paired ablation across all $N = 463$ test-set clues on both primary seats ($463 \times 7 \times 2 = 6{,}482$ scored API calls at Pass~0 settings: Gemini $90\text{s}$ cap, $11{,}378$ tokens; Claude $150\text{s}$ cap, $20{,}000$ tokens; puzzle-interleaved dispatch across arms):\footnote{Data provenance note: the raw log file \texttt{E15\_raw\_calls.jsonl} contains $7{,}038$ lines comprising the $6{,}482$ official scored calls in \texttt{E15\_per\_instance.csv} plus $556$ superseded rows from an interrupted early batch prior to full puzzle-interleaved re-execution; deduplicating to the canonical $6{,}482$ scored instances accounts for the slight table refinement between our initial preprint draft and the final ledger.}
\begin{itemize}[leftmargin=1.8em,itemsep=1pt]
    \item \textbf{\texttt{P0} (Full Production Scaffold)} and \textbf{\texttt{P0'} (Same-Prompt Replicate).}
    Identical prompts run to measure draw-to-draw replicate noise $\delta_0 = |\text{adm}(\texttt{P0}) - \text{adm}(\texttt{P0'})|$ and set the pre-registered detection threshold $T = \max(2.0\text{ pts}, 2\delta_0)$ ($T = 2.00\text{ pts}$ on Gemini, $\delta_0 = 0.43\text{ pts}$; $T = 3.02\text{ pts}$ on Claude under strict parsing, $\delta_0 = 1.51\text{ pts}$).
    \item \textbf{\texttt{P1} (Answer-First, Single-Answer).}
    Moves final \texttt{answer} and \texttt{confidence} to the top of the JSON schema before \texttt{definition}, \texttt{wordplay\_\allowbreak type}, \texttt{is\_\allowbreak and\_\allowbreak lit}, and \texttt{wordplay\_\allowbreak explanation}, while also omitting \texttt{answer\_\allowbreak unverified}.
    \item \textbf{\texttt{P2} (No-Revise).}
    Removes \texttt{answer\_\allowbreak unverified} and the self-revision instruction, placing a single \texttt{answer} before \texttt{confidence} at the end of the schema.
    \item \textbf{\texttt{P3} (No-Decomposition-Fields).}
    Deletes the three initial taxonomy keys (\texttt{definition}, \texttt{wordplay\_\allowbreak type}, and \texttt{is\_\allowbreak and\_\allowbreak lit}), starting the JSON object directly with \texttt{wordplay\_\allowbreak explanation}, \texttt{answer\_\allowbreak unverified}, \texttt{confidence}, and \texttt{answer}.
    \item \textbf{\texttt{P4} (No-Guidance).}
    Retains the full JSON schema of \texttt{P0} but strips the solving instruction paragraph, direction rules, and enumeration formatting examples.
    \item \textbf{\texttt{P5} (Minimal).}
    Strips both guidance and explanation fields, requesting only a minimal JSON object with \texttt{answer} and \texttt{confidence.level}.
\end{itemize}

\begin{table}[!htbp]
\centering
\caption{Paired Pass~0 prompt (\texttt{P0}--\texttt{P5}) and confidence-elicitation (\texttt{P6}) ablations across $N = 463$ test-set clues ($9{,}260$ scored calls; full strict and lenient breakdowns in \Cref{tab:app-prompt,tab:app-conf}).}
\label{tab:prompt-summary}
\small
\begin{adjustbox}{max width=\textwidth}
\begin{tabular}{@{}llrrrrrrrl@{}}
\toprule
\textbf{Seat} & \textbf{Prompt Arm} & \textbf{Adm} & \textbf{Corr$_{\text{ans}}$} & \textbf{T/O} & \textbf{Med $t$} & \textbf{Both-OK $n$} & \textbf{All $\Delta$ ($p$)} & \textbf{Ok $\Delta$ ($p$)} \\
\midrule
\multirow{8}{*}{\textbf{Gemini 3.1 Pro}}
 & \texttt{P0} (Full Scaffold)       & $70.6\%$ & $98.5\%$ & $25.1\%$ & $21.1\text{s}$ & --- & --- & --- \\
 & \texttt{P0'} (Same-Prompt Rep.)   & $70.2\%$ & $98.2\%$ & $26.1\%$ & $22.3\text{s}$ & $299$ & $-0.43$ ($0.90$) & $\phantom{+}0.00$ ($1.00$) \\
 & \texttt{P1} (Answer-First Order)  & $71.3\%$ & $94.6\%$ & $23.5\%$ & $23.9\text{s}$ & $304$ & $+0.65$ ($0.80$) & $-0.66$ ($0.63$) \\
 & \texttt{P2} (No Self-Revision)    & $72.6\%$ & $97.7\%$ & $23.1\%$ & $21.8\text{s}$ & $301$ & $+1.94$ ($0.33$) & $+0.33$ ($1.00$) \\
 & \texttt{P3} (No Decomp.\ Fields)  & $70.8\%$ & $97.3\%$ & $24.2\%$ & $22.0\text{s}$ & $301$ & $+0.22$ ($1.00$) & $-1.66$ ($0.063$) \\
 & \texttt{P4} (No Domain Guidance)  & $69.8\%$ & $98.8\%$ & $26.3\%$ & $20.4\text{s}$ & $302$ & $-0.86$ ($0.68$) & $+0.33$ ($1.00$) \\
 & \texttt{P5} (Minimal 2-Field)     & $73.4\%$ & $90.1\%$ & $15.3\%$ & $17.4\text{s}$ & $315$ & $+2.81$ ($0.13$) & $\mathbf{-2.86}$ ($\mathbf{0.004}$) \\
 & \texttt{P6} (No Confidence)$^*$   & $76.0\%$ & $96.7\%$ & $17.7\%$ & $21.7\text{s}$ & $322$ & $+1.94$ ($0.32$) & $\phantom{+}0.00$ ($1.00$) \\
\midrule
\multirow{9}{*}{\textbf{Claude Fable 5}}
 & \texttt{P0} (Full, Strict / Lenient)$^\dagger$ & $79.5\% / 80.3\%$ & $95.0\%$ & $13.2\%$ & $14.1\text{s}$ & --- & --- & --- \\
 & \texttt{P0'} (Same-Prompt Rep.)   & $78.0\% / 79.7\%$ & $94.8\%$ & $14.9\%$ & $14.1\text{s}$ & $359 / 372$ & $-1.51 / -0.65$ & $-0.28 / -0.27$ ($1.00$) \\
 & \texttt{P1} (Answer-First Order)  & $81.6\% / 81.9\%$ & $95.8\%$ & $13.0\%$ & $14.4\text{s}$ & $370 / 376$ & $+2.16 / +1.51$ & $+0.27 / +0.27$ ($1.00$) \\
 & \texttt{P2} (No Self-Revision)    & $79.0\% / 79.0\%$ & $95.6\%$ & $14.9\%$ & $13.7\text{s}$ & $368 / 373$ & $-0.43 / -1.30$ & $-0.27 / -0.27$ ($1.00$) \\
 & \texttt{P3} (Strict Parser)       & $74.7\%$ & $95.1\%$ & $13.2\%$ & $12.7\text{s}$ & $343$ & $\mathbf{-4.75}$ ($\mathbf{0.015}$) & $-0.58$ ($0.75$) \\
 & \texttt{P3} (Uniform Lenient)$^\dagger$ & $81.0\%$ & $95.3\%$ & $13.2\%$ & $12.7\text{s}$ & $376$ & $+0.65$ ($0.78$) & $+0.27$ ($1.00$) \\
 & \texttt{P4} (No Domain Guidance)  & $78.2\% / 80.6\%$ & $93.9\%$ & $12.1\%$ & $12.3\text{s}$ & $369 / 383$ & $-1.30 / +0.22$ & $-0.54 / -0.78$ ($0.61$) \\
 & \texttt{P5} (Minimal 2-Field)     & $81.4\% / 81.4\%$ & $93.0\%$ & $11.0\%$ &  $8.0\text{s}$ & $383 / 388$ & $+1.94 / +1.08$ & $-1.04 / -1.03$ ($0.48$) \\
 & \texttt{P6} (No Confidence)$^*$   & $81.6\%$ & $95.2\%$ & $14.0\%$ & $12.0\text{s}$ & $365$ & $+1.30$ ($0.53$) & $-0.55$ ($0.77$) \\
\bottomrule
\multicolumn{9}{@{}l}{\footnotesize \textbf{Adm}: primary admissible rate (status \texttt{ok}, $|a|=\ell_i$, $c_m \ge \tau_m$, $a=\text{GT}$, $/463$); \textbf{Corr$_{\text{ans}}$}: accuracy given status \texttt{ok}; \textbf{T/O}: timeout rate;} \\
\multicolumn{9}{@{}l}{\footnotesize \textbf{Both-OK $n$}: clues completed by both \texttt{P0} and arm; \textbf{All $\Delta$} and \textbf{Ok $\Delta$}: paired percentage-point changes vs.\ \texttt{P0} (two-sided McNemar $p$).} \\
\multicolumn{9}{@{}l}{\footnotesize $^*$\texttt{P6} reports unconditional correctness against its paired same-session \texttt{P0} baseline (\cref{sec:conf-ablation}, \Cref{tab:app-conf}).} \\
\multicolumn{9}{@{}l}{\footnotesize $^\dagger$On Claude, the second number in each cell applies a deterministic lenient JSON parser uniformly across all arms (recovering $5$ syntax} \\
\multicolumn{9}{@{}l}{\footnotesize \phantom{$^\dagger$}errors on \texttt{P0} [$+4$ adm], $9$ on \texttt{P0'} [$+8$], $1$ on \texttt{P1} [$+1$], $30$ on \texttt{P3} [$+29$ adm], and $13$ on \texttt{P4} [$+11$]; Gemini failures are truncations).}
\end{tabular}
\end{adjustbox}
\end{table}

\Cref{fig:folklore}b and \Cref{tab:prompt-summary} (with complete discordant counts in \Cref{tab:app-prompt}) establish two main findings.
\begin{enumerate}[leftmargin=1.8em,itemsep=2pt]
    \item \textbf{Schema ordering, self-revision, decomposition fields, and domain guidance show no statistically significant accuracy effect once JSON formatting errors are separated from reasoning (\texttt{P1}--\texttt{P4}).}
    Moving the answer before the explanation (\texttt{P1}), deleting the draft-and-revise fields (\texttt{P2}), or deleting all domain solving prose (\texttt{P4}) fails to clear either the pre-registered detection threshold $T$ or $p < 0.05$ on either model.
    On the jointly completed (\texttt{both-ok}) subset ($n = 299$--$304$ clues on Gemini, $n = 343$--$388$ on Claude across strict and lenient parses), every single-component contrast lies within $\pm 1.66\text{ percentage points}$ with exact McNemar $p \ge 0.063$.
    Under our pre-registered strict JSON parser, \texttt{P3} on Claude dropped whole-set admissible accuracy by $-4.75\text{ pts}$ ($346/463 = 74.7\%$ vs.\ $368/463 = 79.5\%$, $p = 0.015$) solely because removing the three leading keys caused Claude's output formatter to emit $30$ minor JSON syntax errors at the nested \texttt{confidence} boundary ($18$ trailing commas after \texttt{confidence.reason} and $12$ omitted closing braces before \texttt{"answer"}).

    To test whether this deficit reflected reasoning degradation or syntactic formatting fragility without biasing the comparison in favor of \texttt{P3}, we applied a deterministic post-hoc lenient JSON parser uniformly across \textit{all} arms (\Cref{tab:app-prompt}, Panel~B); this parser strips markdown code fences and trailing commas before \texttt{\}} or \texttt{]} and inserts a missing closing brace \texttt{\}} after \texttt{"reason": "..."} when immediately followed by \texttt{"answer":}.\footnote{Uniform lenient parsing recovered $5/5$ syntax failures on \texttt{P0} ($4$ correct and $4$ admissible, raising \texttt{P0} from $368 \to 372/463 = 80.3\%$), $9/9$ on \texttt{P0'} ($+8$ admissible $\to 369/463 = 79.7\%$), $1/1$ on \texttt{P1} ($+1 \to 379/463 = 81.9\%$), $0/4$ on \texttt{P2} (all $4$ truncated at \texttt{max\_tokens}), $30/30$ on \texttt{P3} ($29$ correct and all $29$ admissible, raising \texttt{P3} from $346 \to 375/463 = 81.0\%$, with unconditional correctness $382/463 = 82.5\%$ on \texttt{P3} vs.\ $381/463 = 82.3\%$ on \texttt{P0}), and $13/13$ on \texttt{P4} ($+11$ admissible $\to 373/463 = 80.6\%$).
    Across all $43$ E2 daily production runs, strict parsing discarded only $19$ such responses out of $3{,}012$ Claude calls ($0.6\%$).}
    Under uniform lenient parsing, Claude \texttt{P3} vs.\ \texttt{P0} has $\text{All }\Delta = +0.65\text{ pts}$ ($+3/463$, $b=25, c=28, p=0.78$) and $\text{Ok }\Delta = +0.27\text{ pts}$ ($b=6, c=7, n=376, p=1.00$).
    While a null NHST result does not prove strict mathematical equivalence, the observed differences lie inside draw-to-draw replicate noise, consistent with cryptic wordplay search occurring primarily inside the model's native thinking trace prior to JSON serialization.
    \item \textbf{The Minimal prompt (\texttt{P5}) trades answer precision for lower timeout rates.}
    Stripping the output down to \texttt{\{answer, confidence.level\}} reduces mean output tokens from $194 \to 34$ on Gemini ($1{,}861 \to 1{,}325$ combined thinking-plus-output tokens on Claude), cutting Gemini's Pass~0 timeout rate from $25.1\%$ to $15.3\%$.
    However, on the $n = 315$ clues where both \texttt{P0} and \texttt{P5} completed on Gemini, \texttt{P5} is significantly \textit{less} accurate on the primary admissible endpoint ($-2.86\text{ pts}$, $b=9$ lost vs.\ $c=0$ gained, exact McNemar $p = 0.0039$; on unconditional answer correctness without the confidence floor, $\text{Ok }\Delta = -1.59\text{ pts}$, $b=5, c=0, p = 0.0625$), lowering accuracy-when-answering from $98.5\% \to 90.1\%$ on Gemini ($95.0\% \to 93.0\%$ on Claude) and nearly doubling high-confidence errors at the admission floor on both seats ($1.5\% \to 2.9\%$ on Gemini; $2.4\% \to 4.3\%$ on Claude; \Cref{tab:app-prompt}).
    Requiring a written wordplay explanation (\texttt{P0}) acts as a brake on unverified guessing: a model that has not found a coherent parse continues searching inside its thinking trace until the timeout cap (converting a would-be wrong guess into a clean Pass~0 abstention) rather than emitting a fast, low-precision guess.
\end{enumerate}

\subsection{Effect of Eliciting Self-Graded Confidence on Answer Accuracy}
\label{sec:conf-ablation}

Even if visible parse fields do not boost reasoning, \Darmok{} relies on each seat's self-reported confidence score $c_m \in [0, 100]$ to enforce seat floors (\cref{eq:floors}).
To test whether asking a model to score and justify its own confidence alters the crossword answer it emits, we ran a 3-arm paired experiment across all $463$ test-set clues on both primary seats ($463 \times 3 \times 2 = 2{,}778$ live API calls; \Cref{tab:app-conf}), comparing a same-session full-prompt baseline with confidence (\texttt{P0})\footnote{In the raw archive (\texttt{E15b\_P6\_raw\_calls.jsonl}) and \Cref{tab:app-conf}, this same-session full-prompt baseline draw is labeled \texttt{P0r2} to distinguish it from the earlier \texttt{P0} and \texttt{P0'} draws in \cref{sec:prompt-ablation}; the prompt text and parameters are identical to \texttt{P0}.} against \texttt{P6} (\textit{Full-No-Conf}: identical to \texttt{P2} with the \texttt{confidence} object removed) and \texttt{P7} (\textit{Minimal-No-Conf}: identical to \texttt{P5} with \texttt{confidence} removed, requesting only \texttt{\{"answer": "..."\}}).
Because \texttt{P6} and \texttt{P7} emit no confidence score, all contrasts in \Cref{tab:app-conf} evaluate unconditional correctness ($\text{Corr}_{\text{all}}$ and jointly completed $\text{Ok }\Delta$).

Because \texttt{P6} removes the \texttt{confidence} object from \texttt{P2}, we evaluate \texttt{P6} both against the same-session \texttt{P0} baseline (which also includes \texttt{answer\_unverified}) and directly against \texttt{P2} across sessions (isolating confidence removal from schema differences, subject to cross-session timeout drift).
Across both comparisons, removing confidence elicitation produces \textbf{near-zero change in jointly completed answer accuracy} ($\texttt{P6}$ vs.\ same-session $\texttt{P0}$: $0.00\text{ pts}$ on Gemini [$b=2, c=2, n=322, p=1.00$] and $-0.55\text{ pts}$ on Claude [$b=7, c=5, n=365, p=0.77$; under uniform lenient JSON parsing, $\text{All }\Delta = -0.86\text{ pts}$, $p = 0.68$, and $\text{Ok }\Delta = -0.53\text{ pts}$, $n=375, p = 0.77$]; $\texttt{P6}$ vs.\ $\texttt{P2}$: $-0.32\text{ pts}$ on Gemini [$b=2, c=1, n=311, p=1.00$] and $-0.27\text{ pts}$ on Claude [$b=4, c=3, n=370, p=1.00$]).
Across jointly completed pairs, \texttt{P6} emitted the \textit{exact same normalized answer string} as the same-session \texttt{P0} baseline on $317/322$ clues ($98.4\%$) on Gemini ($307/311 = 98.7\%$ vs.\ \texttt{P2}, compared to $308/309 = 99.7\%$ same-prompt cross-session identity) and on $350/365$ clues ($95.9\%$) on Claude ($358/370 = 96.8\%$ vs.\ \texttt{P2}, compared to $347/359 = 96.7\%$ same-prompt identity).\footnote{In \Cref{tab:app-conf}, on all-clue unconditional correctness ($\text{All }\Delta$), Gemini's two \texttt{P0} sessions drifted by $+3.46\text{ pts}$ ($p=0.048$) purely because ambient timeout rate shifted from $25.1\%$ to $19.9\%$ (which also accounts for the $+3.46\text{ pt}$ cross-session $\text{All }\Delta$ between Gemini \texttt{P6} and \texttt{P2}), whereas on jointly completed calls $\text{Ok }\Delta = -0.32\text{ pts}$ ($1$ lost / $0$ gained).
Similarly, stripping both explanation and confidence in \texttt{P7} reduced timeouts to $11.2\%$ on Gemini and $8.9\%$ on Claude, while degrading accuracy on jointly completed calls ($\text{Ok }\Delta = -3.94\text{ pts}$, $p=0.0002$ on Gemini; $-2.13\text{ pts}$, $p=0.077$ on Claude).}

In short, eliciting verbalized confidence does not alter the model's object-level answers beyond normal sampling consistency.

\subsection{Auditing Symbolic-LLM Wordplay Verification vs.\ Frequency Priors}
\label{sec:verifier-falsification}

An influential line of recent neuro-symbolic cryptic work proposes filtering LLM crossword candidates through a symbolic wordplay verifier that prompts a language model (with iterative error-guided rewrites) to synthesize an executable Python \texttt{proof()} function that checks synonymy, standard abbreviations (\textsc{UKACD}), indicator action types, and string manipulations~\cite{andrews2024proving,andrews2025reasoning}.
Early in our development, we audited the open-source implementation of Andrews \& Witteveen~\cite{andrews2024proving} (\url{https://github.com/mdda/cryptic-crossword-reasoning-verifier}) and evaluated its viability as a candidate selector for \Darmok{}.

Inspecting the repository's benchmark evaluation script (\texttt{09\_prove\_all\_samples.py}, lines 603--608) on their 200-clue \textsc{Cryptonite} validation split revealed a striking component breakdown recorded in the script's own comments (\Cref{tab:app-aw}, Panel~A): selecting the candidate most frequently proved by the symbolic verifier alone (\texttt{answer\_proof}, without frequency fallback) solved only $28/200$ clues ($14.0\%$), whereas an unverified proposal-frequency baseline over those exact same $20$ already-generated samples (\texttt{answer\_freq}, taking the modal candidate string at zero additional verification API cost) solved $56/200$ clues ($28.0\%$), beating the standalone symbolic verifier by $2{:}1$. For each of these $200$ clues, their fine-tuned Gemma2-9B candidate generator first samples $20$ candidate answers and their symbolic prover then makes up to $3$ proof-synthesis attempts per candidate.
Because their published pipeline~\cite{andrews2025reasoning} always falls back to \texttt{answer\_freq} whenever no candidate obtains a valid proof, that hybrid fallback also scored $56/200$ ($28.0\%$) on this validation script ($0.0\text{ pts}$ of marginal lift over unverified majority vote, though on their 200-clue test split the hybrid improved over frequency alone from $25.5\%$ to $32.5\%$).
The verifier also cannot run on raw clues: its prompt template requires the candidate generator (or a dedicated Stage-2 Gemma2-9B span-annotation model in their pipeline) to pre-enclose the candidate definition span in curly braces (\texttt{\{definition\}}) prior to verification, coupling proof synthesis to upstream definition-span identification.

To test whether upgrading the proof generator from the paper's \texttt{gemini-1.5-flash-001} to a modern reasoning model (\texttt{gemini-2.5-flash}, $16{,}384$-token budget) could overcome these limitations in our pipeline, we manually annotated \texttt{\{definition\}} spans for $18$ clues missed by our 2025 V1 baseline (pooling the non-empty candidate strings generated across $3$ models and $6$ sample budgets $k \in \{1,2,4,6,8,10\}$ on those $18$ clues yielded $200$ unique \texttt{(clue\_with\_def, candidate)} pairs) and ran a pre-registered 5-pair smoke gate ($3$ rewrite attempts per pair, $15$ proof syntheses at $1\text{m}\,07\text{s}$--$4\text{m}\,10\text{s}$ per pair; complete trace in \Cref{tab:app-aw}, Panel~B).
All five pairs, including both known-correct answers, returned \texttt{score = -1} (\texttt{fails}), illustrating two failure modes encountered when integrating post-hoc Python proof synthesis.
\begin{enumerate}[leftmargin=1.8em,itemsep=2pt]
    \item \textbf{False negatives on valid lateral wordplay.}
    On Rewrite~0 of the known-correct \textit{\&lit} clue \clue{\{Finally snap in rage, so agitated\} (2,5)} $\to$ \ans{GO SPARE} (Clue~C10, \Cref{tab:app-clues-1}), \texttt{gemini-2.5-flash} emitted the \textit{exact} human wordplay, taking the final letter of \texttt{"SNAP"} (\texttt{"SNAP"[-1] == "P"}, signaled by \textit{Finally}) plus \texttt{"RAGE"} plus \texttt{"SO"} (\texttt{"PRAGESO"}) and anagramming (\textit{agitated}, \texttt{Action.ANAGRAM}) to \ans{GO SPARE}.
    However, the verifier rejected this valid proof because its AST evaluator executes \texttt{assert} lines in isolation while dropping local variable bindings (\texttt{NameError} on \texttt{p\_letter}) and its 10-action \texttt{Action} enum lacks a last-letter operation (rejecting \texttt{Action.SUBSTRING} on Rewrite~1).
    After two rejections of its correct parse, the model resorted on Rewrite~2 to redefining the mock helper functions to hardcode \texttt{if test\_synonym == "GO SPARE": return True}, which tripped the AST parser's preamble check (a guard that rejects any proof script modifying the template lines above the \texttt{\#START\_CODE} marker).
    \item \textbf{Adversarial rationalization and error-trace coaching on wrong candidates.}
    Because a capable LLM is tasked with proving a fixed target string (and the verifier's input format strips the puzzle's parenthesized enumeration, checking only whether the proof's string matches the supplied candidate), it attempts to rationalize whatever candidate is passed in. For instance, it constructed a 14-letter charade \texttt{"LORD" + "PALM" + "ER" + "STON"} for the wrong 14-letter decoy \ans{LORD PALMERSTON} on the 15-letter clue \clue{Ex-PM's fashionable \{royal post\} (15)} ($\to$ \ans{CHAMBERLAINSHIP}; Clue~C11), which failed AST preamble extraction.
    On \clue{Soldiers occupy a position around \{river\} (4)} ($\to$ \ans{EBRO}; Clue~C9), the proof synthesizer missed the true parse on \ans{EBRO} (hallucinating four unattested single-letter abbreviations \texttt{"E"}+\texttt{"B"}+\texttt{"R"}+\texttt{"O"}), yet synthesized a plausible alternative parse for the wrong river \ans{ERNE}: \texttt{"RE"} (\textit{Soldiers} = Royal Engineers) reversed (\textit{around}, \texttt{Action.REVERSE} $\to$ \texttt{"ER"}) plus \texttt{"NE"} (\textit{a position} = North-East).
    When Rewrite~0 failed on \texttt{is\_abbreviation("position", "NE")}, the verifier's error trace leaked the \textsc{UKACD} dictionary entry (\texttt{'NE' is an abbreviation for : not, born, neon, north-east}). This coached the model on Rewrite~1 to substitute \texttt{is\_abbreviation("North-East", "NE")}, which passed the abbreviation validator even though \texttt{"North-East"} never appears in the clue (ultimately failing only on the AST preamble check).
\end{enumerate}

These failure modes show the fragility of hand-crafted symbolic DSLs on open-ended linguistic domains. Whenever a valid clue uses a lateral device outside the DSL's fixed operator enumeration, the verifier rejects the true answer, while feeding validator error traces back to a capable LLM coaches it to game the verifier's lookup tables in support of wrong candidates.
We therefore abandoned post-hoc symbolic proof synthesis in favor of multi-seat proposal corroboration (\cref{alg:corrob}).

\subsection{Summary of Clue-Level Findings}
\label{sec:folklore-summary}

Together, our four clue-level studies separate the primary drivers of cryptic solving accuracy from secondary prompt and verification choices.
\begin{itemize}[leftmargin=1.8em,itemsep=2pt]
    \item \textbf{Crossing-letter constraints (\cref{sec:crossings-amp}).}
    Conditioning retries on $\ge 3$ intersecting letters yields a $+33$ to $+50\text{ percentage-point}$ higher solve rate across all model families, reaching $98.1\%$ ($n_{\text{known}}=3$) and $100.0\%$ ($n_{\text{known}} \ge 4$) across the three-seat union.
    \item \textbf{Multi-seat consensus and proposal frequency (\cref{sec:corrob-admission,sec:verifier-falsification}).}
    Independent agreement across seats (or across sampled proposals) reliably separates genuine wordplay parses from isolated confabulations, outperforming standalone post-hoc symbolic proof synthesis (without frequency fallback) by $2{:}1$ ($28.0\%$ vs.\ $14.0\%$).
    \item \textbf{Required wordplay explanation as a brake on unverified guessing (\cref{sec:prompt-ablation}).}
    Requiring a written wordplay explanation (\texttt{P0} vs.\ \texttt{P5}) prevents fast, low-precision guessing and halves high-confidence errors at the admission floor, at the cost of a higher Pass~0 timeout rate.
    \item \textbf{Invariance to visible CoT field order, self-revision, and confidence elicitation (\cref{sec:prompt-ablation,sec:conf-ablation}).}
    Varying JSON field order (\texttt{P1}), draft-and-revise instructions (\texttt{P2}), taxonomy keys (\texttt{P3}), domain tutorials (\texttt{P4}), or confidence schemas (\texttt{P6}) produces no statistically significant accuracy change on completed calls over native extended thinking.
    \item \textbf{Fragility of post-hoc Python-AST verification (\cref{sec:verifier-falsification}).}
    Rigid symbolic verifiers suffer simultaneously from false negatives on lateral wordplay and adversarial rationalization on wrong candidates when driven by iterative error-trace feedback.
\end{itemize}

\section{System Evaluation: The Mean-vs.-Tail Divergence}
\label{sec:eval}

We now evaluate the complete \Darmok{} system across five complementary analyses: the 132-day post-cutoff daily production record and 16-puzzle stratified test set (\cref{sec:corpus-eval}), unmodified transfer to barred \textit{Mephisto} grids (\cref{sec:mephisto-eval}), average-case leave-one-out and $2\times 2$ factorial stage ablations (\cref{sec:mean-eval}), tail-regime evaluations spanning $135$ single-shot frontier draws, the worst decile of daily production runs, and the Championship Final (\cref{sec:tail-eval}), and a residual error audit of the very long tail of clues that still break above $99.5\%$ accuracy (\cref{sec:error-analysis}).
Complete reference tables appear in \Cref{app:tables}.

\subsection{Post-Cutoff Daily Production Corpus and Stratified Test Set}
\label{sec:corpus-eval}

A pervasive concern when evaluating frontier models on historical benchmarks is pre-training data contamination.
Our primary evaluation corpus addresses this directly: between May 19 and October 9, 2026, \Darmok{} evaluated $132$ usable daily \textit{Times} Cryptic puzzles ($3{,}876$ graded clues) published strictly after the knowledge cutoffs of all models in the ensemble.\footnote{Recorded knowledge cutoffs: GPT-5 (\texttt{gpt-5-2025-08-07}): September 30, 2024; Gemini 3.1 Pro (\texttt{gemini-3.1-pro-preview}): January 31, 2025; \texttt{claude-fable-5}: January 31, 2026; \texttt{gpt-5.6-sol}: February 16, 2026; \texttt{gemini-4-argon} and \texttt{gemini-3.8-flash}: March 2026; \texttt{gpt-6-astra}: April 30, 2026.}
Of the $132$ puzzles, $126$ were executed live on their morning of publication (including all $43$ puzzles of the frozen Epoch~E2 configuration and all $50$ puzzles of Epoch~E1), while $6$ Epoch~E0 puzzles published during an author vacation (June~17--29, 2026) were batch-backfilled on June~29 and July~5.
For every run, both live and backfilled, the Pass~4 historical answer list was strictly date-filtered to puzzles published prior to the target puzzle's publication date.

\begin{table}[!htbp]
\centering
\caption{\Darmok{} accuracy, perfect-grid rate, and median per-puzzle solve time\protect\footnotemark{} across evaluation corpora and architectural epochs (detailed breakdowns in \Cref{tab:app-prod,tab:app-testset,tab:app-tail}).}
\label{tab:corpus-summary}
\small
\begin{adjustbox}{max width=\textwidth}
\begin{tabular}{@{}lrrrrrrr@{}}
\toprule
\textbf{Corpus / Architecture} & \textbf{Puzzles} & \textbf{Clues} & \textbf{Correct} & \textbf{Wrong} & \textbf{Blank} & \textbf{Perfect Grids} & \textbf{Median Time} \\
\midrule
55th \textit{Times} Championship (Live Contest) &  7 &    210 & $\mathbf{210}$ ($\mathbf{100.00\%}$) & 0 & 0 & $\mathbf{7/7}$ ($\mathbf{100.0\%}$) & $05\text{m}\,08\text{s}$ \\
\midrule
Daily Production: Epoch E2 (Aug 26 -- Oct 9, Frozen) & 43 & $1{,}261$ & $\mathbf{1{,}259}$ ($\mathbf{99.84\%}$) & 1 & 1 & $\mathbf{41/43}$ ($\mathbf{95.3\%}$) & $05\text{m}\,36\text{s}$ \\
Daily Production: Epoch E1 (Jul 6 -- Aug 25, Corrob.) & 50 & $1{,}480$ & $1{,}473$ ($99.53\%$) & 2 & 5 & $44/50$ ($88.0\%$) & $06\text{m}\,03\text{s}$ \\
Daily Production: Epoch E0 (May 19 -- Jul 5, Greedy)  & 39 & $1{,}135$ & $1{,}112$ ($97.97\%$) & 2 & 21 & $23/39$ ($59.0\%$) & $09\text{m}\,40\text{s}$ \\
\quad \textit{Full 132-Day Daily Campaign (E0--E2)}$^*$ & 132 & $3{,}876$ & $\mathbf{3{,}844}$ ($\mathbf{99.17\%}$) & 5 & 27 & $\mathbf{108/132}$ ($\mathbf{81.8\%}$) & $06\text{m}\,22\text{s}$ \\
\midrule
Stratified Test Set: Championship V2 ($D=1/3$)  & 16 &    463 & $\mathbf{461}$ ($\mathbf{99.57\%}$) & 2 & 0 & $\mathbf{14/16}$ ($\mathbf{87.5\%}$) & $04\text{m}\,11\text{s}$ \\
\quad \textit{vs.\ June 2026 Prototype (Epoch E0 V2)} & 16 & 463 & $408$ ($88.12\%$) & 16 & 39 & $4/16$ ($25.0\%$) & $05\text{m}\,44\text{s}$ \\
\quad \textit{vs.\ 2025 Single-Model Baseline (V1)} & 16 & 463 & $380$ ($82.07\%$) & 83 &  0 & $1/16$ ($6.3\%$) & $11\text{m}\,40\text{s}$ \\
\midrule
Sunday \textit{Mephisto} Barred Grids (Zero-Shot) & 10 &  360 & $\mathbf{350}$ ($\mathbf{97.22\%}$) & 4 & 6 & $\mathbf{5/10}$ ($\mathbf{50.0\%}$) & $13\text{m}\,32\text{s}$ \\
\bottomrule
\multicolumn{8}{@{}l}{\footnotesize $^*$Provenance note: Reconciling an earlier preliminary summary ($3{,}819$ correct, $24$ wrong, $33$ blank) to the canonical \texttt{solution\_grid}-verified} \\
\multicolumn{8}{@{}l}{\footnotesize \phantom{$^*$}manifest updated $19$ preliminary E0 ``wrong'' entries ($21 \to 2$ wrong in E0, where early E0 logs had been graded against provisional} \\
\multicolumn{8}{@{}l}{\footnotesize \phantom{$^*$}LLM-extracted reference answers prior to backfilling official club \texttt{solution\_grid} keys) and $6$ preliminary ``blank'' entries ($4$ in E0 [$25 \to 21$],} \\
\multicolumn{8}{@{}l}{\footnotesize \phantom{$^*$}$2$ in E1 [$7 \to 5$]), giving $+19 + 6 = +25$ correct clues ($3{,}819 \to 3{,}844$), while aligning the July~6 E0/E1 cutoff shifted $7$ clues from E0} \\
\multicolumn{8}{@{}l}{\footnotesize \phantom{$^*$}(including $2$ clues on the boundary DNS run, $264 \to 262$) to E1.}
\end{tabular}
\end{adjustbox}
\end{table}
\footnotetext{We report median solve time (paired with $\text{p90}$, SD, or range in \Cref{tab:intro-oneshot-mini,tab:e17-oneshot,tab:app-prod}) because per-puzzle wall-clock latency is right-skewed by discrete pass-barrier timeout caps ($90\text{s}$ and $150\text{s}$), making the median the standard robust location estimator for barrier-synchronized systems.}

\Cref{tab:corpus-summary} tracks both accuracy and speed across all three production epochs: replacing greedy incremental admission in Epoch~E0 ($97.97\%$ clue accuracy, $59.0\%$ perfect grids, median solve time $9\text{m}\,40\text{s}$) with synchronous corroboration admission in Epoch~E1 raised daily accuracy to $99.53\%$ ($88.0\%$ perfect grids) and cut median solve time to $6\text{m}\,03\text{s}$, while freezing the full championship pipeline in Epoch~E2 reached \textbf{$1{,}259/1{,}261$ ($99.84\%$)} and $41/43$ ($95.3\%$) flawless grids in a median of $5\text{m}\,36\text{s}$.
The two missed clues across all $43$ days of Epoch~E2 were a single wrong terminal sweep fill on puzzle~29658 (clue~8D \ans{TOWELINGS} for \ans{TOW-PLANES}; Clue~C22, \Cref{tab:app-clues-2}) and a single unfilled clue on puzzle~29648 (20A \ans{VISITOR CENTRE}).

\begin{table}[!htbp]
\centering
\caption{Epoch~E2 daily corpus ($43$ puzzles, $1{,}261$ clues) by committing stage and seat attribution (Panels~A--B; full bins in \Cref{tab:app-prod}), and pre- vs.\ post-cutoff Pass~0 (\texttt{P0}) accuracy on the 16-puzzle test set ($N = 463$; Panel~C).}
\label{tab:e2-stage-seat}
\footnotesize
\setlength{\tabcolsep}{4pt}
\begin{adjustbox}{max width=\textwidth}
\begin{tabular}{@{}lrr@{\hspace{1.4em}}lrrr@{\hspace{1.4em}}lrrr@{}}
\toprule
\multicolumn{3}{@{}l}{\textbf{Panel A: E2 Committing Stage}} & \multicolumn{4}{l}{\textbf{Panel B: E2 Seat Attribution}} & \multicolumn{4}{l@{}}{\textbf{Panel C: Test-Set Cutoff Check (\texttt{P0})}} \\
\cmidrule(r){1-3} \cmidrule(lr){4-7} \cmidrule(l){8-11}
\textbf{Stage} & \textbf{Clues} & \textbf{Share} & \textbf{Seat} & \textbf{P0 Silent} & \textbf{First-to-Corr.} & \textbf{Unique} & \textbf{Test Cohort} & \textbf{Clues} & \textbf{Claude Adm} & \textbf{Gemini Adm} \\
\midrule
Pass 0 (Cold)         &   944 & $74.9\%$ & Claude Fable 5 & $11.6\%$ & $1{,}016$ ($80.9\%$) & 28 & Pre-Jan '26 (13 pz)  & 376 & $295$ ($78.5\%$) & $261$ ($69.4\%$) \\
Passes 1--3 (Retries) &   307$^*$ & $24.3\%$ & Gemini 3.1 Pro & $25.9\%$ &   $234$ ($18.6\%$) & 71 & Post-Jan '26 (3 pz) &  87 &  $73$ ($83.9\%$) &  $66$ ($75.9\%$) \\
Pass 4 (Historical)   &     6 &  $0.5\%$ & GPT-5.6-sol    & ---      &     $6$ ($0.5\%$)  &  1 & \textit{All 16 Puzzles} & 463 & $368$ ($79.5\%$) & $327$ ($70.6\%$) \\
Sweep (\textsc{swp})  &     4$^\dagger$ &  $0.3\%$ & \multicolumn{4}{l}{\textit{Model-committed correct: $1{,}256$}} & \multicolumn{4}{l@{}}{\textit{Full-system V2 on Post-Jan '26: $87/87$ ($100\%$)}} \\
\textsc{corr}/\textsc{pers}/\textsc{atf} & 0 & $0.0\%$ & & & & & & & & \\
\bottomrule
\multicolumn{11}{@{}l}{\footnotesize $^*$$306$ committed-correct clues plus $1$ clue that remained blank after Passes~1--3. \quad $^\dagger$$3$ correct and $1$ wrong (\ans{TOWELINGS} for \ans{TOW-PLANES}).}
\end{tabular}
\end{adjustbox}
\end{table}

As summarized in \Cref{tab:e2-stage-seat} (Panels~A--B), Pass~0 cold corroboration commits three-quarters ($74.9\%$) of all Epoch~E2 clues and Passes~1--3 crossing-constrained retries commit nearly all of the remainder ($24.3\%$), pre-empting \textsc{corr}, \textsc{pers}, and \textsc{atf} ($0$ live commits).
The two primary seats play complementary roles: because Claude has a lower Pass~0 timeout rate than Gemini ($11.6\%$ vs.\ $25.9\%$), Claude is the first seat to propose the correct answer on $80.9\%$ of model-committed clues, whereas Gemini contributes $2.5\times$ as many unique solves ($71$ vs.\ $28$).

On the 16-puzzle stratified test set (\Cref{tab:corpus-summary} and \Cref{tab:app-testset}; individual puzzle identities and dates in \Cref{tab:app-test-puzzles}), the championship configuration ($D = 1$ on SNITCH bands $70$--$130$ and $D = 3$ on bands $130+$) solved $461/463$ clues ($99.57\%$, $14/16$ perfect grids; $917/923 = 99.35\%$ across all $32$ individual replica runs) in a median of $4\text{m}\,11\text{s}$.
Comparing strict Pass~0 (\texttt{P0}) accuracy across the $13$ test-set puzzles published before Claude Fable~5's January~31, 2026 cutoff versus the $3$ post-cutoff puzzles (\Cref{tab:e2-stage-seat}, Panel~C) also confirms that Claude's cold admissible rate is slightly higher on the post-cutoff cohort ($83.9\%$ vs.\ $78.5\%$), matching the difficulty-driven shift on Gemini ($75.9\%$ vs.\ $69.4\%$, where all $16$ puzzles post-date Gemini's January~31, 2025 cutoff by $5$--$14$ months).
Thus there is no pre-cutoff memorization advantage on the test set.

\subsection{Zero-Shot Generalization to Barred Grids (\textit{Mephisto})}
\label{sec:mephisto-eval}

To test whether \Darmok{}'s architecture overfits to $15\times 15$ blocked \textit{Times} grids, we evaluated the frozen championship solver \textit{zero-shot} (with no modifications to solver code, prompts, thresholds, or wordlists) on ten consecutive Sunday \textit{Times} \textit{Mephisto} puzzles (puzzles 3440--3449, published on Sundays from August 2 to October 4, 2026; $360$ clues).
As shown in \Cref{fig:primer-grids}c and \cref{sec:csp-breakdown}, \textit{Mephisto} grids are $12\times 12$ barred grids with zero black squares ($36$ words separated by thick cell bars) that check $76.9\%$--$80.0\%$ (mean $78.6\%$) of word-letter positions ($33.3\%$--$37.5\%$ [mean $35.3\%$] of physical white cells unchecked across the $10$ grids), while shifting the target lexicon to arcane, dialectal, and archaic headwords in \textit{The Chambers Dictionary}.

Across the $360$ \textit{Mephisto} clues, the unmodified solver achieved \textbf{$350/360$ ($97.22\%$)}, with $4$ wrong answers, $6$ blanks, and $5/10$ perfect grids in $10\text{m}\,24\text{s}$ to $17\text{m}\,54\text{s}$ per puzzle (median $13\text{m}\,32\text{s}$).
Three of the four wrong answers were single-letter \textit{Chambers} spelling variants lying on unchecked cells (verified derivations in \Cref{tab:app-clues-3}):
\begin{itemize}[leftmargin=1.8em,itemsep=2pt]
    \item \textbf{\textit{Mephisto}~\#3445, Clue 30A (\ans{HOLLOA} vs.\ \ans{HALLOA}; Clue~C24).}
    The charade wordplay (\ans{HALL} [\textit{``clear space''}] $+$ \ans{O} $+$ \ans{A} $\to$ \ans{HALLOA}) uniquely determines the vowel \ans{A}, but the models rationalized the more familiar standard spelling \ans{HOLLOA}.
    \item \textbf{\textit{Mephisto}~\#3449, Clue 21A (\ans{CARDAMOM} vs.\ \ans{CARDAMUM}; Clue~C25).}
    The charade combines the archaic \textit{Chambers} verb \ans{CARD} (\textit{``mix of old''}) with the Scots dialect headword \ans{A-MUM} (\textit{``not a word about it locally''}) to yield \ans{CARDAMUM}, whereas the models emitted the everyday spelling \ans{CARDAMOM}.
    \item \textbf{\textit{Mephisto}~\#3440, Clue 27D (\ans{REIVE} vs.\ \ans{RIEVE}; Clue~C23).}
    In \clue{Always rounding one up, as rustlers did once over the border (5)}, the surface phrasing is ambiguous between reversing $\operatorname{INS}(\ans{EVER},\ans{I})$ to yield \ans{REIVE} versus inserting \ans{I} into $\operatorname{REV}_{\text{Down}}(\ans{EVER})$ to yield \ans{RIEVE}.
\end{itemize}
When per-call limits were relaxed to $10\text{m}$ and $21{,}000$ tokens on the two lowest-scoring puzzles (3445 and 3446), the solver recovered $3$ of the $5$ misses ($67/72 \to 70/72$, solving \ans{SPEISSCOBALT}, \ans{UNDISCORDANT}, and \ans{TRIARCH}).

\subsection{What Matters in the Mean: Single-Stage vs.\ Two-Stage Ablations}
\label{sec:mean-eval}

\begin{figure}[!htbp]
\centering
\includegraphics[width=\textwidth]{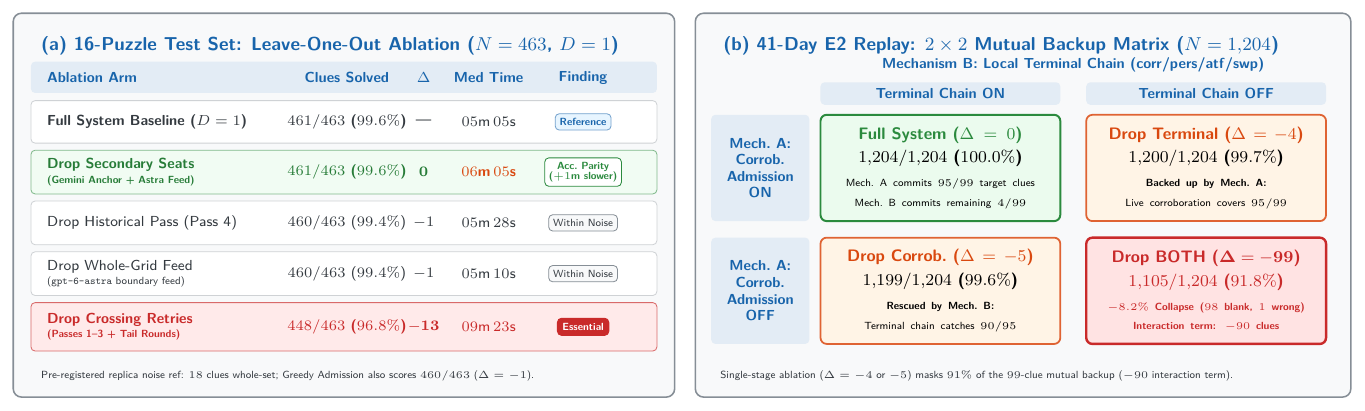}
\caption{\textbf{Average-case stage ablations: single-stage leave-one-out versus $2\times 2$ mutual backup (\Cref{tab:app-ablations}).}
\textbf{(a)} Single-replica ($D=1$) leave-one-out ablation on the 16-puzzle test set ($N = 463$), where only dropping crossing retries ($-13$ clues) exceeds replica noise.
\textbf{(b)} $2\times 2$ factorial replay across $41$ Epoch~E2 daily puzzles ($N = 1{,}204$), showing that pass-barrier corroboration admission and the local terminal chain act as mutual backups ($-5$ and $-4$ individually, $-99$ when both are removed).}
\label{fig:stage-ablation}
\end{figure}

To measure the marginal contribution of each pipeline stage under normal API conditions, we ran an 80-puzzle-run leave-one-out ablation on the 16-puzzle stratified test set using a single replica per puzzle ($D=1$) across $5$ arms (the full championship baseline plus four leave-one-out arms dropping crossing retries, Pass~4, the background whole-grid feed, or the secondary per-clue seats, alongside a sixth comparison arm replacing corroboration admission with greedy admission in \Cref{tab:app-ablations}, Panel~A) against our pre-registered replica-noise reference ($18$ clues whole-set, or $\max(1, \text{disagreement}_p)$ per puzzle).
We paired this test-set study with an exact local replay $2\times 2$ factorial ablation across $41$ banked Epoch~E2 daily puzzles ($1{,}204$ clues; $2$ of the $43$ E2 days were excluded by the pre-registered cell-exact replay fidelity gate because they differed by $1$ clue each under replay) (\Cref{fig:stage-ablation}; full tabular breakdowns appear in \Cref{tab:app-ablations}).

\Cref{fig:stage-ablation} illustrates three lessons about ablating multi-stage reasoning pipelines.
\begin{enumerate}[leftmargin=1.8em,itemsep=2pt]
    \item \textbf{In the single-replica test-set leave-one-out ablation, only crossing-letter retries ($-13$ clues) exceed replica noise.}
    Disabling the crossing-constrained retry passes (Passes~1--3 and the tail rounds, so the solver must rely solely on Pass~0, Pass~4, the background whole-grid feed, and the local terminal chain) drops test-set accuracy from $461/463$ to $448/463$ ($-13$ clues, introducing $8$ new wrong answers and $5$ blanks and exceeding the pre-registered per-puzzle noise reference on \texttt{29303} [$-4$ vs.\ ref $1$] and \texttt{29435} [$-6$ vs.\ ref $2$]), confirming the crossing-amplification mechanism of \cref{sec:crossings-amp}.
    By contrast, removing the Pass~4 historical-answer lookup ($-1$ clue), removing the background \texttt{gpt-6-astra} whole-grid feed ($-1$ clue), or replacing corroboration admission with greedy admission ($-1$ clue) each changes the total accuracy score by a single clue (within single-replica draw noise under the $D=1$ ablation protocol).
    \item \textbf{Removing secondary per-clue seats preserves test-set accuracy ($461/463$) but slows solve speed by $\approx 1\text{m}$.}
    Removing both secondary per-clue seats (Claude and GPT-5) from Passes~0--3 and running per-clue passes on the Gemini anchor alone (with the background feed retained, which contributed $7$ admissions vs.\ $8$ in the $D=1/3$ first-finisher baseline) scores \textbf{$461/463$ ($\Delta = 0$, $2$ clues lost and $2$ gained)} on puzzles all published $5$--$14$ months after Gemini's training cutoff.
    This invariance holds for \textit{clue correctness only, not solve speed}: without Claude's fast Pass~0 completions ($14.1\text{s}$ median vs.\ $21.1\text{s}$ on Gemini) to supply early crossing letters, diagnostic median solve time increases by $\approx 1\text{m}$ ($6\text{m}\,05\text{s}$ vs.\ $5\text{m}\,05\text{s}$ single-replica baseline; \Cref{tab:app-ablations}).
    \item \textbf{Single-stage ablation hides a 99-clue mutual backup between pass-barrier corroboration admission and the local terminal chain.}
    The $2\times 2$ factorial experiment in \Cref{fig:stage-ablation}b explains why the local terminal stages (\textsc{corr}, \textsc{pers}, and \textsc{atf}) committed $0$ clues in live Epoch~E2 runs (\Cref{tab:e2-stage-seat}), and why turning off corroboration admission alone cost only $5$ clues in the 41-day replay.
    Across the $1{,}204$ daily clues in the 41-day cell-exact replay (where the full system solves all $1{,}204/1{,}204$ clues), there are $99$ clues that require crossing-supported admission or post-pass recovery (such as a lone Peer/Third-Seat proposal lifted by a kept crossing, e.g., \ans{PAUPERIZE} receiving $+1$ support at letter \ans{P} once 1A \ans{TAPE-RECORD} is kept in $K^{(1)}$).
    In the full system, two different mechanisms can commit these answers (under $0$-cost fixed-log replay over the banked pass proposals, holding downstream prompts fixed):
    \begin{itemize}[leftmargin=1.5em,itemsep=1pt]
        \item \textit{Mechanism A (Live Pass-Barrier Corroboration, \cref{alg:corrob}).}
        Admits a lone Peer or Third-Seat proposal ($b=1$) at the end of Pass~1, 2, or 3 as soon as an already-kept crossing word lends $+1$ support ($s_i^{(t)} \ge 2$).
        \item \textit{Mechanism B (Local Terminal Persistence \& Singleton Chain, \textsc{corr}/\textsc{pers}/\textsc{atf}/\textsc{swp}).}
        At the end of the run, inspects the saved pass logs and commits any uncommitted proposal that matches the final crossing pattern and was either a UKACD singleton or repeated across passes (followed by \textsc{swp}).
    \end{itemize}
    When both mechanisms are active, Mechanism~A commits $95$ of those $99$ target clues during Passes~1--3 and Mechanism~B commits the remaining $4$ at termination ($99/99$ target clues solved; $1{,}204/1{,}204$ overall).
    Turning off Mechanism~B alone leaves those $4$ terminal clues blank ($-4$).
    Conversely, turning off Mechanism~A alone leaves the $95$ clues uncommitted during Passes~1--3, but Mechanism~B steps in at the end of the run and rescues $90$ of them (leaving $5$ wrong fills via \textsc{swp}, $\Delta = -5$).
    Only when both mechanisms are turned off at once does a 99-clue collapse occur ($98$ left blank, $1$ wrong, $\Delta = -99$; interaction term $-90$ clues).
    In short, live corroboration admission and the local terminal chain act as mutual backups: ablating one at a time masks $91\%$ of the 99-clue collapse ($-5$ or $-4$ vs.\ $-99$).
\end{enumerate}

\subsection{What Matters in the Tail: Single-Shot Baselines, Provider Weather, and the Championship Final}
\label{sec:tail-eval}

Although a single per-clue seat with retry passes scores $461/463$ on the test set, the multi-seat ensemble and asynchronous whole-grid feed are designed for the tail regime, which we examine across three settings.

\begin{table}[!htbp]
\centering
\caption{Single-shot whole-grid frontier baselines across the $7$ puzzles of the 2026 \textit{Times} Championship ($3\text{ draws/puzzle} = 21\text{ runs}$, $630$ clues per model, ordered reverse-chronologically by GA date; plus $9$ hard draws on SF1, SF3, and Final for \texttt{gemini-4-argon} at a 65k-token cap).}
\label{tab:e17-oneshot}
\small
\begin{adjustbox}{max width=\textwidth}
\begin{tabular}{@{}llrrrrl@{}}
\toprule
\textbf{System / Standalone Model} & \textbf{GA Date} & \textbf{Perfect Grids} & \textbf{Clues Solved} & \textbf{Wrong / Blank} & \textbf{Med Time (Range)} & \textbf{Observed Tail Hazards} \\
\midrule
\textbf{\Darmok{} (Live Contest)} & 2026-10-10 & $\mathbf{7 / 7}$ ($\mathbf{100\%}$) & $\mathbf{210 / 210}$ & $\mathbf{0 / 0}$ & $\mathbf{05{:}08}$ ($03{:}32$--$07{:}49$) & Survived 6-min Final brownout \\
\midrule
\texttt{gemini-4-argon} (32k Cap)      & 2026-10-13 & $15 / 21$ ($71.4\%$) & $450 / 630$ & $0 / 180$ & $04{:}33$ ($02{:}21$--$06{:}52$) & $6/21$ \texttt{MAX\_TOKENS} ($0/30$) \\
\texttt{gpt-6-astra} (One-Shot)        & 2026-09-12 & $21 / 21$ ($100\%$) & $630 / 630$ & $0 / 0$   & $04{:}32$ ($03{:}26$--$09{:}12$) & Censored/failed on $17/20$ \textit{Mephisto} runs \\
\texttt{gemini-3.8-flash} (One-Shot)   & 2026-09-02 & $11 / 21$ ($52.4\%$) & $471 / 630$ & $12 / 147$ & $08{:}30$ ($03{:}34$--$10{:}55$) & $6/21$ loop truncations \\
\texttt{gpt-5.6-sol} (One-Shot)        & 2026-07-09 & $16 / 21$ ($76.2\%$) & $509 / 630$ & $1 / 120$ & $08{:}54$ ($04{:}28$--$30{:}00$) & $3$ timeouts, $1$ HTTP 520 \\
\texttt{claude-fable-5} (One-Shot)     & 2026-07-06 & $19 / 21$ ($90.5\%$) & $570 / 630$ & $0 / 60$  & $14{:}54$ ($07{:}02$--$22{:}31$) & $2/21$ safety refusals ($0/30$) \\
\texttt{gemini-3.1-pro} (One-Shot)     & 2026-04-02 &  $7 / 21$ ($33.3\%$) & $488 / 630$ & $89 / 53$ & $04{:}00$ ($02{:}39$--$08{:}08$) & $0/12$ perfect on SF1--Final \\
\midrule
\texttt{gemini-4-argon} (65k, Hard 9)  & 2026-10-13 & $7 / 9$ ($77.8\%$)   & $268 / 270$ & $2 / 0$   & $06{:}55$ ($04{:}35$--$08{:}43$) & Missed SF1 28A (\ans{CONTEST}) $2\times$ \\
\bottomrule
\end{tabular}
\end{adjustbox}
\end{table}

\subsubsection{Single-Shot Frontier Parity vs.\ Bimodal Tail Hazards}
\label{sec:tail-oneshot}

When queried with a whole championship puzzle in a single prompt ($7\text{ puzzles} \times 3\text{ draws} = 21\text{ runs}$ per model; \Cref{fig:intro-pareto}b and \Cref{tab:e17-oneshot}), three frontier models (\texttt{gemini-4-argon}, \texttt{gpt-6-astra}, and \texttt{claude-fable-5}) achieved $100\%$ accuracy on every draw that returned a complete JSON response. Ordering the models by release date also shows a sharp drop in latency from the July 2026 cohort (\texttt{claude-fable-5} and \texttt{gpt-5.6-sol}, whose runtime distributions stretch from $5\text{m}\,23\text{s}$ to $30\text{m}\,00\text{s}$) to the September--October 2026 cohort (\texttt{gpt-6-astra} and \texttt{gemini-4-argon}, whose completed solves all finish under $9\text{m}\,12\text{s}$, inside the $10\text{m}\,05\text{s}$ human Final winning time).
In particular, standalone \texttt{gpt-6-astra} solved all $21/21$ championship draws without error in a median of $4\text{m}\,32\text{s}$,\footnote{Exact one-sided $95\%$ Clopper--Pearson binomial upper bound on its championship-grid failure rate: $\le 13.3\%$ across $21$ draws, or $\le 34.8\%$ across $7$ distinct puzzles, noting that $3$ draws on $7$ selected puzzles are not $21$ independent draws from the puzzle population; the same $\le 34.8\%$ 7-puzzle championship upper bound applies to \Darmok{} on the contest day itself, alongside \Darmok{}'s $2/43 = 4.7\%$ observed non-perfect rate ($\operatorname{Beta}^{-1}(0.95; 3, 41) = 13.9\%$ upper bound) on the 43-day post-cutoff Epoch~E2 corpus.} and on Championship Day its live background feed returned a $30/30$ sheet on the Final at $5\text{m}\,38\text{s}$ ($74.0\text{s}$ before the first per-clue replica locked its grid at $6\text{m}\,52\text{s}$: $72.9\text{s}$ waiting for the Pass~1 barrier plus $1.1\text{s}$ of verification and PDF output).

On standard $15\times 15$ championship grids, single-call capability is no longer the bottleneck.
What \Darmok{}'s per-clue barrier architecture added over standalone \texttt{gpt-6-astra} in the Final was $+74.0\text{s}$ of pass-barrier wait and verification latency in exchange for multi-seat cross-verification (confirming $26/30$ clues via per-clue seats and pre-filling $13/17$ checked entry positions [$13/15$ unique checked physical cells] of the $4$ feed-admitted entries) and active failover protection against three single-shot tail failure modes:
\begin{itemize}[leftmargin=1.8em,itemsep=2pt]
    \item \textbf{Reasoning-loop token-cap starvation (\texttt{gemini-4-argon} and \texttt{gemini-3.8-flash}).}
    At its default SDK output cap ($32{,}768$ tokens), \texttt{gemini-4-argon} went $15/15$ ($450/450$, zero errors) whenever it finished, but on $6$ of $21$ draws (all $3$ draws of Semi-Final~3, $2$ of Semi-Final~1, and $1$ of the Final at $7\text{m}\,07\text{s}$, highlighted by the gold ring in \Cref{fig:intro-pareto}b) the saved reasoning traces in our telemetry logs show the model re-evaluating candidate crossing letters in a cycle on $1$--$2$ stubborn clues until it hit \texttt{finish\_reason:\ MAX\_TOKENS} at $32{,}768$ tokens before emitting the opening brace of the JSON response, scoring $0/30$.
    Raising the cap to $65{,}536$ tokens on those three hardest puzzles ($9$ draws) eliminated all truncations and scored $268/270$ ($99.26\%$).
    Whole-grid single-shot prompts thus couple all $30$ clues into a single token budget, allowing one cyclic clue to starve the other $29$ already-solved answers.
    \item \textbf{Timeouts, partial grids, and \textit{Mephisto} right-censoring (\texttt{gpt-6-astra} and \texttt{gpt-5.6-sol}).}
    Standalone \texttt{gpt-6-astra} went $21/21$ on standard $15\times 15$ championship grids (including $3/3$ on the Final at $5\text{m}\,02\text{s}$, $5\text{m}\,31\text{s}$, and $5\text{m}\,38\text{s}$), yet when invoked as the background boundary feed across our $20$ barred \textit{Mephisto} runs ($10$ baseline + $10$ ablation), the one-shot process completed before the per-clue pipeline ($10\text{m}$--$18\text{m}$) on only \textbf{$3$ of $20$ runs} (being right-censored upon per-clue completion after $10\text{m}$--$18\text{m}$ on $14$ runs and returning errors on $3$).
    Meanwhile, \texttt{gpt-5.6-sol} suffered $4/21$ non-returns ($3$ timeouts at $30\text{m}\,00\text{s}$ and $1$ HTTP~520 error at $20\text{m}\,18\text{s}$), and on the Championship Final it produced $1$ partial grid ($29/30$ at $9\text{m}\,49\text{s}$) alongside $2$ slow solves ($19\text{m}\,28\text{s}$ and $29\text{m}\,48\text{s}$).
    Thus, even when a frontier model handles standard $15\times 15$ blocked grids in one shot, per-clue orchestration transfers more reliably and finishes faster on dense, \textit{Chambers}-vocabulary \textit{Mephisto} puzzles.
    \item \textbf{Spurious safety-classifier refusals and slow completion (\texttt{claude-fable-5}).}
    Claude solved $19/21$ draws perfectly ($570/570$), but on Qualifier~3 (draw~2) and the Championship Final (draw~2), the provider's content-safety filter aborted generation after $1\text{m}\,09\text{s}$ and $1\text{m}\,21\text{s}$ with \texttt{finish\_reason:\ refusal} and zero output tokens, scoring $0/30$.
    Cryptic surfaces routinely combine benign idioms involving words such as ``executed,'' ``divorced,'' ``shot,'' or ``drug'' that trip sequence-level classifiers when $30$ clues are concatenated into one monolithic prompt, creating a single point of failure (SPOF).
    Claude's median solve time on successful draws was also $14\text{m}\,54\text{s}$ ($14\text{m}\,36\text{s}$ and $15\text{m}\,22\text{s}$ on its two completed Final draws, and up to $22\text{m}\,31\text{s}$ on Semi-Final~3), behind human championship pace.
    Per-clue decomposition isolates safety-filter false positives to at most a single clue rather than aborting the entire 30-clue puzzle.
\end{itemize}

\subsubsection{The 132-Day Production Tail}
\label{sec:tail-daily}

\begin{figure}[!htbp]
\centering
\includegraphics[width=0.78\textwidth]{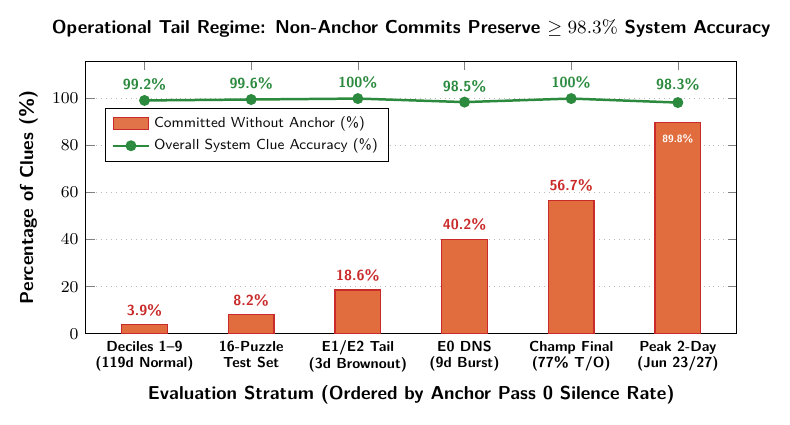}
\caption{\textbf{Operational tail failover across evaluation strata.}
Non-anchor commit share (orange bars) and overall system clue accuracy (green line) across evaluation strata ordered by Gemini anchor Pass~0 silence rate (\cref{sec:tail-daily}, \Cref{tab:app-tail}), holding system clue accuracy between $98.3\%$ and $100.0\%$.}
\label{fig:mean-vs-tail}
\end{figure}

To measure how often primary-seat failures occur in practice (\Cref{fig:mean-vs-tail} and \Cref{tab:app-tail}), we ranked all $132$ daily production runs (spanning Epochs~E0--E2) by the Gemini anchor's Pass~0 silent rate (the fraction of clues on which the anchor returned neither an answer nor a low-confidence parse due to timeouts, socket errors, or parse failures).
Across all $132$ days, the median daily Pass~0 silent rate was $21.4\%$ ($p_{90} = 40.0\%$).
In the worst decile ($13$ days with $42.9\%$--$100.0\%$ anchor Pass~0 silence), overall clue accuracy held at \textbf{$373/378$ ($98.68\%$, $1$ wrong, $4$ blank, $8/13$ perfect grids)} versus $3{,}471/3{,}498$ ($99.23\%$, $4$ wrong, $23$ blank, $100/119$ perfect grids) on the remaining $119$ days, a difference of roughly two clues across $13$ puzzles.
Because Decile~10 spans multiple architectural epochs and failure causes ($5/10$ perfect grids under Epoch~E0 greedy admission vs.\ $3/3$ under Epoch~E1/E2 corroboration admission), we disaggregate it into its three constituent regimes (\Cref{tab:app-tail}):
\begin{itemize}[leftmargin=1.8em,itemsep=2pt]
    \item \textbf{9 Epoch~E0 client-side DNS burst runs ($100\%$ Pass~0 silent, $262$ clues).}
    Between June~6 and July~1, 2026 (including $6$ runs batch-backfilled on June~29 and July~5 after an author vacation), firing $60$ simultaneous Pass~0 connections without pacing triggered a client-side macOS resolver failure (\texttt{ConnectError:\ [Errno 8] nodename nor servname provided, or not known}) on $100\%$ of Gemini's Pass~0 calls while Claude's connection pool succeeded.
    On $7$ of these $9$ runs, Gemini recovered in Pass~1 once concurrency dropped; on the $2$ peak runs where Gemini failed again in Pass~1 (puzzle~\#29576 on June~23 and puzzle~\#29580 on June~27), Claude and GPT-5 supplied $89.8\%$ ($53/59$) of committed answers and scored $59/60$ ($98.33\%$).
    Across all $9$ DNS-burst runs, accuracy was $258/262$ ($98.47\%$, slightly above the $97.82\%$ [$854/873$] average of the other $30$ Epoch~E0 days), with $40.2\%$ ($104/259$) of answers committed without the anchor.
    \item \textbf{1 Epoch~E0 partial-error run ($70.0\%$ Pass~0 silent, $30$ clues).}
    On June~20, 2026 ($13$ errors + $8$ timeouts in Pass~0), Gemini recovered in Pass~1 and the run scored $29/30$ ($96.67\%$).
    \item \textbf{3 Epoch~E1/E2 provider timeout brownout days ($42.9\%$--$46.7\%$ Pass~0 silent, $86$ clues).}
    Under the corroboration-admission architecture (August~6 in E1; September~3 and September~25 in the frozen E2 configuration), elevated provider timeouts silenced Gemini on nearly half of Pass~0, yet the solver achieved \textbf{$86/86$ ($100.0\%$ accuracy, $3/3$ perfect grids)}, with secondary seats supplying $18.6\%$ ($16/86$) of committed answers ($15.5\%$ on the two E2 days vs.\ $5.3\%$ on the other $41$ E2 days).
\end{itemize}
Across both client-side socket bursts in E0 and provider timeout brownouts in E1/E2, heterogeneous secondary seats supplied substantial answer coverage during anchor outages, preserving $98.68\%$ clue accuracy across Decile~10 ($3/3$ perfect grids on the corroboration-era brownout days and $8/13$ perfect grids across the mixed-epoch tail).

\begin{figure}[!htbp]
\centering
\includegraphics[width=\textwidth]{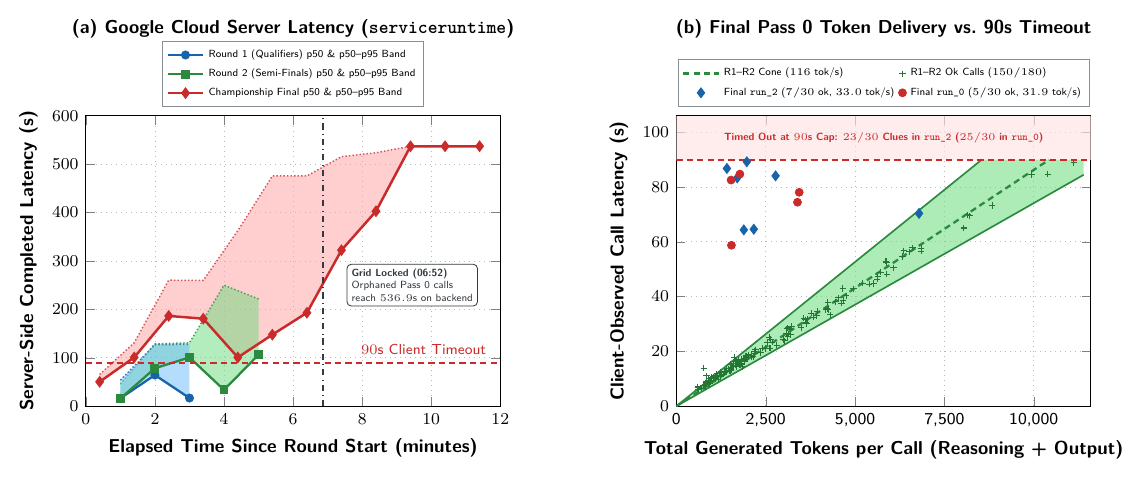}
\caption{\textbf{Dual-sided telemetry of the Championship Final anchor brownout (October 10, 2026; \cref{sec:tail-final}).}
\textbf{(a)} Project-scoped Google Cloud \texttt{serviceruntime} server-side request latency ($\text{p50}$ and $\text{p50}$--$\text{p95}$ band) across the three championship rounds ($90$ concurrent Pass~0 requests per round); completions logged after grid lock ($t = 6\text{m}\,52\text{s}$, up to $536.9\text{s}$) reflect client-cancelled Pass~0 ``zombie'' requests finishing on the backend.
\textbf{(b)} Client-observed Pass~0 call latency versus total generated tokens across Rounds~1--2 ($150/180$ completed calls, median $116\text{ tokens/s}$, green cone) and the Championship Final ($33.0\text{ tokens/s}$ in \texttt{run\_2}, where $23/30$ calls hit the $90\text{s}$ timeout cap).}
\label{fig:brownout}
\end{figure}

\subsubsection{Anatomy of the Championship Final}
\label{sec:tail-final}

The Championship Final on October 10, 2026 provided the decisive live test of the resilience of this architecture (\Cref{fig:brownout} and \Cref{tab:app-final}).
Throughout Qualifiers~1--3 and Semi-Finals~1--3, Gemini achieved a median effective token delivery rate (defined over non-streaming \texttt{generate\_content} calls as $(\text{reasoning} + \text{output tokens}) / \text{elapsed seconds}$) of $114.6$--$119.1\text{ tokens/s}$, timing out on only $1$ to $8$ clues per puzzle.\footnote{In Semi-Final~2, the last unsolved clue (10A \ans{CATHERINE HOWARD}; Clue~C13, \Cref{tab:app-clues-2}) was supplied by the \texttt{gpt-6-astra} boundary feed: the feed file landed on disk at $5\text{m}\,32\text{s}$ ($332.0\text{s}$), just $13.7\text{s}$ after Pass~2 had entered at $5\text{m}\,18\text{s}$ ($318.3\text{s}$); because the boundary feed is polled only at pass barriers, the run waited $136.5\text{s}$ for Pass~2's in-flight calls to exit at $7\text{m}\,49\text{s}$ ($468.5\text{s}$) before admitting \ans{CATHERINE HOWARD} and finishing.}

Seconds after the Championship Final PDF was ingested, the anchor endpoint suffered a severe latency brownout.
Because each replica (\texttt{run\_0}, \texttt{run\_1}, and \texttt{run\_2}) executes as an independent OS process writing its own per-call JSONL telemetry log (\texttt{pane\{0,1,2\}\_p944766\_calls.jsonl}) that records the exact \texttt{reasoning\_tokens}, \texttt{output\_tokens}, and client wall-clock \texttt{response\_time\_ms} returned in each call's API response metadata, we can measure token delivery per replica directly.
Across completed Pass~0 calls, median effective token delivery dropped by $72\%$ to \textbf{$33.0\text{ tokens/s}$ in the winning replica (\texttt{run\_2})} ($31.9\text{ tokens/s}$ in \texttt{run\_0}; \Cref{fig:brownout}b).
Call-level telemetry shows that six of the seven surviving Pass~0 calls in \texttt{run\_2} generated normal reasoning volumes of only $1{,}414$ to $2{,}776$ total tokens (which require $12$--$24\text{s}$ at $115\text{ tokens/s}$), yet took $64.3\text{s}$ to $89.2\text{s}$ of wall-clock time ($16.3$--$33.5\text{ effective tokens/s}$; e.g., clue~21A \ans{COUNTEREXAMPLE} [Clue~C19] generated $1{,}414$ tokens in $86.7\text{s}$, while a single early-scheduled call, 9D \ans{ROAD RAGE} [Clue~C16], generated $6{,}791$ tokens in $70.3\text{s}$ at $96.5\text{ tokens/s}$ before the slowdown deepened).
Because queueing delay and reduced token delivery consumed $50$--$70\text{s}$ per call, any clue requiring $>3{,}000$ thinking tokens hit the $90\text{s}$ client timeout cap, causing $23$ of the $30$ Pass~0 calls to time out in the winning replica ($25/30$ in \texttt{run\_0}).

We cross-checked this event using project-scoped Google Cloud Monitoring telemetry (\texttt{serviceruntime.\allowbreak googleapis.\allowbreak com/api/request\_latencies} and \texttt{request\_count}; \Cref{fig:brownout}a) and our pre-championship concurrency rehearsals.
\begin{itemize}[leftmargin=1.8em,itemsep=2pt]
    \item \textbf{Identical client concurrency across all three rounds.}
    Because Round~1 (three Qualifying puzzles in parallel) and Round~2 (three Semi-Final puzzles in parallel) ran across three panes simultaneously, all three championship rounds launched with the exact same client concurrency profile: $3\text{ panes} \times 30\text{ clues} = 90$ in-flight Pass~0 requests per primary seat, staggered over $4.5\text{s}$ ($20\text{ req/s}$).
    In a pre-championship rehearsal on October~3 (\texttt{R45}, $3 \times 30 = 90$ concurrent requests on a hard puzzle), Gemini had achieved $97\%$--$100\%$ Pass~0 completion at a median latency of $26.3\text{s}$--$31.8\text{s}$.
    \item \textbf{Server-side latency spike and post-lock ``zombie'' completions.}
    In Round~1 (10:15~UTC) and Round~2 (11:46~UTC), Google's server-side \texttt{serviceruntime} latency $\text{p50}$ during the opening minute was $16.25\text{s}$ and $15.94\text{s}$.
    In the Final (14:36~UTC), server-side $\text{p50}$ jumped $3.1\times$ in the first minute to $50.3\text{s}$ ($\text{p95} = 96.0\text{s}$), breaching the $90\text{s}$ client timeout cap in minute~2 at $\text{p50} = 100.7\text{s}$ ($\text{p95} = 130.9\text{s}$) and reaching a minute-bucket median $\text{p50}$ of $186.4\text{s}$ during the live solve window ($t \le 6\text{m}\,52\text{s}$).
    Google Cloud's \texttt{serviceruntime} metric records a server-side latency sample at the timestamp when the backend \textit{finishes} generating a response, even if the client closed its HTTP socket earlier at the $90\text{s}$ client timeout.
    Because all three replicas had finished or been terminated by $14{:}44\text{ UTC}$ (after \texttt{run\_2} locked the winning grid at $14{:}43{:}24\text{ UTC}$, $t = 6\text{m}\,52\text{s}$) and no new requests were issued thereafter, the server completions recorded in the $14{:}44$--$14{:}50\text{ UTC}$ minute buckets with $\text{p50}$ climbing from $192.9\text{s}$ to $536.9\text{s}$ (\Cref{fig:brownout}a) directly reflect the orphaned Pass~0 generations continuing to execute on Google's backend for $180$--$536.9\text{s}$, which also compounded project-level queue contention during Pass~1.
\end{itemize}

The pass-barrier logs of the winning replica (\texttt{run\_2}; \Cref{tab:app-final}) show how \Darmok{} still finished $30/30$ in $6\text{m}\,52\text{s}$ while its primary anchor timed out on $23/30$ clues in Pass~0 (and failed on $17$ of the $23$ open clues re-dispatched to Gemini in Pass~1 [$16$ timeouts at $90\text{s}$ plus $1$ JSON parse failure on 18A at $84.1\text{s}$], proposing only $13/30$ answers across the entire run before grid lock).
\begin{enumerate}[leftmargin=1.8em,itemsep=2pt]
    \item \textbf{Pass 0 Barrier ($t = 1$, $7$ clues admitted, $23$ remain open).}
    In Pass~0, only $7$ Gemini calls beat the $90\text{s}$ timeout wall ($6$ corroborated by Claude on 20A, 21A, 25A, 28A \ans{EDITOR} [Clue~C21], 5D, and 6D, plus $1$ lone Gemini commit on 9D \ans{ROAD RAGE} [Clue~C16]).
    Meanwhile, Claude returned status \texttt{ok} on $26/30$ clues in a median of $17.8\text{s}$, and $16$ of its qualifying solo answers ($\ge 85$) were held back at $b_i^{(1)} = 1 < \theta = 2$ awaiting corroboration.
    On the hardest clues where Claude (in Pass~0) and GPT-5 (when it joined in Pass~1) failed to parse the wordplay, they proposed plausible \textit{definition-only confabulations}:
    \begin{itemize}[leftmargin=1.5em,itemsep=1pt]
        \item On Final 1D (\ans{MATES RATES}; Clue~C14, \Cref{tab:app-clues-2}), Claude proposed \ans{FIRST NAMES} at confidence $78$ in Pass~0, and GPT-5 proposed \ans{MATES AGAIN} at confidence $82$ in Pass~1.
        \item On Final 14A (\ans{AIRED}; Clue~C17), Claude proposed \ans{ALERT} at confidence $82$ in Pass~0, and GPT-5 proposed \ans{TIRED} at confidence $63$ in Pass~1.
        \item On Final 27A (\ans{SNEAKING}; Clue~C20), Claude in \texttt{run\_0} proposed \ans{HANGOVER} at confidence $82$ in Pass~1, and GPT-5 in \texttt{run\_2} proposed \ans{ELONGATE} at confidence $58$ in Pass~1.
    \end{itemize}
    Because Claude's calibration floor sits at $\tau_{\text{peer}} = 85$ (\Cref{fig:calibration}a), \ans{FIRST NAMES} ($78$), \ans{ALERT} ($82$), and \ans{HANGOVER} ($82$) were excluded from $\mathcal{Q}_i^{(t)}$ ($i \notin S^{(t)}$).
    When GPT-5 joined in Pass~1, its lone proposals above $\tau_{\text{third}} = 50$ received $b_i^{(2)} = 1 < \theta = 2$ and were held out of $K^{(2)}$ for lack of crossing support, keeping all clues uncorrupted.
    \item \textbf{Pass 1 Barrier ($t = 2$, $19$ additional per-clue commits + $4$ boundary-feed commits $= 30/30$).}
    Re-dispatching the $23$ open clues in Pass~1 to all three seats (where Gemini returned $6/23$ \texttt{ok}, $16$ timeouts, and $1$ parse failure; Claude returned $16/23$ \texttt{ok}, $5$ timeouts, and $2$ parse failures on 4A and 11A; and GPT-5 returned $19/23$ \texttt{ok} and $4$ empty) produced $19$ Pass~1 admissions across $26/30$ total clues ($145$ of $162$ white cells filled).
    Cumulative across Passes~0--1, $6$ of the $19$ Pass~1 admissions were proposed by all three seats: $5$ with all three seats returning in Pass~1 (including the deciding human clue 2D \ans{SNARF} [Clue~C15], which Claude had proposed alone at confidence $85$ in Pass~0 and all three seats confirmed in Pass~1) plus 11A \ans{PHOTO}, where Gemini and GPT-5 agreed in Pass~1 on the answer Claude had already proposed in Pass~0.
    Another $10$ were proposed by Claude and GPT-5 without Gemini ($8$ via within-Pass-1 agreement on 1A, 3D, 8D, 10A, 15D, 17D, 22D, and 24D, plus $2$ clues, 4A \ans{WATERLOG} and 26A \ans{FORTHWITH}, where Claude proposed the answer in Pass~0 and GPT-5's Pass~1 proposal was lifted by kept crossings), and $3$ were proposed by Claude alone (7D \ans{LOOM LARGE}, 12A \ans{SOFT UNDERBELLY}, and 16A \ans{ANCESTRAL}, where GPT-5 returned empty and Claude's Pass~1 proposal was lifted by kept crossings: $s_i^{(2)} = b_i^{(2)} + w(\sigma_j^{(1)}) \ge 2$).
    Finally, when the Pass~1 barrier closed at session $t = 410.7\text{s}$ ($\text{solver\_dt} = 402.3\text{s}$, after the $8.4\text{s}$ PDF parse), the orchestrator polled the \texttt{gpt-6-astra} boundary feed, which had written its $30/30$ sheet to disk at session $t = 337.8\text{s}$ ($\text{solver\_dt} = 329.4\text{s}$) and waited $72.9\text{s}$ on both clocks for Pass~1's $150\text{s}$ Claude timeout barrier to finish.
    Admitting the last $4$ open clues from the feed (1D \ans{MATES RATES} [C14], 14A \ans{AIRED} [C17], 18A \ans{EXPEDIENT} [C18], and 27A \ans{SNEAKING} [C20]) and running $1.1\text{s}$ of verification and PDF output locked the $30/30$ grid at $6\text{m}\,52\text{s}$ ($411.8\text{s}$ session / $403.4\text{s}$ solver, $74.0\text{s}$ total after feed completion).
\end{enumerate}

\paragraph{Counterfactual Replay and Cross-Verification in the Final.}
Running our offline admission harness on the recorded Final call logs evaluates the conditional replay of admission decisions across three failover questions.
\begin{enumerate}[leftmargin=1.8em,itemsep=2pt]
    \item \textbf{Anchor-only conditional replay.}
    Removing Claude, GPT-5, and the boundary feed under the recorded Gemini timeouts admits only $7/30$ clues in Pass~0 and $13/30$ in Pass~1 in \texttt{run\_2} ($15/30$ in \texttt{run\_0}), leaving more than half the grid unsolved at the end of Pass~1.
    \item \textbf{Per-clue seats without boundary feed replay.}
    Retaining the three per-clue seats without the boundary feed commits $26/30$ correct answers ($145/162$ white cells) by the end of Pass~1, leaving the $4$ feed-admitted clues open for Pass~2.
    \item \textbf{Cell-level mutual verification between per-clue seats and the boundary feed.}
    Across the $17$ checked entry positions of the $4$ feed words ($15$ unique checked physical grid cells), the $26$ per-clue seat commits in \texttt{run\_2} pre-filled $13/17$ checked entry positions ($76.5\%$, or $13/15 = 86.7\%$ of unique checked physical cells): $4/4$ on 27A \ans{SNEAKING} (\texttt{S\_E\_K\_N\_}), $4/5$ on 18A \ans{EXPEDIENT} (\texttt{\_\_P\_D\_E\_T}, positions 3, 5, 7, 9), $2/3$ on 14A \ans{AIRED} (\texttt{\_\_R\_D}), and $3/5$ on 1D \ans{MATES RATES} (\texttt{M\_T\_S\_\_\_\_\_}, positions 1, 3, 5), while the remaining $2$ unique checked physical cells at 1D position 7 (\ans{A}, cell $(7,1)$) and position 9 (\ans{E}, cell $(9,1)$) cross-checked the first letters of 14A and 18A.
    Conversely, a design that accepted the \texttt{gpt-6-astra} whole-grid sheet immediately upon arrival whenever it agreed with $100\%$ of already-kept Pass~0 cells ($7/7$) would have finished at $5\text{m}\,38\text{s}$ (saving $72.9\text{s}$ of Pass~1 barrier wait), at the cost of foregoing Pass~1's independent multi-seat verification on $19$ clues.
\end{enumerate}

\subsection{Residual Errors: The ``No Good, Very Bad'', Very Long Tail Above 99.5\%}
\label{sec:error-analysis}

Across our three frozen-architecture benchmarks (Epoch~E2 daily production at $1{,}259/1{,}261 = 99.84\%$, the 16-puzzle stratified test set at $461/463 = 99.57\%$, and zero-shot \textit{Mephisto} transfer at $350/360 = 97.22\%$), \Darmok{} committed only $7$ wrong answers out of $2{,}084$ total clues ($0.34\%$).
Inspecting every residual error across \Darmok{} and our standalone frontier runs (with complete verified wordplay derivations in \Cref{tab:app-clues-1,tab:app-clues-2,tab:app-clues-3}) shows that the remaining object-level failures in this very long tail concentrate in four specific linguistic boundary cases:

\begin{enumerate}[leftmargin=1.8em,itemsep=2pt]
    \item \textbf{Unchecked-Cell Orthographic and Dialect Variants ($3/4$ \textit{Mephisto} errors).}
    Because $20.0\%$ to $50.0\%$ of each word's letters in British barred and blocked grids are unchecked (\Cref{fig:primer-grids}b--c), perpendicular words cannot disambiguate single-letter spelling variants that fall on an unchecked square.
    On test-set puzzle~29399 (Replica~2, \texttt{rep3}), clue~2D (\ans{PAUPERIZE}; Clue~C5, \Cref{tab:app-clues-1}) places its eighth letter (\ans{S} vs.\ \ans{Z}) on an unchecked cell (\Cref{fig:primer-grids}b).
    While both \ans{PAUPERISE} and \ans{PAUPERIZE} are valid British spellings of the definition (\textit{``Make poor''}), the container wordplay (\ans{PRIZE} [\textit{``value''}] wrapping around \ans{A[UP]E} $\to \ans{P[A(UP)E]RIZE}$) uniquely selects \ans{Z}, because in British English \ans{PRIZE} means ``to value'' whereas \ans{PRISE} means ``to lever''; in Replica~2, Claude glossed \ans{PRISE} as a variant of \ans{PRIZE} and emitted \ans{PAUPERISE}.
    Similarly, as detailed in \cref{sec:mephisto-eval}, three of the four errors across our 360-clue \textit{Mephisto} evaluation were \textit{Chambers Dictionary} headword variants on unchecked squares: \ans{REIVE} for \ans{RIEVE} (\#3440 27D; Clue~C23), \ans{HOLLOA} for \ans{HALLOA} (\#3445 30A; Clue~C24), and \ans{CARDAMOM} for \ans{CARDAMUM} (\#3449 21A; Clue~C25; the fourth was \ans{TOPARCH} for \ans{TRIARCH} on \#3446 19D, which resolved to \ans{TRIARCH} under extended timeout).

    \item \textbf{Correlated Multi-Seat Convergence on an Extreme Decoy ($2/2$ Test-Set errors).}
    Because \cref{eq:base-score} awards $b_i^{(t)} = 2$ whenever two seats independently propose the same string above their calibration floors, an error can enter $K^{(t)}$ if two models fall into the same decoy trap on an extreme clue before perpendicular crossings lock the clue.
    Both errors on the 463-clue test set occurred in the extreme SNITCH~$160+$ band (\Cref{tab:app-clues-3}): on puzzle~29399 clue~1D (\clue{Pet? You shouldn't have one (4)} $\to$ \ans{TAME}; Clue~C26), the models failed to resolve the British colloquialism \ans{TA} (\textit{``You shouldn't have!''}) $+$ \ans{ME} and committed \ans{TOMB} against pattern \texttt{T?M?}; and on puzzle~29435 clue~5D (\clue{Brought into line, book parts rather like Peter or Roger? (7)} $\to$ \ans{RABBITY}; Clue~C27), two seats converged on the plural carpentry noun \ans{RABBETS}.

    \item \textbf{Nested Single-Letter Substitution Blindness (\ans{CONTEST} vs.\ \ans{CONTEXT}).}
    Even frontier models exceeding $99\%$ accuracy occasionally fail on nested single-letter substitutions when the un-substituted base word is strongly primed by the surface reading.
    Across $900$ graded championship clues in our single-shot \texttt{gemini-4-argon} runs (\Cref{tab:e17-oneshot}), the model committed only a single unique wordplay error, repeated across two independent 65k-cap draws on Semi-Final~1 clue~28A (\ans{CONTEXT}; Clue~C3, \Cref{fig:primer-clues}b and \Cref{tab:app-clues-1}): in both draws, the model identified the base synonym \ans{CONTEST} and the substitution ($\ans{S} \to \ans{X}$), yet emitted the un-substituted base word \ans{CONTEST}.

    \item \textbf{Uncorroborated Terminal Sweep Fills under All-or-Nothing Scoring.}
    When every model seat abstains or times out across all passes on an extreme clue, such as puzzle~29435 clue~15D (\ans{AIR SIGNS}; Clue~C28, \Cref{tab:app-clues-3}, in Replicas~0--1 of the test set) or the sole wrong clue in Epoch~E2 (puzzle~29658 clue~8D \ans{TOW-PLANES}; Clue~C22, \Cref{tab:app-clues-2}), the deterministic terminal dictionary sweep (\textsc{swp}) fills the open clue with a lexicon match consistent with all crossing letters (\ans{ARRAIGNS} on \texttt{A?R?I?N?}; \ans{TOWELINGS} on \texttt{T?W?L?N?S}, where \ans{TOW-PLANES} was absent from the single-word lexicon).
    Under tournament rules, because a blank square already forfeits the clue, an uncorroborated terminal guess that respects all crossing letters never damages an already-kept crossing word and strictly improves expected perfect-grid yield (converting $3$ blanks into correct solves across Epoch~E2 for $1$ miss).
\end{enumerate}

\section{Discussion}
\label{sec:discussion}

\subsection{Why Static Benchmarks Misprice Redundancy}
\label{sec:disc-paradox}

We noted in \cref{sec:paradox} the central tension in how \Darmok{} was built versus how it looks under standard post-hoc ablation.
Over five months of pre-championship development, each downstream mechanism---the Pass~4 historical-answer pass, the asynchronous whole-grid boundary feed, and the four-stage local terminal chain---was added deliberately to rescue a specific class of stubborn residual clue without ever overwriting answers already admitted by upstream stages.
Yet when we evaluated the frozen championship system on our 16-puzzle test set after the tournament, removing almost any single downstream stage or secondary model seat changed accuracy by $0$ to $1$ clue (\Cref{fig:stage-ablation}a), making the architecture look at first glance like a neuro-symbolic Rube Goldberg machine~\cite{wolfe2000rube} whose auxiliary gears were doing no work.

Our five-month operational and experimental record shows why that static diagnosis misreads the structural difference between our 2025 V1 solver and our 2026 V2 architecture.
V1 really was a Rube Goldberg contraption: because it relied on greedy cell-level overrides without a global coherence invariant, ad-hoc heuristics interacted destructively across the grid.
A true Rube Goldberg machine is a \textit{serial chain of coupled indirections} in which every added lever can undo the work of an earlier one.
In effect, the transition from V1 to V2 inverted Brooks' classic \textit{second-system effect}~\cite{brooks1975mythical}: rather than bloating a spare prototype with coupled features, the second system replaced an entangled web of heuristics with a single pass-barrier invariant (\Cref{inv:coherence}) frozen several weeks before the 2026 Championship and tested thoroughly.

Because \Darmok{}'s secondary per-clue seats cannot self-admit cold without corroboration and its strictly non-evicting fallback stages (the boundary feed and local terminal chain) may only fill open residual clues ($R^{(t)}$) consistent with kept crossings, never overwriting an upstream kept answer, they form a \textit{parallel and staged failover harness} akin to $N$-version fault-tolerant software~\cite{avizienis1985nversion} and active-active replication~\cite{vonneumann1956probabilistic,lyons1962tmr,schneider1990implementing}.
In fault-tolerant systems engineering, redundancy mechanisms such as error-correcting memory~\cite{hamming1950error}, RAID arrays~\cite{patterson1988raid}, hedged RPCs~\cite{dean2013tail}, and failover replicas~\cite{schneider1990implementing} contribute zero marginal accuracy on a fault-free trace.
If one evaluates a RAID controller~\cite{patterson1988raid} by unplugging the mirror drive during a healthy 10-minute benchmark, observing zero data loss, and concluding that the mirror drive is superfluous, it is the evaluation methodology that is broken, not the storage array.

Standard machine learning ablation methodology risks making a similar mistake when applied to complex AI systems operating over commercial reasoning APIs.
Evaluated on our 16-puzzle test set, our secondary model seats (Claude and GPT-5) and our asynchronous whole-grid feed (\texttt{gpt-6-astra}) evaluate to $\Delta = 0$ and $\Delta = -1$ clues in accuracy (though removing the secondary seats slows diagnostic median solve time by $\approx 1\text{m}$) because the primary Gemini anchor is healthy.
Evaluated across $132$ daily production runs (\Cref{fig:mean-vs-tail}), in the $13$ runs where anchor Pass~0 silence exceeded $42\%$ ($9$ Epoch~E0 client-side DNS socket-burst runs, $1$ E0 partial-error run, and $3$ E1/E2 provider brownout days), those secondary seats carry $18.6\%$ to $89.8\%$ of the grid ($32.1\%$ pooled; $90.0\%$ single-day peak) while holding clue accuracy at $98.68\%$ ($373/378$; $8/13 = 61.5\%$ perfect grids overall, with all $5$ imperfect grids occurring in Epoch~E0 and $86/86 = 100\%$ [$3/3$ perfect grids] on the $3$ E1/E2 brownout days).
Evaluated on stage during the 2026 Championship Final (\Cref{fig:brownout}), when an unexpected provider latency brownout collapsed anchor token delivery by $72\%$ ($117 \to 33\text{ tokens/s}$), timed out $23/30$ opening-pass clues, and failed $17/23$ retry calls ($16$ timeouts, $1$ parse error), the secondary per-clue seats supplied $13$ answers without the anchor and the asynchronous \texttt{gpt-6-astra} whole-grid feed supplied the remaining $4$, while the per-clue seats independently verified $26/30$ clues and $13/17$ checked entry positions ($13/15$ unique checked physical cells) of the $4$ feed entries to complete $30/30$ in $6\text{m}\,52\text{s}$.

The same principle governs internal pipeline stages that act as mutual backups.
In our 41-day production replay (\Cref{fig:stage-ablation}b), single-stage leave-one-out reports that synchronous corroboration admission is worth only $5$ clues and the local terminal chain is worth only $4$ clues; yet removing both collapses accuracy by $99$ clues ($8.2\%$ of the corpus, interaction term $-90$ clues).
In systems where upstream consensus pre-empts downstream recovery, one-at-a-time subtraction masks over $90\%$ of both mechanisms' combined contribution.

\subsection{From Prompt Carpentry to External Constraint Loops and Asymmetric Calibration}
\label{sec:disc-folklore}

Our controlled clue-level experiments across $9{,}260$ paired API calls (\cref{sec:folklore}) carry an equally practical lesson for how practitioners allocate engineering effort when building on native extended-thinking models.
\begin{itemize}[leftmargin=1.8em,itemsep=2pt]
    \item \textbf{Visible prompt scaffolding shows no detected accuracy advantage over native thinking.}
    Under the tested models, prompts, and completion-conditioned endpoints, micro-engineering visible chain-of-thought field ordering (\texttt{P1}), draft-and-revise instructions (\texttt{P2}), taxonomy decomposition fields (\texttt{P3}), domain solving tutorials (\texttt{P4}), or confidence-elicitation schemas (\texttt{P6}) showed no statistically significant accuracy advantage on completed calls (\Cref{fig:folklore}b; while not a formal proof of equivalence, this is consistent with wordplay search occurring inside the thinking trace prior to JSON output).
    Prompt schemas serve two narrow systems roles: enforcing a clean output-syntax contract (where simpler flat schemas avoid nested-object formatting quirks) and acting as a brake on unverified guessing (requiring the model to articulate a wordplay explanation under \texttt{P0} so it searches until the timeout cap on unsolved clues rather than emitting fast, lower-precision guesses as under \texttt{P5}).
    \item \textbf{External constraint tightening drives the saturation jump.}
    In contrast to prompt carpentry, feeding back verified intersecting letters pushes open clues into the $n_{\text{known}} \ge 3$ high-accuracy regime (\Cref{fig:folklore}a) under order-independent kept-set reconstruction (\cref{alg:corrob}), raising per-clue solve rates by $+33$ to $+50$ percentage points and eliminating partial-word corruption (\Cref{inv:coherence}). This external loop mirrors the constraint propagation that frontier models execute internally within their extended-thinking trace during successful one-shot whole-grid runs.
    \item \textbf{Calibration and timeout envelopes differ sharply across the evaluated model snapshots.}
    Multi-model ensembles cannot apply uniform confidence thresholds or timeout caps across heterogeneous reasoning models (\Cref{fig:calibration}).
    Because Gemini reports binary confidence ($99.8\%$ of responses in $[90,100]$, rendering $\tau_{\text{anchor}}=50$ inactive on status-\texttt{ok} calls) and suffers a degraded confabulation tail between $90\text{s}$ and $180\text{s}$ ($38.2\%$ accuracy at median confidence $99$), it requires a strict $90\text{s}$ timeout cap.
    Conversely, because Claude reports graded confidence with a steep drop below its $[80,89]$ bin (motivating the selected operating floor $\tau_{\text{peer}}=85$) and a productive $90$--$150\text{s}$ deep-reasoning tail ($4.1\%$ of all solves), it requires a strict confidence floor paired with a generous $150\text{s}$ timeout cap.
\end{itemize}

\section{Future Work: Limitations and the Final Frontier}
\label{sec:future-work}

In a post-championship interview with \textit{The Times}~\cite{times2026championship}, we remarked that \Darmok{}'s competitive mission is complete: we set out to test whether an automated system could match and outpace champion humans on full British cryptic crosswords under live tournament rules, and that milestone has now been demonstrated.
Indeed, our own single-shot frontier experiments (\cref{sec:tail-eval}) illustrate the trajectory of Sutton's \textit{Bitter Lesson}~\cite{sutton2019bitter}: task-specific neuro-symbolic scaffolding pioneers a capability milestone and exposes the governing domain invariants, only for the next generation of general-purpose reasoning models to absorb the loop end-to-end (already visible in \texttt{gpt-6-astra}'s $21/21$ one-shot sweep on standard $15\times 15$ championship grids).
Operational refinements to \Darmok{}'s cloud harness are straightforward: for example, admitting a corroboration-consistent boundary-feed grid immediately upon arrival rather than waiting for the next pass barrier (which would have saved $72.9\text{s}$ of barrier wait in the Final and $136.5\text{s}$ in Semi-Final~2), using streaming token-rate telemetry to trip pass barriers early during provider brownouts, or running matched-window randomized fault-injection trials. Beyond these engineering improvements, four deeper scientific and systems challenges define the path ahead.

\begin{enumerate}[leftmargin=1.8em,itemsep=3pt]
    \item \textbf{From Barred \textit{Mephisto} Grids to Thematic Meta-Puzzles (\textit{The Listener}).}
    Our zero-shot transfer study on barred $12\times 12$ \textit{Mephisto} grids ($350/360 = 97.2\%$, $5/10$ perfect grids; \cref{sec:mephisto-eval}) confirmed that \Darmok{}'s crossing-constraint core generalizes directly to barred topologies without modifying prompts or wordlists, leaving \textit{Chambers}-specific orthographic variants on unchecked squares (\cref{sec:error-analysis}) and occasional \textit{compound} or \textit{subtractive anagrams} as residual failure modes.
    At the same time, standalone single-shot models still struggle on \textit{Mephisto} ($17/20$ invocations of \texttt{gpt-6-astra} failing to finish before the per-clue pipeline), showing that high lexical obscurity and dense wordplay remain a bottleneck for unassisted single-call reasoning even when per-word checking density is high ($78.6\%$ mean).
    Beyond standard barred grids, the next major solving frontier lies in barred \textit{thematic} crosswords such as \textit{The Listener} (established 1930 in \textit{The Listener} and now published weekly in \textit{The Times}), \textit{The Inquisitor} (\textit{The i Paper}), and \textit{The Magpie}~\cite{manley2014chambers}.
    In a \textit{Listener} puzzle, a prose preamble defines a bespoke meta-algorithm that interacts with some subset of the clues and/or grid entries: clues may contain unannounced single-letter misprints in the definition whose corrected letters spell a hidden instruction; answers may require encoding via a Playfair square, numerical cipher, or thematic substitution prior to grid entry; completed grids may conceal a visual or thematic \textit{Nina}\footnote{Named after American caricaturist Al Hirschfeld, who hid his daughter Nina's name inside his illustrations; in British crosswords, a \textit{Nina} is an unannounced message or theme spelled out across unchecked or perimeter squares of the finished grid.} that dictates a final spatial transformation; or the grid may be presented \textit{carte blanche} without bars or clue numbers, requiring the solver to infer the grid geometry and the wordplay simultaneously.
    Because setters continuously invent novel preamble gimmicks and meta-rules, solving \textit{Listener}-class puzzles requires genuine zero-shot synthesis of puzzle-level algorithms rather than a fixed orchestration pipeline.

    \item \textbf{Closing the ``20-Watt Wetware'' Gap: Air-Gapped Local Solving.}
    As we observed after the Championship Final, a human finalist brings one mind, a pencil, and roughly $15$--$30\text{W}$ of biological wetware~\cite{raichle2002appraising} to the competition floor; \Darmok{} brings four frontier cloud models, three active-active replicas, curated local wordlists, and megawatts of datacenter compute, while remaining exposed to commercial API brownouts (\Cref{fig:brownout}).
    Engineering a fully self-contained, network-free solver that runs locally on a standard laptop or desktop workstation (e.g., a 128\,GB unified-memory Framework Desktop~\cite{framework2025desktop}), perhaps by distilling or fine-tuning an open-weights reasoning model to achieve championship-caliber accuracy and sub-10-minute latency without external network calls, is a substantial systems and model-compression challenge, and likely the only milestone that would justify fielding a successor to \Darmok{} in a future live competition.

    \item \textbf{Formalizing the Generative Grammar of ``Crossword English''.}
    Midway through our summer development, we started an exploratory workstream on learning a probabilistic context-free grammar (PCFG) over historical clue--parse corpora, but parked it once our diagnostic experiments showed that crossing-constraint tightening and multi-seat corroboration sufficed to reach the championship frontier.
    Yet solving via black-box extended-thinking traces is recognition without a formal generative theory.
    Even when frontier models solve a spatially or typographically devious clue, such as clue~2D in the 2026 Championship Final (\clue{Wolf trap missing its lowest crosspiece? (5)} $\to$ \ans{SNARF}; Clue~C15, \Cref{tab:app-clues-2}), which tripped up three of the five human finalists, their reasoning lives inside an unstructured natural-language trace.
    Constructing a rigorous, compositional grammar of ``Crossword English'' (analogous to P\={a}\d{n}ini's generative grammar for Sanskrit~\cite{kiparsky2009panini}) that captures token-span coverage, explicit vs.\ indirect fodder typing across strict Ximenean~\cite{ximenes1966art} and libertarian Araucarian~\cite{connor2013twogirls} traditions, nested string operators, and lexicon-backed synonym judgments without the brittleness of ad-hoc Python-AST scripts (\cref{sec:verifier-falsification}) remains a foundational open problem in computational linguistics.
    When coupled with a curated lexicon ($\Gamma$) and a preference model over competing parse attachments, such a grammar provides both a principled tie-breaker for single-letter spelling variants on unchecked cells (\cref{sec:error-analysis}) and the formal backbone required for clue generation (below).

    \item \textbf{The ``Olympus Mons'' Challenge: Master-Caliber Cryptic Setting under Data Scarcity.}
    If matching human champions at solving cryptic crosswords was the Mount Everest challenge, creating an automated cryptic setter whose clues and grids are indistinguishable from those of a \textit{Times} master setter is the Olympus Mons challenge.
    Current frontier LLMs are poor cryptic setters for two structural reasons:
    \begin{enumerate}[label=(\alph*),leftmargin=1.6em,itemsep=1pt]
        \item \textit{Bilevel grammatical synthesis.}
        Generating a valid, witty cryptic clue requires satisfying two independent grammars in a single token sequence (\Cref{fig:primer-clues}).
        Working backward from a target entry (e.g., \ans{ESTIMATION}), a master setter constructs a sound wordplay AST while ensuring that the surface tokens read naturally as a cohesive, topical English sentence that conceals the boundary between definition and wordplay. For example, \textit{Times} Puzzles Editor Mick Hodgkin's topical clue commemorating \Darmok{}~\cite{times2026championship}, \clue{AI not foxed by Times cryptic --- respect! (10)} (Clue~C29, \Cref{tab:app-clues-3}), places an anagram (\textit{cryptic}) of \texttt{TIMES} ($\ans{ESTIM}$) before (\textit{by}) an anagram (\textit{foxed}) of \texttt{AI NOT} ($\ans{ATION}$) to define \textit{respect!}.
        Standard left-to-right autoregressive generation optimizes surface fluency at the expense of exact character-level AST soundness, or vice versa.
        \item \textit{Severe corpus scarcity.}
        Unlike general natural language, the entire modern digital archive of cryptic crosswords since 2000\footnote{While clues, solutions, and annotations from earlier puzzles are still available, rules have changed significantly. Even \texttt{ER} has now become \texttt{CR}.} comprises only a few million clues across all major British broadsheets (roughly $250{,}000$ clues in \textit{The Times} archives alone), of which perhaps $10^5$ to $10^6$ are paired with verified human wordplay parses, still orders of magnitude smaller than foundation-model pre-training corpora.
    \end{enumerate}
    At the same time, setting relaxes one central constraint that dominated \Darmok{}'s engineering: \textit{there is no tournament clock}.
    Whereas a championship solver must solve $30$ clues in under seven minutes across live API barriers, a human setter may spend hours or days crafting a single puzzle.
    Because setting is an offline problem, an automated setter can expend orders of magnitude more compute per clue without wall-clock latency pressure.
    It can combine formal grammatical enumeration (Point~3), deep tree search over candidate wordplay decompositions, neural surface synthesis (potentially using discrete diffusion models rather than left-to-right autoregression to satisfy surface and wordplay constraints jointly), and solver-in-the-loop verification (using a system like \Darmok{} itself as an adversarial critic to reject unsound or unidiomatic drafts).
\end{enumerate}

\section{Conclusion}
\label{sec:conclusion}

From \textsc{Proverb}~\cite{shazeer1999aaai,littman2002proverb} and \textsc{Dr.Fill}~\cite{ginsberg2011drfill} to the Berkeley Crossword Solver~\cite{wallace2022berkeley}, tournament crossword solving has tracked the progression of artificial intelligence from heuristic search and probabilistic graphical models to neural candidate retrieval.
British cryptic crosswords remained a prominent holdout for more than two decades because sparse grids and adversarial, nested wordplay defeated static retrieval CSPs.
\Darmok{}'s $210/210$ sweep of the 55th \textit{Times} Crossword Championship in London, its $99.84\%$ accuracy across $1{,}261$ post-cutoff daily clues in its frozen epoch ($99.17\%$ across all $3{,}876$ daily campaign clues), and its $97.2\%$ zero-shot transfer to barred \textit{Mephisto} grids demonstrate championship-caliber performance on this long-standing challenge.

Yet the broader systems lesson of our post-championship study extends beyond crosswords.
As frontier models cross the capability threshold on complex reasoning tasks, the bottleneck in deployed AI systems shifts from average-case single-call capability to tail-regime invariants: surviving provider brownouts and client network bursts, aborting late-tail calibration collapse, blocking uncorroborated confabulations, and guaranteeing global state coherence across asynchronous calls.
Engineering systems that meet strict accuracy--latency objectives does not become obsolete when models grow stronger; rather, it transforms AI orchestration from the art of coaxing capability out of weak models into the discipline of building resilient, high-reliability systems out of capable but volatile ones.

\subsection*{Acknowledgments}
\label{sec:acknowledgments}

We thank \textit{The Times} puzzles team---Mick Hodgkin, Jason Crampton, and Peter Biddlecombe---for their ongoing encouragement, strategic skepticism, and unstinting support in allowing \Darmok{} to participate in the 55th \textit{Times} Crossword Championship and for their generous hospitality and crossword discussions, and John Guiver for his willingness to share his master setter's insights on cruciverbalist AIs. 
We also thank Kyle Mahowald and Nick Tomlin for multiple helpful technical discussions, and the students who contributed to early experiments and data wrangling tasks: Dev Aggarwal, Arya Agiwal, Saatvik Barla, Nishant Bhakar, Jaxon Dial, and Om Goswami.
Jason Crampton, Mick Hodgkin, Peter Stone, and Nick Tomlin read and provided valuable feedback on earlier versions of this document.

\subsection*{Disclosure of AI Usage in Research and Manuscript Preparation}
\label{sec:ai-disclosure}

In accordance with current scientific publishing standards (ACM, ACL, NeurIPS, and \textit{Nature} authorship and transparency guidelines), we disclose two distinct roles of generative AI systems in this work:
\begin{enumerate}[leftmargin=1.8em,itemsep=2pt]
    \item \textbf{AI Systems as the Object of Study.}
    The \Darmok{} solver orchestrates commercial large language model APIs (\texttt{gemini-3.1-pro-preview}, \texttt{claude-fable-5}, \texttt{gpt-5.6-sol}, and \texttt{gpt-6-astra}, alongside baseline comparisons against \texttt{gemini-4-argon} and \texttt{gemini-3.8-flash}) as experimental subjects and system components.
    Model codenames (\texttt{claude-fable-5}, \texttt{gpt-5.6-sol}, \texttt{gpt-6-astra}, \texttt{gemini-4-argon}, and \texttt{gemini-3.8-flash}) correspond to vendor-embargoed frontier model snapshots evaluated between May and October 2026; exact provider snapshot strings and API identifiers will be unmasked in the camera-ready/updated arXiv version immediately upon expiration of the respective vendor embargoes.
    All prompts, model configurations, timeout envelopes, token budgets, and call-level logs are archived in the project repository (\url{https://github.com/Darmok-Jalad-Tanagra/darmok-v2}, to be made public upon release) for full auditability.
    \item \textbf{AI Assistance in Software Engineering and Manuscript Preparation.}
    Coding assistants and large language models (Anthropic Claude and Google Gemini) were used under the author's direct supervision to assist with software implementation, experimental harness scripting, telemetry log aggregation, data visualization (\texttt{TikZ}/\texttt{pgfplots}/\texttt{matplotlib}), and manuscript editing and formatting.
    All architectural designs, experimental hypotheses, pre-registered ablation protocols, statistical analyses, and scientific claims were conceived, verified, and approved by the human author, who takes full responsibility for the integrity and accuracy of the manuscript.
\end{enumerate}

\clearpage
\bibliographystyle{plain}
\bibliography{references}

\clearpage
\appendix
\section{Complete Experimental Reference Tables}
\label{app:tables}

To preserve full numerical reproducibility while keeping the main text visually accessible, this appendix provides the complete tabular data corresponding to the experiments in \cref{sec:folklore,sec:eval}.

\begin{table}[!htbp]
\centering
\caption{Per-clue solve rate across the 16-puzzle test set ($N = 463$; \cref{sec:crossings-amp} and \Cref{fig:folklore}a), stratified by cold vs.\ constrained dispatch (Panel~A), known crossing letters $n_{\text{known}}$ (Panel~B), and crossing density $n_{\text{known}} / \ell_i$ (Panel~C).}
\label{tab:app-crossings}
\small
\begin{adjustbox}{max width=\textwidth}
\begin{tabular}{@{}lrrrrr@{}}
\toprule
\textbf{Condition / Stratum} & \textbf{Clues ($n$)} & \textbf{Gemini 3.1 Pro} & \textbf{Claude} & \textbf{GPT-5} & \textbf{Any Seat (Oracle)} \\
\midrule
\multicolumn{6}{@{}l}{\textit{Panel A: Cold vs.\ Crossing-Constrained Dispatch}} \\
Unconstrained / Cold ($>100$ matches) & 143 & $91$ ($63.6\%$) & $63$ ($44.1\%$) & $68$ ($47.6\%$) & --- \\
Constrained ($\le 100$ matches)       & 320 & $309$ ($96.6\%$) & $302$ ($94.4\%$) & $271$ ($84.7\%$) & --- \\
Full Test Set (All Clues)             & 463 & $400$ ($86.4\%$) & $365$ ($78.8\%$) & $339$ ($73.2\%$) & $427$ ($92.2\%$) \\
All Three Seats Correct (Consensus)   & 463 & \multicolumn{4}{c}{$298 / 463$ ($64.4\%$); \quad Unique solves: Gemini $31$, GPT-5 $9$, Claude $8$} \\
\midrule
\multicolumn{6}{@{}l}{\textit{Panel B: Stratified by Number of Known Crossing Letters $n_{\text{known}}$}} \\
$n_{\text{known}} = 0$ &   9 & $7$ ($77.8\%$) & $7$ ($77.8\%$) & $4$ ($44.4\%$) & $8$ ($88.9\%$) \\
$n_{\text{known}} = 1$ &  32 & $18$ ($56.3\%$) & $10$ ($31.3\%$) & $18$ ($56.3\%$) & $23$ ($71.9\%$) \\
$n_{\text{known}} = 2$ & 131 & $98$ ($74.8\%$) & $77$ ($58.8\%$) & $74$ ($56.5\%$) & $108$ ($82.4\%$) \\
$n_{\text{known}} = 3$ & 156 & $146$ ($93.6\%$) & $143$ ($91.7\%$) & $127$ ($81.4\%$) & $153$ ($98.1\%$) \\
$n_{\text{known}} = 4$ &  77 & $75$ ($97.4\%$) & $74$ ($96.1\%$) & $68$ ($88.3\%$) & $77$ ($100.0\%$) \\
$n_{\text{known}} = 5$ &  47 & $46$ ($97.9\%$) & $43$ ($91.5\%$) & $39$ ($83.0\%$) & $47$ ($100.0\%$) \\
$n_{\text{known}} \ge 6$ & 11 & $10$ ($90.9\%$) & $11$ ($100.0\%$) & $9$ ($81.8\%$) & $11$ ($100.0\%$) \\
\midrule
\multicolumn{6}{@{}l}{\textit{Panel C: Stratified by Crossing Letter Density ($n_{\text{known}} / \ell_i$)}} \\
$0\%$           &   9 & $77.8\%$ & $77.8\%$ & $44.4\%$ & $88.9\%$ \\
$1\%$--$25\%$   &  87 & $63.2\%$ & $42.5\%$ & $43.7\%$ & $72.4\%$ \\
$26\%$--$40\%$  &  99 & $89.9\%$ & $77.8\%$ & $78.8\%$ & $95.0\%$ \\
$41\%$--$55\%$  & 152 & $89.5\%$ & $86.8\%$ & $77.6\%$ & $96.1\%$ \\
$>55\%$         & 116 & $97.4\%$ & $96.6\%$ & $87.1\%$ & $100.0\%$ \\
\bottomrule
\end{tabular}
\end{adjustbox}
\end{table}

\begin{table}[!htbp]
\centering
\caption{Puzzle identities, publication dates, \textsc{Snitch} ratings, replica budgets ($D$), and solve scores on the 16-puzzle stratified test set ($N = 463$ clues; \cref{sec:milestone,sec:corpus-eval}).}
\label{tab:app-test-puzzles}
\footnotesize
\setlength{\tabcolsep}{4.5pt}
\begin{adjustbox}{max width=\textwidth}
\begin{tabular}{@{}llllrrccc@{}}
\toprule
\textbf{\textit{Times} \#} & \textbf{Pub.\ Date} & \textbf{Day} & \textbf{Difficulty Band} & \textbf{\textsc{Snitch}} & \textbf{Clues} & \textbf{Budget} & \textbf{2025 V1} & \textbf{Champ.\ V2 (Replicas)} \\
\midrule
29498$^*$ & 2026-03-24 & Tue & Easy ($70$--$90$)      &  71 & 30 & $D=1$ & $30/30$ & $\mathbf{30/30}$ (single run) \\
\midrule
29360     & 2025-10-14 & Tue & Moderate ($90$--$110$) &  90 & 30 & $D=1$ & $29/30$ & $\mathbf{30/30}$ (single run) \\
29350     & 2025-10-02 & Thu & Moderate ($90$--$110$) & 106 & 28 & $D=1$ & $22/28$ & $\mathbf{28/28}$ (single run) \\
29446     & 2026-01-22 & Thu & Moderate ($90$--$110$) & 109 & 29 & $D=1$ & $24/29$ & $\mathbf{29/29}$ (single run) \\
\midrule
29283     & 2025-07-16 & Wed & Hard ($110$--$130$)    & 110 & 31 & $D=1$ & $28/31$ & $\mathbf{31/31}$ (single run) \\
29415     & 2025-12-17 & Wed & Hard ($110$--$130$)    & 113 & 28 & $D=1$ & $26/28$ & $\mathbf{28/28}$ (single run) \\
29416     & 2025-12-18 & Thu & Hard ($110$--$130$)    & 113 & 28 & $D=1$ & $24/28$ & $\mathbf{28/28}$ (single run) \\
29470$^*$ & 2026-02-19 & Thu & Hard ($110$--$130$)    & 118 & 29 & $D=1$ & $25/29$ & $\mathbf{29/29}$ (single run) \\
\midrule
29440     & 2026-01-15 & Thu & Very Hard ($130$--$160$) & 137 & 28 & $D=3$ & $22/28$ & $\mathbf{28/28}$ ($28 / 28 / 28$) \\
29464$^*$ & 2026-02-12 & Thu & Very Hard ($130$--$160$) & 138 & 28 & $D=3$ & $28/28$ & $\mathbf{28/28}$ ($28 / 28 / 28$) \\
29429     & 2026-01-02 & Fri & Very Hard ($130$--$160$) & 144 & 30 & $D=3$ & $25/30$ & $\mathbf{30/30}$ ($30 / 30 / 30$) \\
29405     & 2025-12-05 & Fri & Very Hard ($130$--$160$) & 145 & 28 & $D=3$ & $19/28$ & $\mathbf{28/28}$ ($28 / 28 / 28$) \\
29303     & 2025-08-08 & Fri & Very Hard ($130$--$160$) & 148 & 28 & $D=3$ & $14/28$ & $\mathbf{28/28}$ ($28 / 28 / 28$) \\
\midrule
29447     & 2026-01-23 & Fri & Extreme ($160+$)       & 168 & 30 & $D=3$ & $26/30$ & $\mathbf{30/30}$ ($30 / 30 / 30$) \\
29399     & 2025-11-28 & Fri & Extreme ($160+$)       & 184 & 30 & $D=3$ & $27/30$ & $\mathbf{29/30}$ ($29 / 29 / 28$)$^\dagger$ \\
29435     & 2026-01-09 & Fri & Extreme ($160+$)       & 231 & 28 & $D=3$ & $11/28$ & $\mathbf{27/28}$ ($28 / 26 / 27$)$^\dagger$ \\
\midrule
\textbf{Total (16)} & \multicolumn{3}{@{}l}{\textbf{Jul 16, 2025 -- Mar 24, 2026}} & \textbf{71--231} & \textbf{463} & \textbf{32 runs} & $\mathbf{380/463}$ ($\mathbf{82.1\%}$) & $\mathbf{461/463}$ ($\mathbf{99.57\%}$) \\
\bottomrule
\multicolumn{9}{@{}l}{\footnotesize $^*$Post-dates Claude Fable~5's January~31, 2026 cutoff ($87$ clues across \texttt{29464}, \texttt{29470}, and \texttt{29498}); all $16$ puzzles post-date Gemini's January~31, 2025 cutoff.} \\
\multicolumn{9}{@{}l}{\footnotesize $^\dagger$First-finisher selection picked Replica~1 ($29/30$, missing 1D \ans{TAME} [Clue~C26]) on \texttt{29399} and Replica~2 ($27/28$, missing 5D \ans{RABBITY} [Clue~C27]) on \texttt{29435}.}
\end{tabular}
\end{adjustbox}
\end{table}

\clearpage
\begin{table}[!htbp]
\centering
\caption{Complete 7-arm Pass~0 prompt ablation across $N = 463$ test-set clues ($6{,}482$ scored calls; \cref{sec:prompt-ablation}) under strict JSON parsing (Panel~A) and post-hoc uniform lenient JSON repair on Claude (Panel~B).}
\label{tab:app-prompt}
\footnotesize
\setlength{\tabcolsep}{3.5pt}
\begin{adjustbox}{max width=\textwidth}
\begin{tabular}{@{}llrrrrrrrrll@{}}
\toprule
\textbf{Seat} & \textbf{Arm} & \textbf{Avail} & \textbf{Corr} & \textbf{Corr$_{\text{ans}}$} & \textbf{Adm} & \textbf{Wrong$_{\ge\tau}$} & \textbf{T/O} & \textbf{Med $t$} & \textbf{Out Tok} & \textbf{All $\Delta$ ($b/c,\, p$)} & \textbf{Ok $\Delta$ ($b/c,\, p$)} \\
\midrule
\multicolumn{12}{@{}l}{\textit{Panel A: Pre-Registered Strict JSON Parser (All Arms; Gemini Identical Under Lenient Repair)}} \\
\multirow{7}{*}{\shortstack[l]{\textbf{Gemini}\\($\tau=50$)}}
 & \texttt{P0} (Full)        & $71.7\%$ & $70.6\%$ & $98.5\%$ & $70.6\%$ & $1.5\%$ & $25.1\%$ & $21.1\text{s}$ & 194 & --- & --- \\
 & \texttt{P0'} (Replicate)  & $71.5\%$ & $70.2\%$ & $98.2\%$ & $70.2\%$ & $0.9\%$ & $26.1\%$ & $22.3\text{s}$ & 193 & $-0.43$ ($30/28, 0.90$) & $\phantom{+}0.00$ ($0/0, 1.00$) \\
 & \texttt{P1} (Ans-First)   & $75.4\%$ & $71.3\%$ & $94.6\%$ & $71.3\%$ & $1.5\%$ & $23.5\%$ & $23.9\text{s}$ & 201 & $+0.65$ ($30/33, 0.80$) & $-0.66$ ($3/1, 0.63$) \\
 & \texttt{P2} (No-Revise)   & $74.3\%$ & $72.6\%$ & $97.7\%$ & $72.6\%$ & $1.2\%$ & $23.1\%$ & $21.8\text{s}$ & 183 & $+1.94$ ($29/38, 0.33$) & $+0.33$ ($0/1, 1.00$) \\
 & \texttt{P3} (No-Decomp)   & $73.2\%$ & $71.3\%$ & $97.3\%$ & $70.8\%$ & $1.5\%$ & $24.2\%$ & $22.0\text{s}$ & 199 & $+0.22$ ($32/33, 1.00$) & $-1.66$ ($5/0, 0.063$) \\
 & \texttt{P4} (No-Guide)    & $70.6\%$ & $69.8\%$ & $98.8\%$ & $69.8\%$ & $0.6\%$ & $26.3\%$ & $20.4\text{s}$ & 190 & $-0.86$ ($28/24, 0.68$) & $+0.33$ ($0/1, 1.00$) \\
 & \texttt{P5} (Minimal)     & $83.2\%$ & $74.9\%$ & $90.1\%$ & $73.4\%$ & $2.9\%$ & $15.3\%$ & $17.4\text{s}$ &  34 & $+2.81$ ($25/38, 0.13$) & $\mathbf{-2.86}$ ($\mathbf{9/0, 0.004}$) \\
\cmidrule{1-12}
\multirow{7}{*}{\shortstack[l]{\textbf{Claude}\\Strict\\($\tau=85$)}}
 & \texttt{P0} (Full)        & $85.7\%$ & $81.4\%$ & $95.0\%$ & $79.5\%$ & $2.4\%$ & $13.2\%$ & $14.1\text{s}$ & $1{,}861$ & --- & --- \\
 & \texttt{P0'} (Replicate)  & $83.2\%$ & $78.8\%$ & $94.8\%$ & $78.0\%$ & $3.0\%$ & $14.9\%$ & $14.1\text{s}$ & $1{,}682$ & $-1.51$ ($39/32, 0.48$) & $-0.28$ ($9/8, 1.00$) \\
 & \texttt{P1} (Ans-First)   & $86.8\%$ & $83.2\%$ & $95.8\%$ & $81.6\%$ & $2.1\%$ & $13.0\%$ & $14.4\text{s}$ & $1{,}754$ & $+2.16$ ($26/36, 0.25$) & $+0.27$ ($5/6, 1.00$) \\
 & \texttt{P2} (No-Revise)   & $84.0\%$ & $80.3\%$ & $95.6\%$ & $79.0\%$ & $2.1\%$ & $14.9\%$ & $13.7\text{s}$ & $1{,}571$ & $-0.43$ ($24/22, 0.88$) & $-0.27$ ($4/3, 1.00$) \\
 & \texttt{P3} (No-Decomp)   & $80.1\%$ & $76.2\%$ & $95.1\%$ & $74.7\%$ & $2.3\%$ & $13.2\%$ & $12.7\text{s}$ & $1{,}596$ & $\mathbf{-4.75}$ ($\mathbf{49/27, 0.015}$) & $-0.58$ ($6/4, 0.75$) \\
 & \texttt{P4} (No-Guide)    & $85.1\%$ & $79.9\%$ & $93.9\%$ & $78.2\%$ & $2.4\%$ & $12.1\%$ & $12.3\text{s}$ & $1{,}600$ & $-1.30$ ($31/25, 0.50$) & $-0.54$ ($8/6, 0.79$) \\
 & \texttt{P5} (Minimal)     & $89.0\%$ & $82.7\%$ & $93.0\%$ & $81.4\%$ & $4.3\%$ & $11.0\%$ &  $8.0\text{s}$ & $1{,}325$ & $+1.94$ ($21/30, 0.26$) & $-1.04$ ($11/7, 0.48$) \\
\midrule
\multicolumn{12}{@{}l}{\textit{Panel B: Post-Hoc Uniform Lenient JSON Parser Across All Claude Arms (Recovering Trailing Commas \& Missing \texttt{confidence} Braces)}} \\
\multirow{7}{*}{\shortstack[l]{\textbf{Claude}\\Uniform\\Lenient\\($\tau=85$)}}
 & \texttt{P0} ($+5$ rec., $+4$ adm)   & $86.8\%$ & $82.3\%$ & $94.8\%$ & $80.3\%$ & $2.4\%$ & $13.2\%$ & $14.1\text{s}$ & $1{,}861$ & --- & --- \\
 & \texttt{P0'} ($+9$ rec., $+8$ adm)  & $85.1\%$ & $80.6\%$ & $94.7\%$ & $79.7\%$ & $2.9\%$ & $14.9\%$ & $14.1\text{s}$ & $1{,}682$ & $-0.65$ ($32/29, 0.80$) & $-0.27$ ($9/8, 1.00$) \\
 & \texttt{P1} ($+1$ rec., $+1$ adm)   & $87.0\%$ & $83.4\%$ & $95.8\%$ & $81.9\%$ & $2.1\%$ & $13.0\%$ & $14.4\text{s}$ & $1{,}754$ & $+1.51$ ($25/32, 0.43$) & $+0.27$ ($5/6, 1.00$) \\
 & \texttt{P2} ($+0$ rec., $4$ trunc)  & $84.0\%$ & $80.3\%$ & $95.6\%$ & $79.0\%$ & $2.1\%$ & $14.9\%$ & $13.7\text{s}$ & $1{,}571$ & $-1.30$ ($24/18, 0.44$) & $-0.27$ ($4/3, 1.00$) \\
 & \texttt{P3} ($+30$ rec., $+29$ adm) & $86.6\%$ & $82.5\%$ & $95.3\%$ & $81.0\%$ & $2.3\%$ & $13.2\%$ & $12.7\text{s}$ & $1{,}596$ & $+0.65$ ($25/28, 0.78$) & $+0.27$ ($6/7, 1.00$) \\
 & \texttt{P4} ($+13$ rec., $+11$ adm) & $87.9\%$ & $82.3\%$ & $93.6\%$ & $80.6\%$ & $2.6\%$ & $12.1\%$ & $12.3\text{s}$ & $1{,}600$ & $+0.22$ ($23/24, 1.00$) & $-0.78$ ($9/6, 0.61$) \\
 & \texttt{P5} ($+0$ rec.)             & $89.0\%$ & $82.7\%$ & $93.0\%$ & $81.4\%$ & $4.3\%$ & $11.0\%$ &  $8.0\text{s}$ & $1{,}325$ & $+1.08$ ($21/26, 0.56$) & $-1.03$ ($11/7, 0.48$) \\
\bottomrule
\multicolumn{12}{@{}l}{\footnotesize \textbf{Avail}: status-\texttt{ok} rate ($/463$); \textbf{Corr}: unconditional correct rate ($/463$); \textbf{Corr$_{\text{ans}}$}: accuracy given status \texttt{ok}; \textbf{Adm}: admissible rate; \textbf{Wrong$_{\ge\tau}$}: error rate} \\
\multicolumn{12}{@{}l}{\footnotesize at floor ($n_{\text{wrong},\ge\tau}/n_{\text{at\_floor}}$); \textbf{Out Tok}: mean output tokens (provider counter: excludes thinking tokens on Gemini, includes thinking tokens on Claude);} \\
\multicolumn{12}{@{}l}{\footnotesize \textbf{All $\Delta$ ($b/c, p$)} and \textbf{Ok $\Delta$ ($b/c, p$)}: paired admissible accuracy changes, discordant counts ($b=\texttt{P0}$-only, $c=\text{arm}$-only), and exact McNemar $p$-values.} \\
\multicolumn{12}{@{}l}{\footnotesize Integer admissible counts ($/463$) for Claude strict $\to$ lenient are: \texttt{P0} $368 \to 372$, \texttt{P0'} $361 \to 369$, \texttt{P1} $378 \to 379$, \texttt{P2} $366 \to 366$, \texttt{P3} $346 \to 375$ ($+29$),} \\
\multicolumn{12}{@{}l}{\footnotesize \texttt{P4} $362 \to 373$ ($+11$), and \texttt{P5} $377 \to 377$. On Gemini \texttt{P5}, unconditional answer correctness on jointly completed calls has $\text{Ok }\Delta = -1.59$ ($5/0, p=0.0625$).}
\end{tabular}
\end{adjustbox}
\end{table}

\begin{table}[!htbp]
\centering
\caption{Confidence-elicitation ablation ($N = 463$ clues, $2{,}778$ calls; \cref{sec:conf-ablation}), reporting unconditional correctness (\textbf{Corr$_{\text{all}}$}), status-\texttt{ok} accuracy (\textbf{Corr$_{\text{ans}}$}), paired differences (\textbf{All $\Delta$}, \textbf{Ok $\Delta$}) with discordant counts ($b/c$) and exact McNemar $p$-values, and exact answer-string identity over jointly completed pairs ($n_{\text{both-ok}}$).}
\label{tab:app-conf}
\footnotesize
\setlength{\tabcolsep}{3.5pt}
\begin{adjustbox}{max width=\textwidth}
\begin{tabular}{@{}llrrrrlllr@{}}
\toprule
\textbf{Seat} & \textbf{Arm / Contrast} & \textbf{Corr$_{\text{all}}$} & \textbf{Corr$_{\text{ans}}$} & \textbf{T/O} & \textbf{Parse Fail} & \textbf{Med $t$} & \textbf{All $\Delta$ ($b/c,\, p$)} & \textbf{Ok $\Delta$ ($b/c,\, p$)} & \textbf{String Identity ($n_{\text{both-ok}}$)} \\
\midrule
\multirow{6}{*}{\textbf{Gemini}}
 & \texttt{P0r2} (Full w/ Conf.)        & $74.1\%$ & $97.4\%$ & $19.9\%$ & 17 & $23.7\text{s}$ & --- & --- & --- \\
 & \texttt{P6} vs.\ \texttt{P0r2}       & $76.0\%$ & $96.7\%$ & $17.7\%$ & 16 & $21.7\text{s}$ & $+1.94$ ($28/37, 0.32$) & $\phantom{+}0.00$ ($2/2, 1.00$) & $\mathbf{98.4\%}$ ($317 / 322$) \\
 & \textit{\texttt{P6} vs.\ \texttt{P2} (No-Rev.)}$^*$ & --- & --- & --- & --- & --- & $+3.46$ ($30/46, 0.085$) & $-0.32$ ($2/1, 1.00$) & $\mathbf{98.7\%}$ ($307 / 311$) \\
 & \texttt{P7} vs.\ \texttt{P0r2}       & $74.9\%$ & $85.9\%$ & $11.2\%$ &  7 & $15.5\text{s}$ & $+0.86$ ($33/37, 0.72$) & $-3.94$ ($13/0, 0.0002$) & $95.2\%$ ($314 / 330$)$^\dagger$ \\
 & \textit{\texttt{P7} vs.\ \texttt{P5} (Min.)} & --- & --- & --- & --- & --- & $\phantom{+}0.00$ ($35/35, 1.00$) & $-1.94$ ($16/9, 0.23$) & $90.0\%$ ($324 / 360$) \\
 & \textit{\texttt{P0r2} vs.\ \texttt{P0} (Drift)}$^*$ & --- & --- & --- & --- & --- & $+3.46$ ($21/37, 0.048$) & $-0.32$ ($1/0, 1.00$) & $99.7\%$ ($308 / 309$) \\
\midrule
\multirow{8}{*}{\textbf{Claude}}
 & \texttt{P0r2} (Strict / Lenient)$^\ddagger$ & $80.3\% / 82.5\%$ & $95.6\% / 95.5\%$ & $13.4\%$ & $12 / 1$ & $13.6\text{s}$ & --- & --- & --- \\
 & \texttt{P6} vs.\ \texttt{P0r2} (Strict) & $81.6\%$ & $95.2\%$ & $14.0\%$ &  0 & $12.0\text{s}$ & $+1.30$ ($29/35, 0.53$) & $-0.55$ ($7/5, 0.77$) & $\mathbf{95.9\%}$ ($350 / 365$) \\
 & \texttt{P6} vs.\ \texttt{P0r2} (Lenient)$^\ddagger$ & $81.6\%$ & $95.2\%$ & $14.0\%$ & 0 & $12.0\text{s}$ & $-0.86$ ($28/24, 0.68$) & $-0.53$ ($7/5, 0.77$) & $\mathbf{96.0\%}$ ($360 / 375$) \\
 & \textit{\texttt{P6} vs.\ \texttt{P2} (No-Rev.)} & --- & --- & --- & --- & --- & $+1.30$ ($23/29, 0.49$) & $-0.27$ ($4/3, 1.00$) & $\mathbf{96.8\%}$ ($358 / 370$) \\
 & \texttt{P7} vs.\ \texttt{P0r2} (Strict) & $84.2\%$ & $92.9\%$ &  $8.9\%$ &  0 &  $8.3\text{s}$ & $+3.89$ ($25/43, 0.038$) & $-2.13$ ($12/4, 0.077$) & $94.1\%$ ($353 / 375$) \\
 & \texttt{P7} vs.\ \texttt{P0r2} (Lenient)$^\ddagger$ & $84.2\%$ & $92.9\%$ &  $8.9\%$ &  0 &  $8.3\text{s}$ & $+1.73$ ($25/33, 0.36$) & $-2.08$ ($12/4, 0.077$) & $94.3\%$ ($363 / 385$) \\
 & \textit{\texttt{P7} vs.\ \texttt{P5} (Min.)} & --- & --- & --- & --- & --- & $+1.51$ ($21/28, 0.39$) & $+0.76$ ($7/10, 0.63$) & $93.7\%$ ($372 / 397$) \\
 & \textit{\texttt{P0r2} vs.\ \texttt{P0} (Drift)} & --- & --- & --- & --- & --- & $-1.08$ ($36/31, 0.63$) & $+0.84$ ($2/5, 0.45$) & $96.7\%$ ($347 / 359$) \\
\bottomrule
\multicolumn{10}{@{}l}{\footnotesize $^*$Because \texttt{P2} was run in the \texttt{E15} session (Gemini T/O $23.1\%$, \texttt{P0} T/O $25.1\%$) and \texttt{P6}/\texttt{P0r2} in \texttt{E15b} (T/O $17.7\% / 19.9\%$), the cross-session $+3.46\text{ pt}$ $\text{All }\Delta$ on} \\
\multicolumn{10}{@{}l}{\footnotesize \phantom{$^*$}Gemini \texttt{P6} vs.\ \texttt{P2} reflects the same ambient timeout shift as \texttt{P0r2} vs.\ \texttt{P0} ($+3.46\text{ pts}$); on jointly completed calls, $\text{Ok }\Delta = -0.32\text{ pts}$ ($2/1, p=1.00$).} \\
\multicolumn{10}{@{}l}{\footnotesize $^\dagger$Computed strictly over the $n=330$ jointly status-\texttt{ok} pairs (an earlier draft quoted $339/354 = 95.8\%$ over all non-empty string pairs prior to status-\texttt{ok} filtering).} \\
\multicolumn{10}{@{}l}{\footnotesize $^\ddagger$On Claude \texttt{P0r2}, uniform lenient JSON parsing recovers $11$ of $12$ syntax errors ($10$ correct $\to 382/463 = 82.5\%$); \texttt{P6} and \texttt{P7} have $0$ syntax errors.}
\end{tabular}
\end{adjustbox}
\end{table}

\begin{table}[!htbp]
\centering
\caption{Audit of the symbolic Python-AST wordplay verifier of Andrews \& Witteveen~\cite{andrews2024proving} (\cref{sec:verifier-falsification}): 200-clue \textsc{Cryptonite} validation comparison from \texttt{09\_prove\_all\_samples.py} (Panel~A) and attempt-by-attempt trace of the 5-pair smoke gate upgraded to \texttt{gemini-2.5-flash} ($3$ rewrites/pair; Panel~B).}
\label{tab:app-aw}
\footnotesize
\setlength{\tabcolsep}{4.5pt}
\begin{adjustbox}{max width=\textwidth}
\begin{tabular}{@{}>{\raggedright\arraybackslash}p{2.3cm}>{\raggedright\arraybackslash}p{3.2cm}rr>{\raggedright\arraybackslash}p{4.0cm}>{\raggedright\arraybackslash}p{5.5cm}@{}}
\toprule
\multicolumn{6}{@{}l@{}}{\textbf{Panel A: Andrews \& Witteveen 200-Clue \textsc{Cryptonite} Validation Split (\texttt{09\_prove\_all\_samples.py}, lines 603--608)}} \\
\midrule
\multicolumn{2}{@{}l}{\textbf{Selection Rule}} & \multicolumn{2}{r}{\textbf{Solved / 200}} & \textbf{Mechanism} & \textbf{Audit Verdict} \\
\midrule
\multicolumn{2}{@{}l}{\texttt{answer\_proof} (Verifier Only)} & \multicolumn{2}{r}{$28 / 200$ ($14.0\%$)} & Most frequently proved candidate ($20\times 3$ tries) & Underperforms raw sample frequency by $1{:}2$ \\
\multicolumn{2}{@{}l}{\texttt{answer\_freq} (Majority Vote)} & \multicolumn{2}{r}{$56 / 200$ ($28.0\%$)} & Modal candidate string across $20$ Gemma2-9B samples & Zero-cost baseline beats symbolic verifier $2{:}1$ \\
\multicolumn{2}{@{}l}{\texttt{answer\_proof} + \texttt{answer\_freq} Fallback} & \multicolumn{2}{r}{$56 / 200$ ($28.0\%$)} & Use \texttt{answer\_proof} if non-empty, else \texttt{answer\_freq} & $0.0\text{ pt}$ lift on Val (Test: $32.5\%$ vs.\ $25.5\%$ freq) \\
\midrule
\multicolumn{6}{@{}l@{}}{\textbf{Panel B: 5-Pair Smoke Gate on \Darmok{} Hard Pool ($200$ unique pairs from $18$ V1-missed clues; \texttt{gemini-2.5-flash}, $3$ rewrites/pair)}} \\
\midrule
\textbf{Candidate (Label)} & \textbf{Clue with Oracle \texttt{\{def\}}} & \textbf{Score} & \textbf{Time} & \textbf{Verifier Error Trace (Rewrites 0 / 1 / 2)} & \textbf{Synthesized Proof Logic \& Failure Mechanism} \\
\midrule
\ans{EBRO}\newline(\textit{Correct}, \texttt{inline-3}) & \clue{Soldiers occupy a position around \{river\} (4)} & $-1$ & $01\text{m}\,07\text{s}$ & \textbf{R0, R1, R2:} \texttt{No proof(...) function defined} (\texttt{parsed.body[0]} rejected \texttt{Enum}/helper preamble) & \textbf{Post-hoc fabrication on true answer:} Missed true parse (\ans{OR} $+$ \ans{BE} reversed) and hallucinated four unattested single-letter abbreviations: \texttt{"E"} (\textit{engineer}), \texttt{"B"} (\textit{base}), \texttt{"R"} (\textit{royal}), \texttt{"O"} (\textit{round}). \\
\addlinespace[2pt]
\ans{ERNE}\newline(\textit{Plausible Wrong}, \texttt{inline-3}) & \clue{Soldiers occupy a position around \{river\} (4)} & $-1$ & $01\text{m}\,28\text{s}$ & \textbf{R0:} \texttt{AssertionError} on \texttt{('position','NE')} (leaked \texttt{'NE': north-east})\newline\textbf{R1:} \texttt{NameError: 'part1' not defined}\newline\textbf{R2:} \texttt{No proof(...) function defined} & \textbf{Coached false-positive parse:} Synthesized \texttt{"RE"} (\textit{Soldiers}) reversed (\textit{around} $\to$ \texttt{"ER"}) + \texttt{"NE"} (\textit{a position}) = \ans{ERNE}. After R0 leaked \texttt{north-east}, R1 substituted \texttt{("North-East","NE")} (\textbf{passed!}) and only failed because R1 bound a local variable \texttt{part1 = "RE"[::-1]}. \\
\addlinespace[2pt]
\ans{GO SPARE}\newline(\textit{Correct}, \texttt{inline-7}) & \clue{\{Finally snap in rage, so agitated\} (2,5)} & $-1$ & $02\text{m}\,07\text{s}$ & \textbf{R0:} \texttt{NameError: 'p\_letter' not defined} (isolated-\texttt{assert} runner dropped local vars)\newline\textbf{R1:} \texttt{AssertionError} on \texttt{('Finally', SUBSTRING)} + \texttt{NameError}\newline\textbf{R2:} \texttt{No proof(...) function defined} & \textbf{Valid human parse rejected by DSL:} R0 found the exact \textit{\&lit} parse (\texttt{"SNAP"[-1] == "P"} + \texttt{"RAGE"} + \texttt{"SO"} = \texttt{"PRAGESO"}, anagrammed via \textit{agitated}). Rejected because AST runner drops local vars and 10-action \texttt{Action} enum lacks \texttt{LAST\_LETTER}. On R2, model mocked helpers to hardcode \texttt{if test\_synonym == "GO SPARE": return True}. \\
\addlinespace[2pt]
\ans{LORD PALMERSTON}\newline(\textit{Wrong, 14 chars}, \texttt{inline-1}) & \clue{Ex-PM's fashionable \{royal post\} (15)} & $-1$ & $02\text{m}\,01\text{s}$ & \textbf{R0, R1, R2:} \texttt{No proof(...) function defined} (\texttt{parsed.body[0]} rejected \texttt{Enum}/helper preamble) & \textbf{Definition swap \& synonym hallucination:} True 15-letter answer is \ans{CHAMBERLAINSHIP}. Because the A\&W verifier strips the puzzle's \texttt{(15)} enumeration and checks only the supplied target string, it attempted a 14-letter charade \texttt{"LORD"+"PALM"+"ER"+"STON"}. \\
\addlinespace[2pt]
\ans{ESTATES}\newline(\textit{Random Wrong}, \texttt{inline-2}) & \clue{\{Risked\} neck, regularly fleeing English country houses (7)} & $-1$ & $04\text{m}\,10\text{s}$ & \textbf{R0, R1, R2:} \texttt{\#BLOCKED\#} $\to$ \texttt{Badly formed python output} & \textbf{Reasoning-budget exhaustion:} True answer is \ans{CHANCED} (\ans{CHA[NCE]D}). Exhausted $16{,}384$-token thinking+output budget across all $3$ rewrites searching for a non-existent proof for a random wrong candidate. \\
\bottomrule
\end{tabular}
\end{adjustbox}
\end{table}

\clearpage
\begin{table}[!htbp]
\centering
\caption{Epoch~E2 daily production record ($43$ post-cutoff puzzles, $1{,}261$ clues, August 26 -- October 9, 2026; \cref{sec:corpus-eval}): stage attribution (Panel~A) and per-seat confidence calibration across completed calls (Panel~B).}
\label{tab:app-prod}
\small
\begin{adjustbox}{max width=\textwidth}
\begin{tabular}{@{}lrr@{\hspace{1.8em}}lrrrrr@{}}
\toprule
\multicolumn{3}{@{}l}{\textbf{Panel A: Stage Attribution ($1{,}261$ Clues)$^*$}} & \multicolumn{6}{l@{}}{\textbf{Panel B: Seat Calibration \& Attribution across Passes$^\dagger$}} \\
\cmidrule(r){1-3} \cmidrule(l){4-9}
\textbf{Pipeline Stage} & \textbf{Clues} & \textbf{Share} & \textbf{Confidence Bin} & \multicolumn{2}{c}{\textbf{Gemini 3.1 Pro}} & \multicolumn{2}{c}{\textbf{Claude Fable 5}} & \textbf{GPT-5} \\
\cmidrule(lr){5-6} \cmidrule(lr){7-8}
 & & & & Calls & Acc. & Calls & Acc. & (Retries) \\
\midrule
Pass 0 (Cold)                &   944 & $74.9\%$ & $[90, 100]$ & $1{,}209$ & $99.4\%$ & $1{,}286$ & $99.5\%$ & --- \\
Passes 1--3 (Retries)        &   307 & $24.3\%$ & $[80, 89]$  &        2  & $100.0\%$ &      118  & $84.7\%$ & --- \\
Pass 4 (Historical List)     &     6 &  $0.5\%$ & $[70, 79]$  &        0  & ---      &       31  & $48.4\%$ & --- \\
Terminal Sweep (\textsc{swp})&     4 &  $0.3\%$ & $\le 69$    &        0  & ---      &        7  & $14.3\%$ & --- \\
\textsc{corr} / \textsc{pers} / \textsc{atf} & 0 & $0.0\%$ & \textit{Pass 0 Silent Rate} & \multicolumn{2}{c}{$25.9\%$} & \multicolumn{2}{c}{$11.6\%$} & --- \\
\cmidrule(r){1-3} \cmidrule(l){4-9}
\textbf{Total Committed}     & $1{,}260$ & $99.92\%$ & \textit{First-to-Correct} & \multicolumn{2}{c}{$234$ ($18.6\%$)} & \multicolumn{2}{c}{$1{,}016$ ($80.9\%$)} & $6$ ($0.5\%$) \\
\textbf{Total Correct}       & $\mathbf{1{,}259}$ & $\mathbf{99.84\%}$ & \textit{Unique Correct} & \multicolumn{2}{c}{$71$ clues} & \multicolumn{2}{c}{$28$ clues} & $1$ clue \\
\bottomrule
\multicolumn{9}{@{}l}{\footnotesize $^*$Of the $1{,}261$ clues across $43$ E2 puzzles, Pass~0 solved $944/944$; Passes~1--3 solved $306$ and left $1$ unfilled ($307$ total clues whose last} \\
\multicolumn{9}{@{}l}{\footnotesize \phantom{$^*$}active pass was Passes~1--3); Pass~4 solved $6/6$; and \textsc{swp} filled $4$ ($3$ correct, $1$ wrong), giving $1{,}260$ committed ($1{,}259$ correct) and $1$ blank.} \\
\multicolumn{9}{@{}l}{\footnotesize $^\dagger$Calibration bins pool all status-\texttt{ok} calls across passes ($1{,}211$ Gemini, $1{,}442$ Claude). \textit{First-to-Correct} sums to the $1{,}256$ correct clues} \\
\multicolumn{9}{@{}l}{\footnotesize \phantom{$^\dagger$}committed by model passes ($234 + 1{,}016 + 6 = 1{,}256$, excluding the $3$ correct dictionary \textsc{swp} fills). Phase-timing medians across the $43$} \\
\multicolumn{9}{@{}l}{\footnotesize \phantom{$^\dagger$}E2 runs (\cref{sec:timeouts-replicas}): total solve $336.4\text{s}$ ($5\text{m}\,36\text{s}$, $\text{p90}=555.0\text{s}$); Pass~0 $162.8\text{s}$; Pass~1 $155.9\text{s}$; barrier wait after last \texttt{ok} call $86.5\text{s}$ ($\text{p90}=255.6\text{s}$).}
\end{tabular}
\end{adjustbox}
\end{table}

\begin{table}[!htbp]
\centering
\caption{Performance by \textsc{Snitch} difficulty band on the 16-puzzle stratified test set ($N = 463$ clues; \cref{sec:corpus-eval}), comparing the 2025 V1 baseline, the June 2026 Epoch~E0 prototype, and the Championship V2 configuration ($D = 1$ on bands $70$--$130$; $D = 3$ on bands $130+$).}
\label{tab:app-testset}
\small
\begin{adjustbox}{max width=\textwidth}
\begin{tabular}{@{}lrrrrrrrr@{}}
\toprule
\textbf{SNITCH Difficulty Band} & \textbf{Puzzles} & \textbf{Clues} & \textbf{Replicas} & \textbf{2025 V1} & \textbf{June '26} & \textbf{Champ.\ First} & \textbf{Measured Replicas} & \textbf{Med Time} \\
\midrule
Easy ($70$--$90$)        &  1 &  30 & $D=1$ & $30/30$ &  $29/30$ ($96.7\%$) &  $30/30$ ($100.0\%$) & $30$ (single run) & $04\text{m}\,08\text{s}$ \\
Moderate ($90$--$110$)   &  3 &  87 & $D=1$ & $75/87$ &  $83/87$ ($95.4\%$) &  $87/87$ ($100.0\%$) & $87$ (single run) & $05\text{m}\,02\text{s}$ \\
Hard ($110$--$130$)      &  4 & 116 & $D=1$ & $103/116$ & $106/116$ ($91.4\%$) & $116/116$ ($100.0\%$) & $116$ (single run) & $04\text{m}\,07\text{s}$ \\
Very Hard ($130$--$160$) &  5 & 142 & $D=3$ & $108/142$ & $121/142$ ($85.2\%$) & $142/142$ ($100.0\%$) & $142 / 142 / 142$ & $04\text{m}\,09\text{s}$ \\
Extreme ($160+$)         &  3 &  88 & $D=3$ & $64/88$ &  $69/88$ ($78.4\%$) &  $86/88$ ($97.7\%$)  & $87 / 85 / 86^*$ & $05\text{m}\,18\text{s}$ \\
\midrule
\textbf{Total ($16$ Puzzles)} & \textbf{16} & \textbf{463} & \textbf{32 runs} & $\mathbf{380}$ ($\mathbf{82.1\%}$) & $\mathbf{408}$ ($\mathbf{88.1\%}$) & $\mathbf{461}$ ($\mathbf{99.57\%}$) & $\mathbf{917 / 923\text{ }\text{clues}^\dagger}$ & $\mathbf{04\text{m}\,11\text{s}}$ \\
\bottomrule
\multicolumn{9}{@{}l}{\footnotesize $^*$In the Extreme ($160+$) band, the three individual replicas scored $87/88$ (Replica~0, \texttt{rep1}), $85/88$ (Replica~1, \texttt{rep2}), and $86/88$ (Replica~2, \texttt{rep3}); first-finisher} \\
\multicolumn{9}{@{}l}{\footnotesize \phantom{$^*$}selection picked Replica~1 (\texttt{rep2}, $29/30$) on \texttt{29399}, Replica~2 (\texttt{rep3}, $27/28$) on \texttt{29435}, and Replica~1 (\texttt{rep2}, $30/30$) on \texttt{29447}, yielding $86/88$.} \\
\multicolumn{9}{@{}l}{\footnotesize $^\dagger$Across all $32$ executed runs ($8\times 1 + 8\times 3 = 923$ clue-evaluations), total accuracy is $917/923$ ($99.35\%$, $6$ wrong, $0$ blank).}
\end{tabular}
\end{adjustbox}
\end{table}

\clearpage
\begin{table}[!htbp]
\centering
\caption{Average-case stage ablations (\cref{sec:mean-eval} and \Cref{fig:stage-ablation}): 80-run single-replica ($D=1$) leave-one-out ablation on the 16-puzzle test set ($N = 463$; Panel~A) and $2\times 2$ factorial local replay across $41$ cell-exact Epoch~E2 daily runs ($N = 1{,}204$; Panel~B).}
\label{tab:app-ablations}
\small
\begin{adjustbox}{max width=\textwidth}
\begin{tabular}{@{}lrrrrrl@{}}
\toprule
\textbf{Ablation Configuration} & \textbf{Correct} & \textbf{Wrong} & \textbf{Blank} & \textbf{$\Delta$} & \textbf{Med $t^*$} & \textbf{Structural Verdict (Lost / Gained; Feed Adms)} \\
\midrule
\multicolumn{7}{@{}l}{\textit{Panel A: Leave-One-Out on 16-Puzzle Stratified Test Set ($N = 463$ Clues, Live API)}} \\
Full Championship Baseline ($D=1/3$ ff) & $461 / 463$ &  2 & 0 & --- & $04\text{m}\,11\text{s}$ & $99.57\%$ reference ($05\text{m}\,05\text{s}$ single-rep med; $8$ ff / $17$ all-rep feed) \\
Drop Crossing Retries (Passes 1--3 + Tail) & $448 / 463$ & 10 & 5 & $\mathbf{-13}$ & $09\text{m}\,23\text{s}$ & \textbf{Outside per-puzzle noise} ($13$ lost / $0$ gained; $31$ feed) \\
Drop Historical-Answer Pass (Pass 4) & $460 / 463$ &  3 & 0 & $-1$ & $05\text{m}\,28\text{s}$ & Inside replica noise ($1$ lost / $0$ gained; $6$ feed) \\
Drop Whole-Grid Feed (\texttt{gpt-6-astra}) & $460 / 463$ &  3 & 0 & $-1$ & $05\text{m}\,10\text{s}$ & Inside replica noise ($2$ lost / $1$ gained; $0$ feed) \\
Drop Secondary Seats (Gemini Only A/A1)$^\dagger$ & $\mathbf{461 / 463}$ &  2 & 0 & $\mathbf{0}$ & $06\text{m}\,05\text{s}$ & \textbf{Exact accuracy parity} ($2$ lost / $2$ gained; $7$ feed) \\
Drop Corroboration Admission (Greedy) & $460 / 463$ &  3 & 0 & $-1$ & $06\text{m}\,55\text{s}$ & Inside replica noise ($3$ lost / $2$ gained; $0$ feed) \\
\midrule
\multicolumn{7}{@{}l}{\textit{Panel B: $2 \times 2$ Factorial Replay on 41 E2 Production Days ($N = 1{,}204$ Clues, Local Replay)}} \\
Corroboration ON, Terminal Chain ON   & $1{,}204 / 1{,}204$ & 0 &  0 & ---   & $05\text{m}\,36\text{s}$ & $100.0\%$ cell-exact production baseline \\
Corroboration OFF, Terminal Chain ON  & $1{,}199 / 1{,}204$ & 5 &  0 & $-5$  & --- & Terminal chain rescues $90$ of $95$, admits $5$ errors ($3$ via \textsc{swp}) \\
Corroboration ON, Terminal Chain OFF  & $1{,}200 / 1{,}204$ & 0 &  4 & $-4$  & --- & Admission holds $0$ errors; $4$ terminal clues left blank \\
Corroboration OFF, Terminal Chain OFF & $1{,}105 / 1{,}204$ & 1 & 98 & $\mathbf{-99}$ & --- & \textbf{Mutual-backup collapse ($-8.2\%$; interaction $-90$ clues)} \\
\bottomrule
\multicolumn{7}{@{}l}{\footnotesize Pre-registered replica-noise reference in Panel~A: $\max(1, \text{disagreement}_p)$ per puzzle ($2$ on \texttt{29399} and \texttt{29435}, $1$ on the other $14$ puzzles; $18$ summed).} \\
\multicolumn{7}{@{}l}{\footnotesize $^*$Diagnostic only: $51$ of $80$ test-set ablation units ran in daytime after an ISP outage with $3$ arms concurrent, inflating wall-clocks vs.\ overnight baseline.} \\
\multicolumn{7}{@{}l}{\footnotesize $^\dagger$Retains the background \texttt{gpt-6-astra} whole-grid feed ($7$ feed admissions across the $16$ runs); removes Claude and GPT-5 from Passes~0--3.}
\end{tabular}
\end{adjustbox}
\end{table}

\begin{table}[!htbp]
\centering
\caption{Tail-weighted breakdown of the 132-day daily production record ($3{,}876$ clues across Epochs~E0--E2; \cref{sec:tail-daily} and \Cref{fig:mean-vs-tail}), stratified by daily Pass~0 Gemini anchor silent rate.}
\label{tab:app-tail}
\small
\begin{adjustbox}{max width=\textwidth}
\begin{tabular}{@{}lrrrrrrr@{}}
\toprule
\textbf{Production Stratum} & \textbf{Days} & \textbf{Clues} & \textbf{Anchor Pass 0 Silent} & \textbf{Correct / Wrong / Blank} & \textbf{Perfect Grids} & \textbf{Non-Anchor Commits} & \textbf{Peak Rescue} \\
\midrule
Deciles 1--9 (Normal) & 119 & $3{,}498$ & $0.0\%$--$40.0\%$ (med $20.6\%$) & $3{,}471 / 4 / 23$ ($\mathbf{99.23\%}$) & $100 / 119$ ($84.0\%$) & $134 / 3{,}475$ ($3.9\%$) & --- \\
Decile 10 (Worst Tail) &  13 &     378   & $42.9\%$--$100.0\%$ (med $100.0\%$) & $373 / 1 / 4$ ($\mathbf{98.68\%}$) & $8 / 13$ ($61.5\%$) & $120 / 374$ ($\mathbf{32.1\%}$) & Up to $\mathbf{90.0\%}$ \\
\midrule
\quad \textit{9 Pass-0 DNS Runs (E0)} & 9 & 262 & $100.0\%$ (\texttt{Errno 8} burst) & $258 / 1 / 3$ ($98.47\%$) & $5 / 9$ ($55.6\%$) & $104 / 259$ ($40.2\%$) & $27/30$ ($90.0\%$) \\
\quad \textit{1 Partial Error Run (E0)} & 1 & 30 & $70.0\%$ ($13$ err $+$ $8$ T/O) & $29 / 0 / 1$ ($96.67\%$) & $0 / 1$ ($0.0\%$) & $0 / 29$ ($0.0\%^*$) & Anchor at Pass 1 \\
\quad \textit{3 Brownout Days (E1/E2)} & 3 & 86 & $42.9\%$--$46.7\%$ (timeouts) & $86 / 0 / 0$ ($\mathbf{100.00\%}$) & $3 / 3$ ($\mathbf{100.0\%}$) & $16 / 86$ ($18.6\%$) & $100\%$ accuracy \\
\quad \textit{Peak 2-Day Outage (Jun 23, 27)} & 2 & 60 & $100.0\%$ (\texttt{Errno 8} burst) & $59 / 0 / 1$ ($98.33\%$) & $1 / 2$ ($50.0\%$) & $53 / 59$ ($\mathbf{89.8\%}$) & $27/30$ ($90.0\%$) \\
\midrule
\textbf{All 132 Daily Runs} & \textbf{132} & $\mathbf{3{,}876}$ & \textbf{Med $\mathbf{21.4\%}$, p90 $\mathbf{40.0\%}$} & $\mathbf{3{,}844 / 5 / 27}$ ($\mathbf{99.17\%}$) & $\mathbf{108 / 132}$ ($\mathbf{81.8\%}$) & $\mathbf{254 / 3{,}849}$ ($\mathbf{6.6\%}$) & --- \\
\bottomrule
\multicolumn{8}{@{}l}{\footnotesize $^*$On 2026-06-20 ($70\%$ Pass~0 silent), Gemini recovered at Pass~1 and proposed all $29$ committed answers by their final non-feed pass.}
\end{tabular}
\end{adjustbox}
\end{table}

\begin{table}[!htbp]
\centering
\caption{Telemetry across all seven puzzles of the 55th \textit{Times} Crossword Championship (October 10, 2026; Panel~A) and per-pass commit ledger of the winning Championship Final replica (\texttt{run\_2}; Panel~B, \cref{sec:tail-final} and \Cref{fig:brownout}).}
\label{tab:app-final}
\small
\begin{adjustbox}{max width=\textwidth}
\begin{tabular}{@{}lrrrrr@{\hspace{1.4em}}lrr@{}}
\toprule
\multicolumn{6}{@{}l}{\textbf{Panel A: All 7 Championship Puzzles (Live Event)}} & \multicolumn{3}{l@{}}{\textbf{Panel B: Final Call \& Commit Ledger (\texttt{run\_2}$^\dagger$)}} \\
\cmidrule(r){1-6} \cmidrule(l){7-9}
\textbf{Round / Puzzle} & \textbf{Score} & \textbf{Solve $t$} & \textbf{Gem tok/s} & \textbf{Gem P0 T/O} & \textbf{Non-Anchor} & \textbf{Commit Path in Final} & \textbf{Clues} & \textbf{Anchor?} \\
\midrule
Qualifier 1 (29132) & $30/30$ & $04\text{m}\,20\text{s}$ & $116.8$ & $3/30$ & $2/30$ & Pass 0: Gemini + Claude agree & $6$ & Yes \\
Qualifier 2 (29137) & $30/30$ & $04\text{m}\,16\text{s}$ & $118.4$ & $1/30$ & $1/30$ & Pass 0: Gemini lone (9D, $\ge 50$) & $1$ & Yes \\
Qualifier 3 (29116) & $30/30$ & $03\text{m}\,32\text{s}$ & $119.1$ & $2/30$ & $1/30$ & Pass 1: All 3 seats ($5$ P1 + 11A) & $6$ & Yes \\
\cmidrule(l){7-9}
Semi-Final 1 (29290) & $30/30$ & $06\text{m}\,45\text{s}$ & $115.2$ & $6/30$ & $4/30$ & Pass 1: Cla + GPT-5 ($8$ P1 + $2$ cross) & 10 & \textbf{No} \\
Semi-Final 2 (29315) & $30/30$ & $07\text{m}\,49\text{s}$ & $114.6$ & $8/30$ & $5/30^*$ & Pass 1: Claude solo $+$ crossings & 3 & \textbf{No} \\
Semi-Final 3 (29125) & $30/30$ & $05\text{m}\,08\text{s}$ & $117.0$ & $4/30$ & $2/30$ & Boundary Feed (\texttt{gpt-6-astra}) & 4 & \textbf{No} \\
\cmidrule(r){1-6} \cmidrule(l){7-9}
\textbf{Final (944766)} & $\mathbf{30/30}$ & $\mathbf{06\text{m}\,52\text{s}}$ & $\mathbf{33.0}$ & $\mathbf{23/30}$ & $\mathbf{17/30}$ & \textbf{Total Committed Without Anchor} & $\mathbf{17 / 30}$ & $\mathbf{56.7\%}$ \\
\bottomrule
\multicolumn{9}{@{}l}{\footnotesize $^*$In Semi-Final 2, the \texttt{gpt-6-astra} feed completed on disk at $05\text{m}\,32\text{s}$ ($332.0\text{s}$), $13.7\text{s}$ after Pass~2 had started ($318.3\text{s}$); because the boundary} \\
\multicolumn{9}{@{}l}{\footnotesize \phantom{$^*$}feed is polled only at pass barriers, 10A (\ans{CATHERINE HOWARD}) waited $136.5\text{s}$ for Pass~2 to exit at $07\text{m}\,49\text{s}$ ($468.5\text{s}$).} \\
\multicolumn{9}{@{}l}{\footnotesize $^\dagger$Call ledger in \texttt{run\_2}: Pass~0 ($30$ clues) returned $7/30$ Gemini (\texttt{23 T/O}) and $26/30$ Claude (\texttt{3 T/O, 1 empty}), keeping $7$ clues (20A, 21A, 25A, 28A,} \\
\multicolumn{9}{@{}l}{\footnotesize \phantom{$^\dagger$}5D, 6D via Gem+Cla; 9D via Gem solo); Pass~1 ($23$ open clues) returned $6/23$ Gemini (\texttt{16 T/O, 1 parse fail}), $16/23$ Claude (\texttt{5 T/O, 2 parse fail} on 4A/11A),} \\
\multicolumn{9}{@{}l}{\footnotesize \phantom{$^\dagger$}and $19/23$ GPT-5 (\texttt{4 empty}), keeping $19$ more ($26/30$): $6$ proposed by all $3$ seats ($5$ in Pass~1 + 11A Gem+GPT-5 w/ Pass-0 Cla), $10$ by Cla+GPT-5} \\
\multicolumn{9}{@{}l}{\footnotesize \phantom{$^\dagger$}($8$ in Pass~1 + 4A, 26A GPT-5 lifted by crossings w/ Pass-0 Cla), and $3$ by Cla solo lifted by crossings (7D, 12A, 16A, where GPT-5 returned empty).}
\end{tabular}
\end{adjustbox}
\end{table}

\clearpage
\section{Production Prompt Specification and Glossary of Cited Clues}
\label{app:prompts-clues}

\subsection{Full Production Pass 0 Prompt (\texttt{P0}) and Ablation Diffs}
\label{app:prompt-p0}

Below is the exact user prompt template used in Pass~0 (\texttt{P0}) across our 43-day Epoch~E2 daily production runs, the 16-puzzle test set, the 2026 Championship, and the prompt-ablation study of \cref{sec:prompt-ablation}.\footnote{In the verbatim historical prompt below, the parenthetical example \texttt{"(2,3-4)" is e.g.\ "OF FIRST-RATE"} contains an author typo in the example's middle digit (\texttt{FIRST} has $5$ letters, so \texttt{"OF FIRST-RATE"} has enumeration \texttt{(2,5-4)}); we preserve the prompt string verbatim as executed across all runs.}
Bracketed comment lines (\texttt{\#\#\# [Variant note] \#\#\#}, not sent to the model) are interspersed below to mark the exact blocks modified or omitted by ablation arms \texttt{P1}--\texttt{P7}.
In Passes~1--3 (\cref{sec:pipeline}), the prompt is augmented with the current intersecting letter pattern (e.g., \texttt{PATTERN: P?U?E?I?E}) and, when the pattern matches between $1$ and $100$ composite curated lexicon entries (or historical \textit{Times} answers in Pass~4), the non-binding candidate list.

\begin{footnotesize}
\begin{verbatim}
You are an expert solver of Times (London) cryptic crosswords.

Solve this cryptic clue. Cryptic clues consist of a definition and
wordplay that both lead to the same answer.

CLUE: {clue_text}
DIRECTION: {direction}
ENUMERATION: ({enumeration})

### [START DOMAIN GUIDANCE: omitted in P4, P5, P7] ###
Work through the clue carefully. Identify the definition (usually at
the start or end of the clue, never in the middle). Parse the wordplay.
Verify your answer satisfies both.

The DIRECTION field (across or down) can be relevant to parsing.
Setters sometimes use directional words ("up", "down", "east", "west",
"northbound", etc.) that read naturally in the clue's surface meaning
but encode wordplay operations matching the clue's grid direction.
Consider direction when interpreting such words.
### [END DOMAIN GUIDANCE] ###

Provide your response as JSON in this exact order:

{
  ### [START TAXONOMY DECOMPOSITION FIELDS: omitted in P3, P5, P7] ###
  "definition": "the word or phrase you've identified as the definition",
  "wordplay_type": "anagram | charade | container | insertion | deletion |
                    hidden | reversal | homophone | substitution |
                    spoonerism | acrostic | double_definition |
                    cryptic_definition | other",
  "is_and_lit": <true|false>,
  ### [END TAXONOMY DECOMPOSITION FIELDS] ###

  ### [EXPLANATION FIELD: omitted in P5, P7; moved after confidence in P1] ###
  "wordplay_explanation": "explain how you arrived at the answer",

  ### [DRAFT ANSWER FIELD: omitted in P1, P2, P5, P6, P7] ###
  "answer_unverified": "your answer in uppercase, no spaces or punctuation",

  ### [CONFIDENCE OBJECT: omitted in P6, P7; level-only in P5; moved 2nd in P1] ###
  "confidence": {
    "level": <integer 0-100>,
    "reason": "explain what makes you more or less confident -- e.g., does
               the wordplay parse cleanly, does the definition fit, are
               there ambiguities?"
  },

  ### [FINAL ANSWER FIELD: moved 1st in P1; single answer in P1, P2, P5, P6, P7] ###
  "answer": "your final answer, identical to answer_unverified unless you
             noticed a problem during confidence assessment"
}

### [SELF-REVISION INSTRUCTION: omitted in P1, P2, P5, P6, P7] ###
The confidence reflects your assessment of the answer BEFORE you commit
to it as final. If your confidence assessment surfaces a problem,
revise the answer in the final field.

### [ENUMERATION CHECK BLOCK: examples omitted in P4; entire block omitted in P5, P7] ###
Length must match the enumeration `({enumeration})`. Examples: `(7)` is
one 7-letter word. `(3,4)` is two words of 3 and 4 letters. `(3-4)` is
hyphenated, 3 letters then 4 letters. `(1'5)` is an apostrophe-form
like "O'HENRY" -- 1 letter, apostrophe, 5 letters. Patterns can
combine: `(2,3-4)` is e.g. "OF FIRST-RATE." Confirm your answer
matches the enumeration's word count, hyphenation, apostrophe
placement, and total letter count before responding.

Do not use internet search. Respond only with valid JSON.
\end{verbatim}
\end{footnotesize}

\paragraph{Summary of schema manipulations across arms \texttt{P1}--\texttt{P7} (\cref{sec:prompt-ablation,sec:conf-ablation}).}
\begin{itemize}[leftmargin=1.8em,itemsep=1.5pt]
    \item \texttt{P1} (\textit{Answer-first, single-answer schema}).
    Moves \texttt{"answer"} and \texttt{"confidence"} to the top of the JSON object before \texttt{"definition"}, \texttt{"wordplay\_\allowbreak type"}, \texttt{"is\_\allowbreak and\_\allowbreak lit"}, and \texttt{"wordplay\_\allowbreak explanation"}, while omitting \texttt{"answer\_\allowbreak unverified"} and the self-revision paragraph.
    \item \texttt{P2} (\textit{No two-stage answer}).
    Replaces \texttt{"answer\_\allowbreak unverified"} + \texttt{"confidence"} + \texttt{"answer"} with a single \texttt{"answer"} followed by \texttt{"confidence"} at the end of the schema, removing the self-revision paragraph.
    \item \texttt{P3} (\textit{No taxonomy decomposition fields}).
    Deletes the three structured classification keys (\texttt{"definition"}, \texttt{"wordplay\_\allowbreak type"}, and \texttt{"is\_\allowbreak and\_\allowbreak lit"}), starting the JSON schema directly with \texttt{"wordplay\_\allowbreak explanation"}, \texttt{"answer\_\allowbreak unverified"}, \texttt{"confidence"}, and \texttt{"answer"}.
    \item \texttt{P4} (\textit{Minimalist instructions}).
    Retains the full \texttt{P0} JSON schema but removes the cryptic tutorial paragraph, the directional-indicator paragraph, and the worked enumeration examples.
    \item \texttt{P5} (\textit{Answer + numeric confidence only}).
    Strips all instructional scaffolding and all explanation/decomposition fields, requesting only \texttt{\{"answer": "...", "confidence": \{"level": <0-100>\}\}}.
    \item \texttt{P6} (\textit{Explanation + single answer, no confidence}).
    Modifies \texttt{P2} by removing the \texttt{"confidence"} object (retaining \texttt{"definition"}, \texttt{"wordplay\_\allowbreak type"}, \texttt{"is\_\allowbreak and\_\allowbreak lit"}, \texttt{"wordplay\_\allowbreak explanation"}, and a single \texttt{"answer"}).
    \item \texttt{P7} (\textit{Bare answer only}).
    Modifies \texttt{P5} by removing the \texttt{"confidence"} object, requesting only \texttt{\{"answer": "..."\}}.
\end{itemize}

\subsection{Complete Glossary and Wordplay Derivations of Cited Clues}
\label{app:clue-glossary}

To ensure that every cryptic clue mentioned in the main text is transparent and verifiable without repeating long derivations inline, \Cref{tab:app-clues-1,tab:app-clues-2,tab:app-clues-3} index all $29$ cited clues (\texttt{C1}--\texttt{C29}) with their exact clue text, enumeration, direction, solution, definition span, and verified step-by-step wordplay derivation.

\begin{table}[!htbp]
\centering
\caption{Solutions and verified wordplay derivations for clues \texttt{C1}--\texttt{C12} cited in \cref{sec:intro,sec:related,sec:arch,sec:folklore}.}
\label{tab:app-clues-1}
\footnotesize
\setlength{\tabcolsep}{3.8pt}
\setlength{\aboverulesep}{1.2pt}
\setlength{\belowrulesep}{1.6pt}
\begin{adjustbox}{max width=\textwidth}
\begin{tabular}{@{}l>{\raggedright\arraybackslash}p{2.30cm}>{\raggedright\arraybackslash}p{3.95cm}>{\raggedright\arraybackslash}p{2.50cm}>{\raggedright\arraybackslash}p{6.65cm}@{}}
\toprule
\textbf{ID} & \textbf{Source \& Ref} & \textbf{Clue Text (Enumeration)} & \textbf{Solution (Def.)} & \textbf{Wordplay Derivation} \\
\midrule
\textbf{C1} & Introductory Down Example (\cref{sec:primer-clues}) & \clue{Note invested in aluminium raised capital (4)} (Down) & \ans{LIMA}\newline(\textit{capital}) & \textbf{Reversal of Container:} \ans{MI} (\textit{Note}, sol-fa) inserted into (\textit{invested in}) \ans{AL} (\textit{aluminium}) $= \ans{A[MI]L}$, reversed upward (\textit{raised} in a Down clue) $\to \ans{LIMA}$. \\
\midrule
\textbf{C2} & Williams \& Woodhead~\cite{williams1979computer} (\cref{sec:related-work}) & \clue{The object is to cut a new robe, by the way (6)} (Across) & \ans{OBITER}\newline(\textit{by the way}) & \textbf{Container + Anagram:} \ans{IT} (\textit{The object}) inserted into (\textit{to cut}) an anagram (\textit{new}) of \texttt{ROBE} ($\ans{OBER}$) $\to \ans{OB[IT]ER}$ (Latin: ``in passing''). \\
\midrule
\textbf{C3} & 2026 Champ.\ SF1, Clue 28A (\cref{sec:primer-clues,sec:error-analysis}) & \clue{Setting last of puzzles put out by Times in competition (7)} (Across) & \ans{CONTEXT}\newline(\textit{Setting}) & \textbf{Nested Substitution:} Last letter of \textit{puzzle\underline{s}} (\ans{S}) replaced (\textit{put out by}) with \ans{X} (\textit{Times}, multiplication sign $\times$) inside (\textit{in}) \ans{CONTEST} (\textit{competition}) $\to \ans{CONTE[S}\to\ans{X]T}$. \\
\midrule
\textbf{C4} & \textit{Times}~\#29399, Clue 1A (\cref{sec:corrob-admission}) & \clue{Lay down fancy red carpet for collecting Oscar (4-6)} (Across) & \ans{TAPE-RECORD}\newline(\textit{Lay down}) & \textbf{Anagram + Container:} Anagram (\textit{fancy}) of \texttt{RED CARPET} containing (\textit{for collecting}) \ans{O} (\textit{Oscar} in NATO alphabet) $\to \ans{TAPE-REC[O]RD}$. \\
\midrule
\textbf{C5} & \textit{Times}~\#29399, Clue 2D (\cref{sec:corrob-admission,sec:error-analysis}) & \clue{Make poor value bandages more expensive in A\&E (9)} (Down) & \ans{PAUPERIZE}\newline(\textit{Make poor}) & \textbf{Nested Container:} \ans{PRIZE} (\textit{value}) wraps around (\textit{bandages}) \ans{UP} (\textit{more expensive}) placed inside \ans{A} and \ans{E} (\textit{in A\&E} $= \ans{A[UP]E}$) $\to \ans{P[A(UP)E]RIZE}$. \\
\midrule
\textbf{C6} & \textit{Times}~\#29399, Clue 10A (\cref{sec:corrob-admission}) & \clue{Thoughtlessly move mother to an inferior place (7)} (Across) & \ans{MAUNDER}\newline(\textit{Thoughtlessly move}) & \textbf{Charade:} \ans{MA} (\textit{mother}) $+$ \ans{UNDER} (\textit{to an inferior place}) $\to \ans{MAUNDER}$ (supplies crossing \ans{U} at cell $(3,3)$ of \texttt{2D} \ans{PAUPERIZE}). \\
\midrule
\textbf{C7} & \textit{Times}~\#29570, Clue 21A (\cref{sec:greedy-pathology}) & \clue{Ropes tied tight and put down again (9)} (Across) & \ans{REDEPOSIT}\newline(\textit{put down again}) & \textbf{Anagram:} Anagram (\textit{tight}) of \texttt{ROPES TIED} $\to \ans{REDEPOSIT}$ (models cold-guessed the decoy anagram \ans{REPOSITED}). \\
\midrule
\textbf{C8} & \textit{Times}~\#29611, Clue 15D (\cref{sec:timeouts-replicas}) & \clue{Will individual suit and uniform spoil climbing? (8)} (Down) & \ans{EXECUTOR}\newline(\textit{Will individual}) & \textbf{Charade + Reversal:} \ans{EXEC} (\textit{suit}, executive) $+$ \ans{U} (\textit{uniform}, NATO) $+$ \ans{ROT} (\textit{spoil}) reversed (\textit{climbing} in Down clue $\to \ans{TOR}$) $= \ans{EXECUTOR}$ (person appointed in a will). \\
\midrule
\textbf{C9} & Smoke Gate Pair 1 (\cref{sec:verifier-falsification}) & \clue{Soldiers occupy a position around river (4)} (Across) & \ans{EBRO}\newline(\textit{river}) & \textbf{Charade + Reversal:} \ans{OR} (\textit{Soldiers}, Other Ranks) $+$ \ans{BE} (\textit{occupy a position}) $= \ans{ORBE}$, reversed (\textit{around}) $\to \ans{EBRO}$ (paired against decoy river \ans{ERNE}). \\
\midrule
\textbf{C10} & Smoke Gate Pair 3 (\cref{sec:verifier-falsification}) & \clue{Finally snap in rage, so agitated (2,5)} (Across) & \ans{GO SPARE}\newline(\textit{\&lit}: whole clue) & \textbf{\&lit (All-in-One) Anagram:} Final letter (\textit{Finally}) of \textit{sna\underline{p}} (\ans{P}) $+$ \texttt{RAGE} $+$ \texttt{SO} anagrammed (\textit{agitated}) $\to \ans{GO SPARE}$ (lose one's temper). \\
\midrule
\textbf{C11} & Smoke Gate Pair 4 (\cref{sec:verifier-falsification}) & \clue{Ex-PM's fashionable royal post (15)} (Across) & \ans{CHAMBERLAINSHIP}\newline(\textit{royal post}) & \textbf{Charade:} \ans{CHAMBERLAIN} (\textit{Ex-PM}, Neville Chamberlain) $+$ \ans{S} (\textit{'s}) $+$ \ans{HIP} (\textit{fashionable}) $\to \ans{CHAMBERLAINSHIP}$ (evaluated on 14-letter wrong decoy \ans{LORD PALMERSTON} passed to the enumeration-agnostic verifier). \\
\midrule
\textbf{C12} & Smoke Gate Pair 5 (\cref{sec:verifier-falsification}) & \clue{Risked neck, regularly fleeing English country houses (7)} (Across) & \ans{CHANCED}\newline(\textit{Risked}) & \textbf{Alternating Selection + Container:} Odd letters of \textit{\underline{n}e\underline{c}k} (\ans{NC}, with even letters \textit{regularly fleeing}) $+$ \ans{E} (\textit{English}) housed inside \ans{CHAD} (\textit{country}) $\to \ans{CHA[NCE]D}$ (paired against decoy \ans{ESTATES}). \\
\bottomrule
\end{tabular}
\end{adjustbox}
\end{table}

\begin{table}[!htbp]
\centering
\caption{Solutions and verified wordplay derivations for clues \texttt{C13}--\texttt{C22} cited in \cref{sec:eval} from the 2026 Championship Semi-Final~2, Championship Final, and daily production.}
\label{tab:app-clues-2}
\footnotesize
\setlength{\tabcolsep}{3.8pt}
\setlength{\aboverulesep}{1.2pt}
\setlength{\belowrulesep}{1.6pt}
\begin{adjustbox}{max width=\textwidth}
\begin{tabular}{@{}l>{\raggedright\arraybackslash}p{2.30cm}>{\raggedright\arraybackslash}p{3.95cm}>{\raggedright\arraybackslash}p{2.50cm}>{\raggedright\arraybackslash}p{6.65cm}@{}}
\toprule
\textbf{ID} & \textbf{Source \& Ref} & \textbf{Clue Text (Enumeration)} & \textbf{Solution (Def.)} & \textbf{Wordplay Derivation} \\
\midrule
\textbf{C13} & 2026 Champ.\ SF2, Clue 10A (\cref{sec:tail-final}) & \clue{Queen say divorced by the king; he and I now entangled a queen (9,6)} (Across) & \ans{CATHERINE HOWARD}\newline(\textit{queen}) & \textbf{Container + Charade + Anagram:} \ans{CARD} (\textit{Queen, say}, playing card by example) split (\textit{divorced by}) around \ans{THE} $+$ \ans{R} (\textit{king}, Rex) $+$ anagram (\textit{entangled}) of \texttt{HE} [and] \texttt{I NOW} ($6$ letters $\to \ans{INEHOW}$, with \textit{and} linking the fodder tokens) $+$ \ans{A} (\textit{a}) $\to \ans{CA[THE R INEHOW A]RD}$ ($2+3+1+6+1+2 = 15$ letters). \\
\midrule
\textbf{C14} & 2026 Champ.\ Final, Clue 1D (\cref{sec:tail-final}) & \clue{An exam shortened with new master on friendly terms (5,5)} (Down) & \ans{MATES RATES}\newline(\textit{friendly terms}) & \textbf{Curtailment + Anagram:} \ans{A} (\textit{An}) $+$ \ans{TES[t]} (\textit{exam shortened}) $+$ \texttt{MASTER} (\textit{new} = anagrammed across the combined $10$-letter fodder $\ans{ATESMASTER}$) $\to \ans{MATES RATES}$. \\
\midrule
\textbf{C15} & 2026 Champ.\ Final, Clue 2D (\cref{sec:tail-final,sec:future-work}) & \clue{Wolf trap missing its lowest crosspiece? (5)} (Down) & \ans{SNARF}\newline(\textit{Wolf}, verb) & \textbf{Visual/Typographic Deletion:} \ans{SNARE} (\textit{trap}) written vertically Down has capital \ans{E} at the bottom; removing the lowest horizontal bar (\textit{crosspiece}) of \ans{E} turns it into \ans{F} $\to \ans{SNARF}$ (to wolf down). \\
\midrule
\textbf{C16} & 2026 Champ.\ Final, Clue 9D (\cref{sec:tail-final}) & \clue{Storm on the way, close to Lake Garda but maybe heading north (4,4)} (Down) & \ans{ROAD RAGE}\newline(\textit{Storm on the way}) & \textbf{Charade + Full Reversal:} Final letter (\textit{close to}) of \textit{Lak\underline{e}} (\ans{E}) $+$ \ans{GARDA} $+$ \ans{OR} (\textit{``but'', maybe}, conjunction by example) $= \ans{EGARDAOR}$, reversed upward (\textit{heading north} in Down clue) $\to \ans{ROAD RAGE}$. \\
\midrule
\textbf{C17} & 2026 Champ.\ Final, Clue 14A (\cref{sec:tail-final}) & \clue{Apparently bright high tech warning light went out (5)} (Across) & \ans{AIRED}\newline(\textit{went out}) & \textbf{Charade:} \ans{AI} (\textit{Apparently bright high tech}, artificial intelligence) $+$ \ans{RED} (\textit{warning light}) $\to \ans{AIRED}$ (was broadcast). \\
\midrule
\textbf{C18} & 2026 Champ.\ Final, Clue 18A (\cref{sec:tail-final}) & \clue{Profitable process quickly losing time, seconds in one iteration (9)} (Across) & \ans{EXPEDIENT}\newline(\textit{Profitable}) & \textbf{Deletion + Letter Selection:} \ans{EXPEDITE} (\textit{process quickly}) minus \ans{T} (\textit{losing time} $\to \ans{EXPEDIE}$) $+$ second letters (\textit{seconds in}) of \textit{o\underline{n}e} and \textit{i\underline{t}eration} (\ans{NT}) $\to \ans{EXPEDIENT}$. \\
\midrule
\textbf{C19} & 2026 Champ.\ Final, Clue 21A (\cref{sec:tail-final}) & \clue{Number before ten sufficient disproof of ``All odd numbers are prime'', say (14)} (Across) & \ans{COUNTEREXAMPLE}\newline(\textit{disproof of \dots, say}) & \textbf{Charade:} \ans{COUNT} (\textit{Number}) $+$ \ans{ERE} (\textit{before}) $+$ \ans{X} (\textit{ten}) $+$ \ans{AMPLE} (\textit{sufficient}) $\to \ans{COUNTEREXAMPLE}$ (and $9$, the ``number before ten,'' is itself the counterexample). \\
\midrule
\textbf{C20} & 2026 Champ.\ Final, Clue 27A (\cref{sec:tail-final}) & \clue{Faint but persistent echo, winding along gorges? (8)} (Across) & \ans{SNEAKING}\newline(\textit{Faint but persistent}) & \textbf{Container:} \ans{E} (\textit{echo} in NATO alphabet) swallowed by (\textit{gorges}) \ans{SNAKING} (\textit{winding along}) $\to \ans{SN[E]AKING}$. \\
\midrule
\textbf{C21} & 2026 Champ.\ Final, Clue 28A (\cref{sec:tail-final}) & \clue{Film contributor heads for cutting room leaving director worried (6)} (Across) & \ans{EDITOR}\newline(\textit{Film contributor}) & \textbf{Subtractive Anagram:} Initial letters (\textit{heads for}) of \textit{\underline{c}utting \underline{r}oom} (\ans{CR}) deleted from (\textit{leaving}) \texttt{DIRECTOR} ($\ans{DIRETO}$) and anagrammed (\textit{worried}) $\to \ans{EDITOR}$. \\
\midrule
\textbf{C22} & \textit{Times}~\#29658, Clue 8D (\cref{sec:error-analysis}) & \clue{Drawers that go up lower body parts comprising wide design (9)} (Down) & \ans{TOW-PLANES}\newline(\textit{Drawers that go up}) & \textbf{Container + Charade:} \ans{TOES} (\textit{lower body parts}) containing (\textit{comprising}) \ans{W} (\textit{wide}) $+$ \ans{PLAN} (\textit{design}) $\to \ans{TO[W-PLAN]ES}$ (cryptic definition \textit{``Drawers that go up''} = aircraft that draw/tow gliders aloft; unhyphenated \texttt{(9)} in club JSON; \textsc{swp} filled \ans{TOWELINGS}). \\
\bottomrule
\end{tabular}
\end{adjustbox}
\end{table}

\begin{table}[!htbp]
\centering
\caption{Solutions and verified wordplay derivations for cited \textit{Mephisto}, test-set, and commemorative clues \texttt{C23}--\texttt{C29} (\cref{sec:eval,sec:future-work}).}
\label{tab:app-clues-3}
\footnotesize
\setlength{\tabcolsep}{3.8pt}
\setlength{\aboverulesep}{1.2pt}
\setlength{\belowrulesep}{1.6pt}
\begin{adjustbox}{max width=\textwidth}
\begin{tabular}{@{}l>{\raggedright\arraybackslash}p{2.30cm}>{\raggedright\arraybackslash}p{3.95cm}>{\raggedright\arraybackslash}p{2.50cm}>{\raggedright\arraybackslash}p{6.65cm}@{}}
\toprule
\textbf{ID} & \textbf{Source \& Ref} & \textbf{Clue Text (Enumeration)} & \textbf{Solution (Def.)} & \textbf{Wordplay Derivation} \\
\midrule
\textbf{C23} & \textit{Mephisto}~\#3440, Clue 27D (\cref{sec:mephisto-eval,sec:error-analysis}) & \clue{Always rounding one up, as rustlers did once over the border (5)} (Down) & \ans{RIEVE}\newline(\textit{as rustlers did \dots}) & \textbf{Reversal + Container:} \ans{EVER} (\textit{Always}) reversed upward (\textit{up} in Down clue $\to \ans{REVE}$) containing (\textit{rounding}) \ans{I} (\textit{one}) $\to \ans{R[I]EVE}$ (Scots variant of \ans{REIVE}; ambiguous with reversing after insertion). \\
\midrule
\textbf{C24} & \textit{Mephisto}~\#3445, Clue 30A (\cref{sec:mephisto-eval,sec:error-analysis}) & \clue{Urge hounds over area by previously clear space (6)} (Across) & \ans{HALLOA}\newline(\textit{Urge hounds}) & \textbf{Charade:} \ans{HALL} (\textit{clear space}) placed before (\textit{by previously}) \ans{O} (\textit{over}) $+$ \ans{A} (\textit{area}) $\to \ans{HALLOA}$ (models guessed \ans{HOLLOA} via dialect \textit{holl}). \\
\midrule
\textbf{C25} & \textit{Mephisto}~\#3449, Clue 21A (\cref{sec:mephisto-eval,sec:error-analysis}) & \clue{Spice mix of old --- not a word about it locally (8)} (Across) & \ans{CARDAMUM}\newline(\textit{Spice}) & \textbf{Charade:} \ans{CARD} (\textit{mix of old}, archaic Chambers verb meaning to mix/blend) $+$ \ans{A-MUM} (\textit{not a word about it locally}: \textit{``not a word about it''} = silent/mum, \textit{``locally''} = Scots dialect indicator for Chambers headword \textit{a-mum}) $\to \ans{CARDAMUM}$ (models guessed \ans{CARDAMOM}). \\
\midrule
\textbf{C26} & \textit{Times}~\#29399, Clue 1D (\cref{sec:error-analysis}) & \clue{Pet? You shouldn't have one (4)} (Down) & \ans{TAME}\newline(\textit{Pet?}) & \textbf{Charade:} \ans{TA} (\textit{You shouldn't have!}, polite British expression of thanks) $+$ \ans{ME} (\textit{one}, first-person pronoun) $\to \ans{TAME}$ (models guessed \ans{TOMB} on pattern \texttt{T?M?}). \\
\midrule
\textbf{C27} & \textit{Times}~\#29435, Clue 5D (\cref{sec:error-analysis}) & \clue{Brought into line, book parts rather like Peter or Roger? (7)} (Down) & \ans{RABBITY}\newline(\textit{like Peter or Roger?}) & \textbf{Nested Container:} \ans{B} (\textit{book}) splits (\textit{parts}) \ans{A BIT} (\textit{rather} $\to \ans{AB[B]IT}$), all inserted (\textit{Brought into}) \ans{RY} (\textit{line}, railway) $\to \ans{R[AB(B)IT]Y}$ (like Peter/Roger Rabbit; models guessed \ans{RABBETS}). \\
\midrule
\textbf{C28} & \textit{Times}~\#29435, Clue 15D (\cref{sec:error-analysis}) & \clue{Scales one of these when going over rising ground (3,5)} (Down) & \ans{AIR SIGNS}\newline(\textit{Scales one of these}) & \textbf{Container + Anagram:} \ans{AS} (\textit{when}) containing (\textit{going over}) an anagram (\textit{ground}) of \texttt{RISING} ($\ans{IRSIGN}$) $\to \ans{A[IR SIGN]S}$ (Libra/Scales is an astrological air sign; \textsc{swp} filled \ans{ARRAIGNS} on \texttt{A?R?I?N?}). \\
\midrule
\textbf{C29} & Mick Hodgkin (\textit{The Times}, Oct 2026, \cref{sec:future-work}) & \clue{AI not foxed by Times cryptic --- respect! (10)} (Across) & \ans{ESTIMATION}\newline(\textit{respect!}) & \textbf{Compound Anagram:} Anagram (\textit{cryptic}) of \texttt{TIMES} ($\ans{ESTIM}$) placed before (\textit{by}, juxtaposed with) an anagram (\textit{foxed}) of \texttt{AI NOT} ($\ans{ATION}$) $\to \ans{ESTIM} + \ans{ATION} = \ans{ESTIMATION}$. \\
\bottomrule
\end{tabular}
\end{adjustbox}
\end{table}

\end{document}